\documentclass[10pt,twocolumn,letterpaper]{article}
\pdfoutput=1
\usepackage{cvpr}
\usepackage{times}
\usepackage{epsfig}
\usepackage{graphicx}
\usepackage{amsmath}
\usepackage{amssymb}
\usepackage{booktabs}
\usepackage{verbatim}
\usepackage{subcaption}
\usepackage{caption}

\newcommand{\R}{\mathbb{R}}
\let \bs=\mathbf
\let \set=\mathcal

\def \coplane {1}
\def \parall {2}
\def \perped {3}

\def \saliency {\textup{\saliency}}

\def \path {\mathit{path}}

\let \set = \mathcal
\let \bs = \boldsymbol

\usepackage[pagebackref=true,breaklinks=true,letterpaper=true,colorlinks,bookmarks=false]{hyperref}

\cvprfinalcopy 

\def\cvprPaperID{0178} 

\ifcvprfinal\fi
\begin{document}

\title{Extreme Relative Pose Network under Hybrid Representations}

\author{Zhenpei Yang\thanks{Both authors contributed equally}\\
UT Austin\\
{\tt\small yzp12@cs.utexas.edu}
\and
Siming Yan\footnotemark[1]\\
UT Austin\\
{\tt\small siming@cs.utexas.edu}
\and
Qixing Huang\\
UT Austin\\
{\tt\small huangqx@cs.utexas.edu}
}

\maketitle
\thispagestyle{empty}
\begin{abstract}

In this paper, we introduce a novel RGB-D based relative pose estimation approach that is suitable for small-overlapping or non-overlapping scans and can output multiple relative poses. Our method performs scene completion and matches the completed scans. However, instead of using a fixed representation for completion, the key idea is to utilize hybrid representations that combine 360-image, 2D image-based layout, and planar patches. This approach offers adaptively feature representations for relative pose estimation. Besides, we introduce a global-2-local matching procedure, which utilizes initial relative poses obtained during the global phase to detect and then integrate geometric relations for pose refinement. 
Experimental results justify the potential of this approach across a wide range of benchmark datasets. For example, on ScanNet, the rotation translation errors of the top-1/top-5 predictions of our approach are $28.6^{\circ}/0.90m$ and $16.8^{\circ}/0.76m$, respectively. Our approach also considerably boosts the performance of multi-scan reconstruction in few-view reconstruction settings.

\end{abstract}

\section{Introduction}
\label{Section:Introduction}

Estimating the relative pose between two RGB-D scans is a crucial problem in 3D vision and robotics. In this paper, we are interested in the setting where the overlapping region between two input scans is small, or they may not even overlap. Efficient and robust solutions to this extreme relative pose problem (c.f.~\cite{Yang_2019_CVPR}) enjoy a wide range of applications. Examples include 3D reconstruction from a few views~\cite{1044493} (e.g., RGB-D scans of an indoor scene captured at a few distinctive locations), enhancing the performance of interactive scanning~\cite{Rusinkiewicz:2002:RMA,Newcombe:2011:KRD} when there are interruptions during the acquisition process, early detection of loop closure \cite{angeli2008fast}, and solving jigsaw puzzles \cite{cho2010probabilistic}.

A standard pipeline in relative pose estimation or object matching, in general, is to first extract features from the input objects and then match these features to derive relative poses or dense correspondences~\cite{huang2017visual,andreasson2012real}. While early works~\cite{Gelfand:2005:RGR, Huang:2006:RFO} use hand-crafted features or matching procedures, recent works~\cite{Goforth-2019-112290,melekhov2017relative,ranftl2018deep} have focused on applying deep neural networks to aid this pipeline. The challenge in the extreme relative pose setting is that there are insufficient features to match, and one has to leverage priors about the underlying object/scene. In this paper, we focus on two aspects of utilizing deep neural networks to instill priors into feature extraction and feature matching.

\begin{figure}
\def\imh{0.16\textwidth}
\def\imw{0.16\textwidth}
\newcommand{\T}[1]{\raisebox{-0.5\height}{#1}}
\setlength{\tabcolsep}{1pt}
\begin{tabular}{ccc}

\T{\includegraphics[width=\imw]     {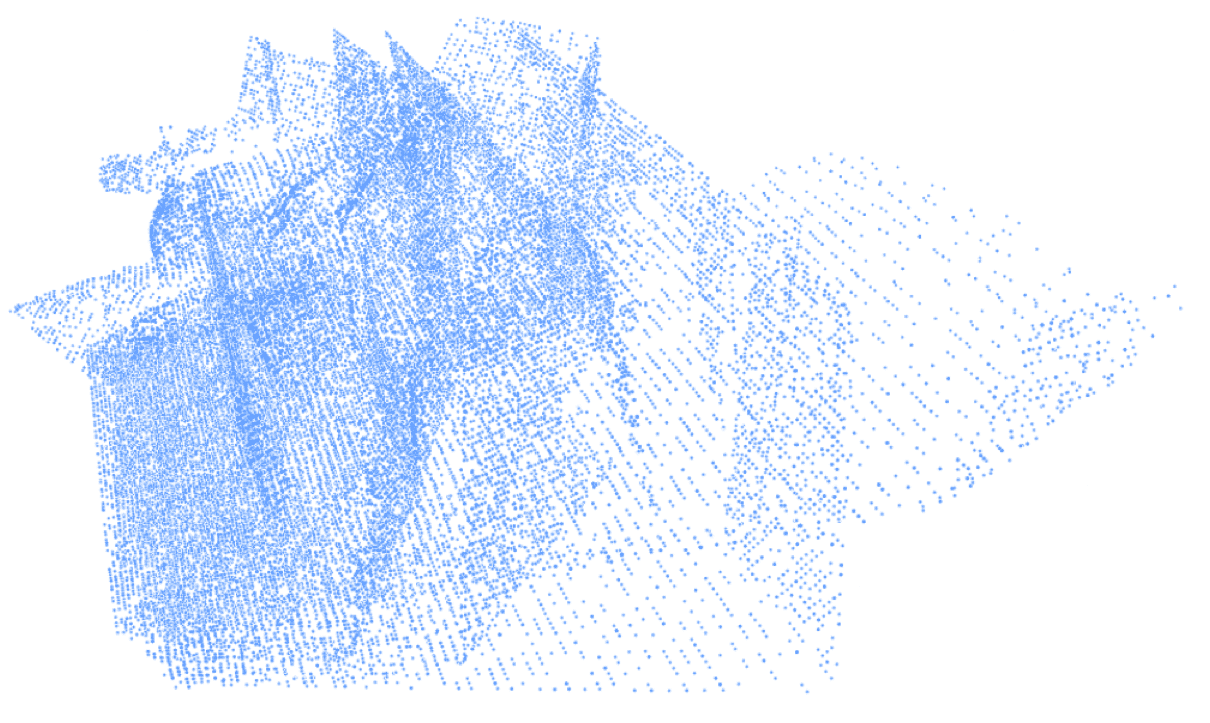}} & 
\T{\includegraphics[width=\imw]   {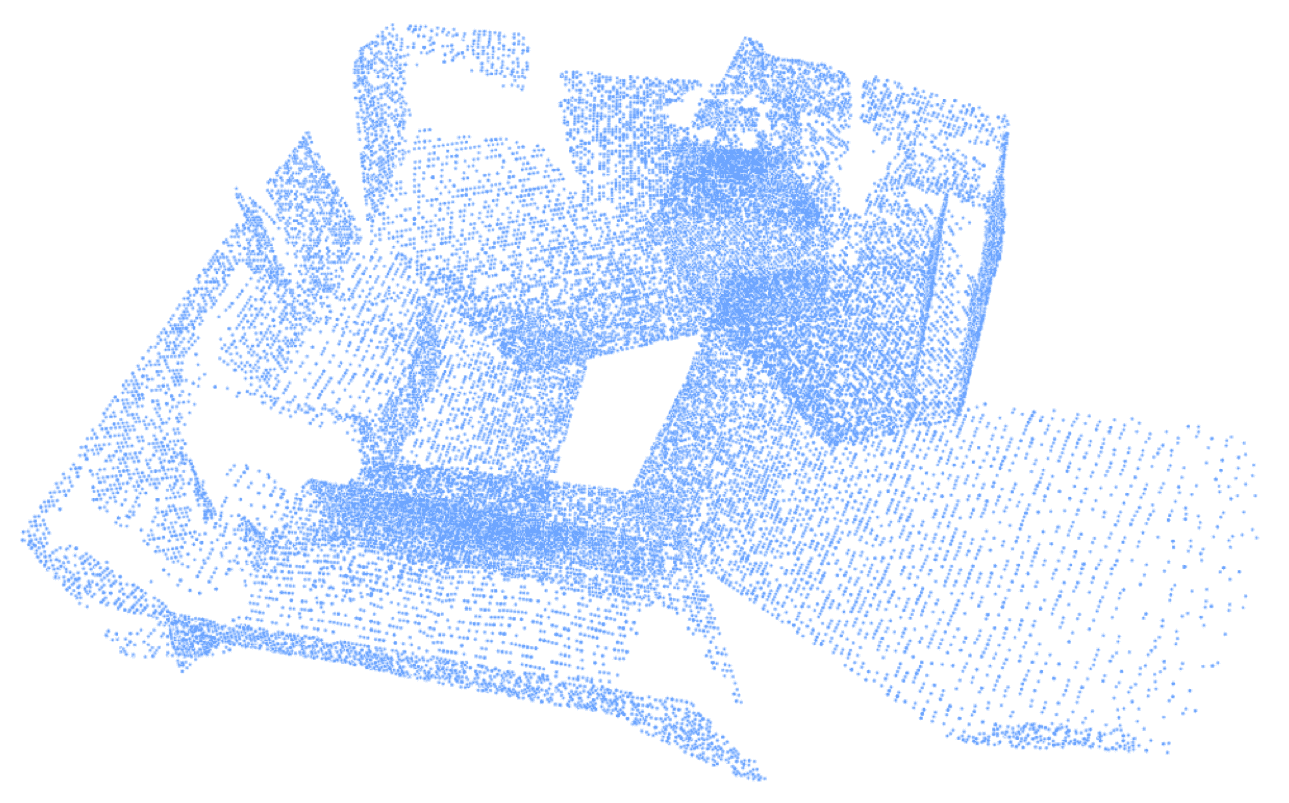}} & 
\T{\includegraphics[width=\imw]      {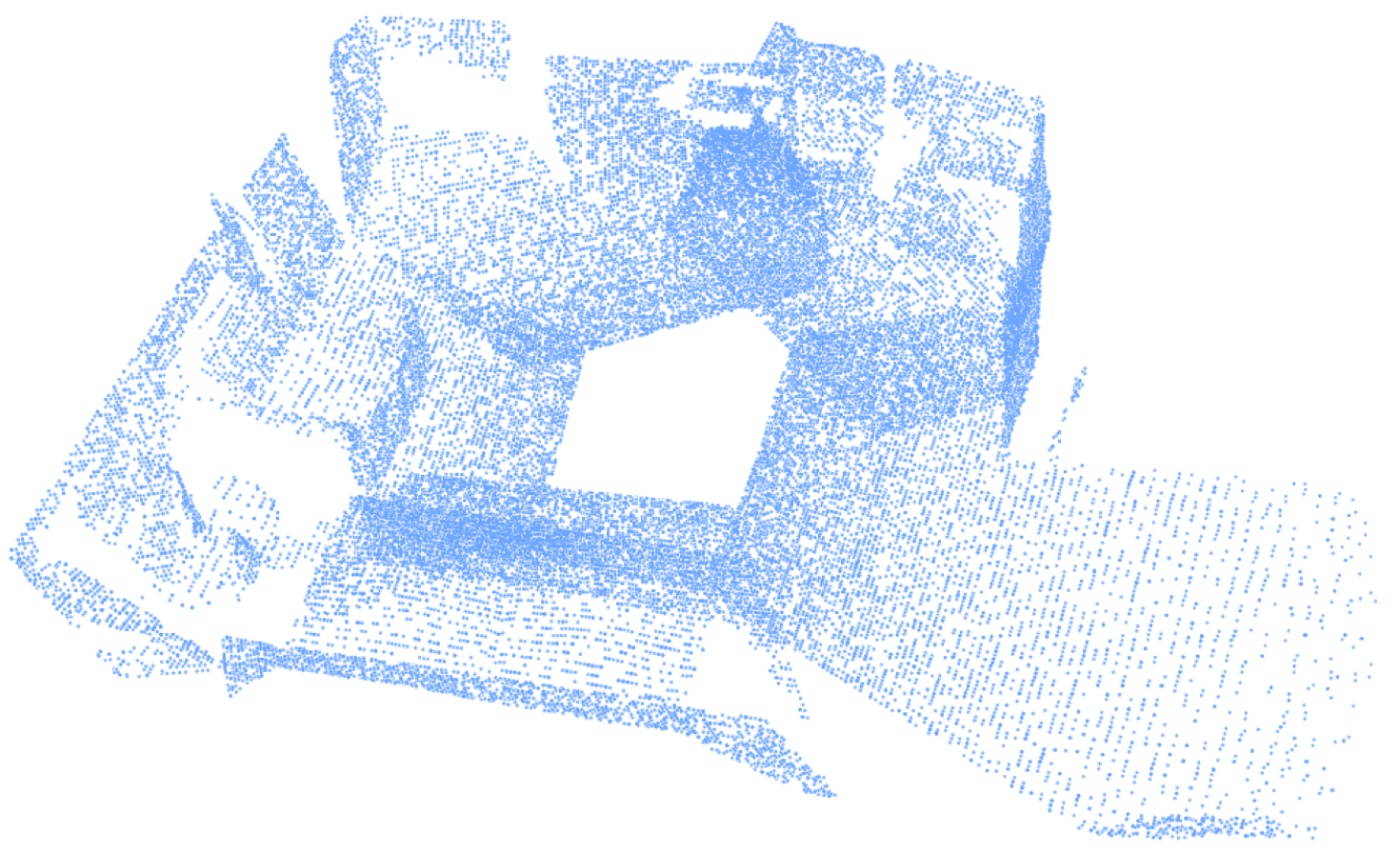}} \\

\T{\includegraphics[width=\imw]     {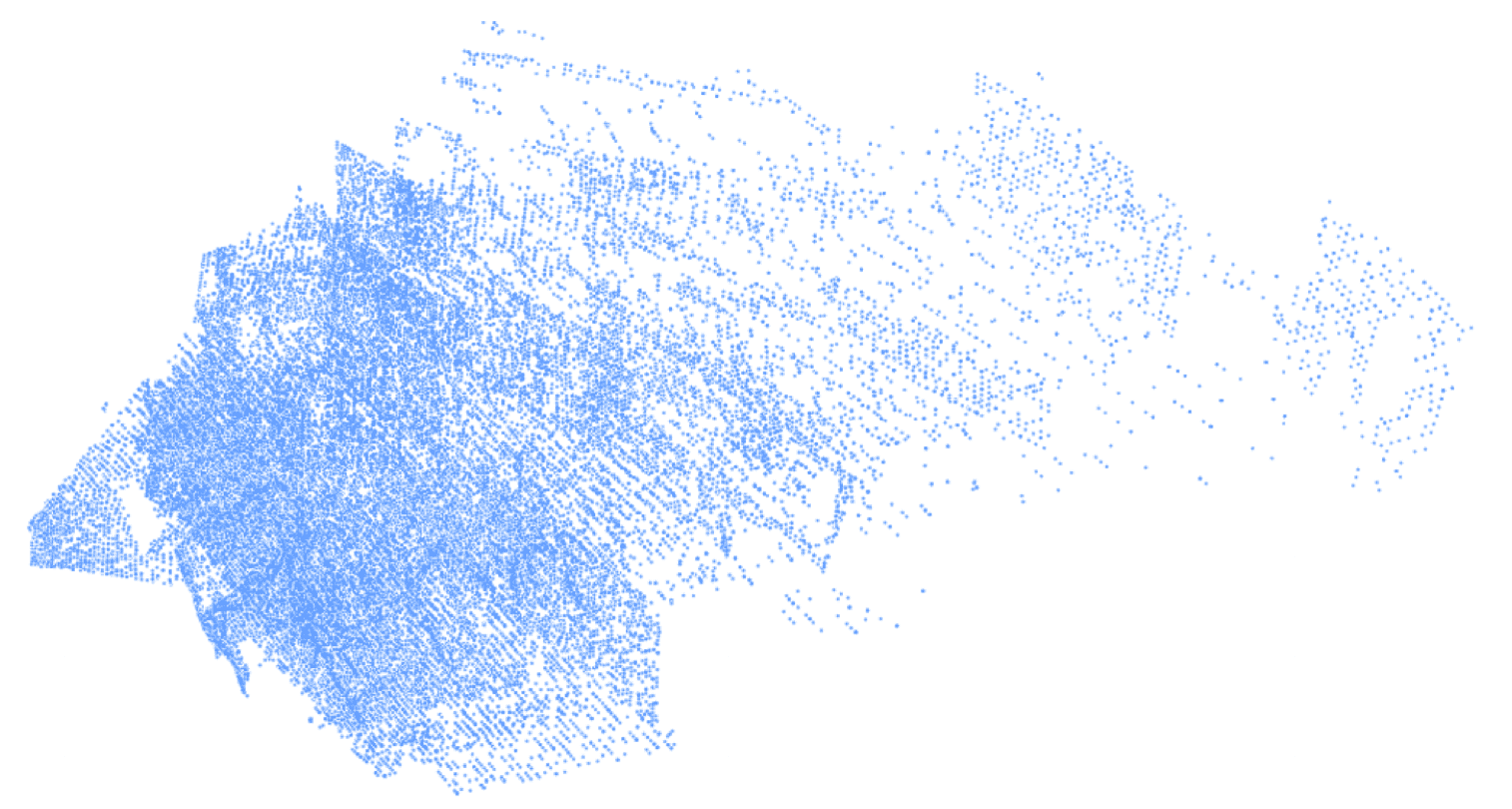}} & 
\T{\includegraphics[width=\imw]   {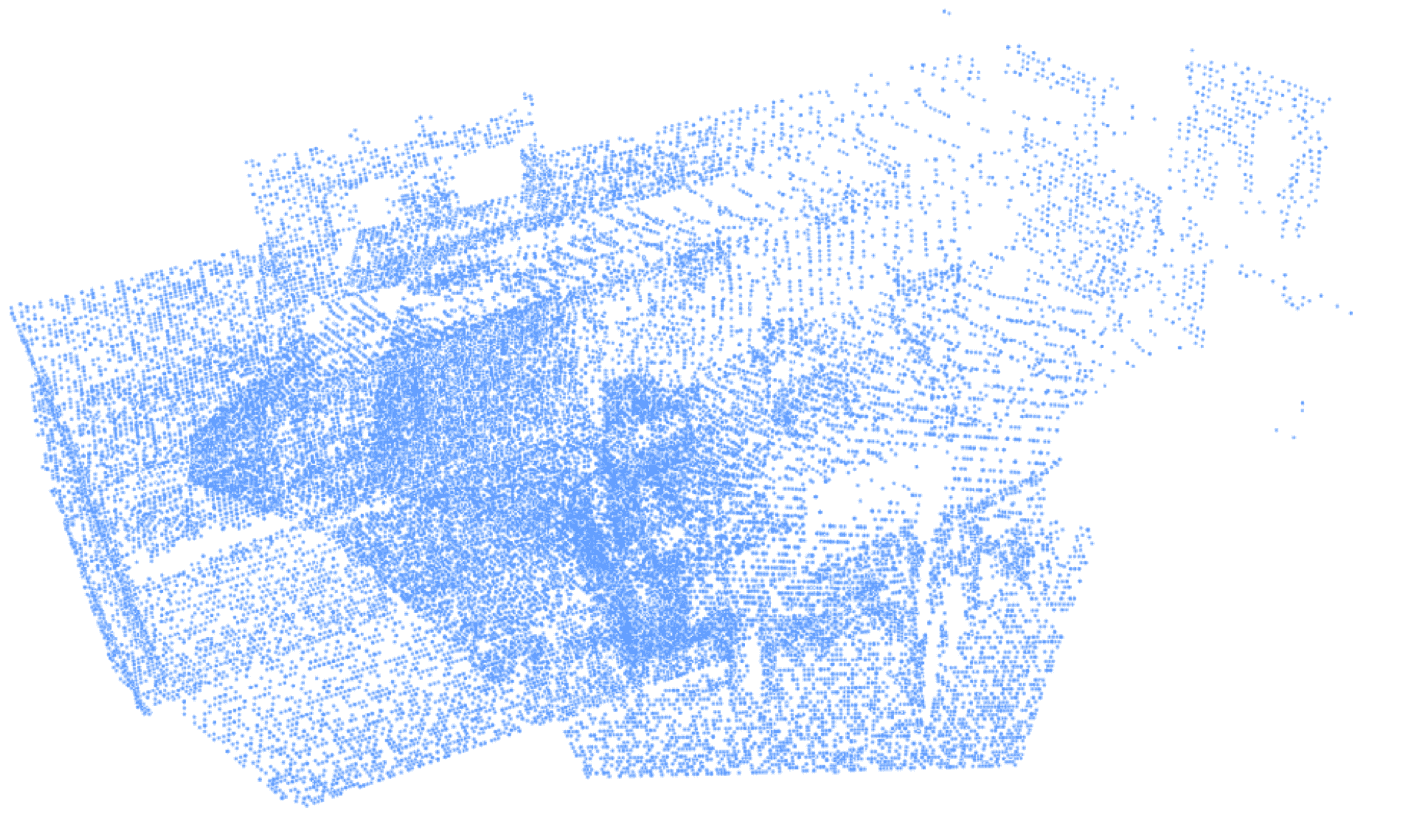}} & 
\T{\includegraphics[width=\imw]      {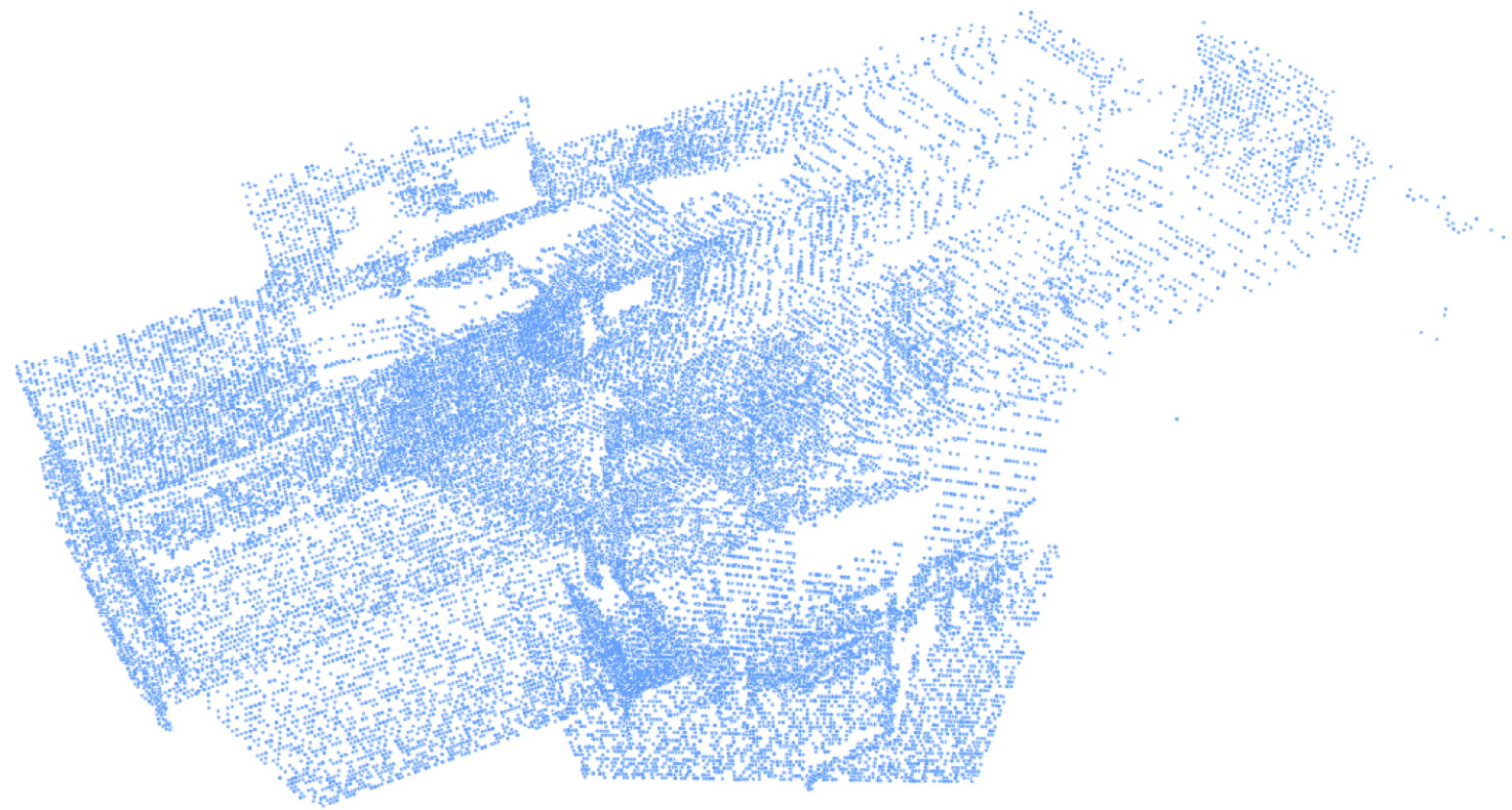}} \\

\T{\includegraphics[width=\imw]     {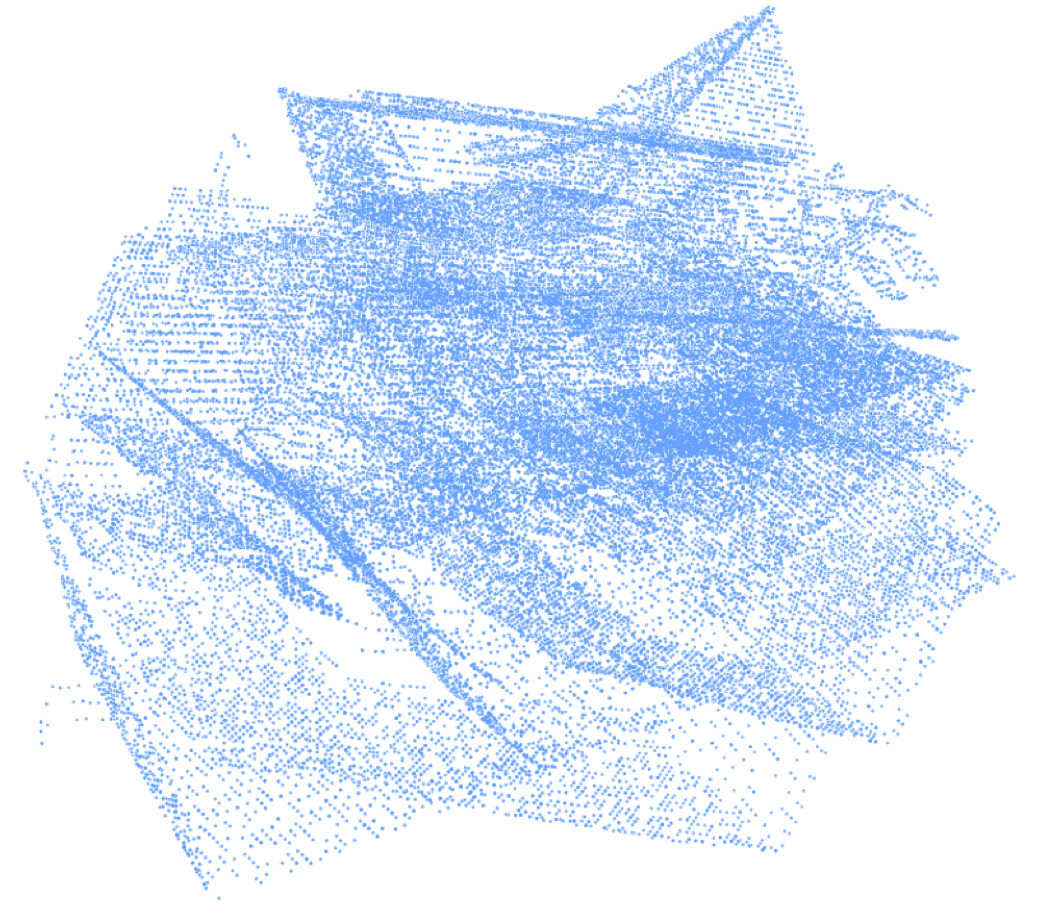}} & 
\T{\includegraphics[width=\imw]   {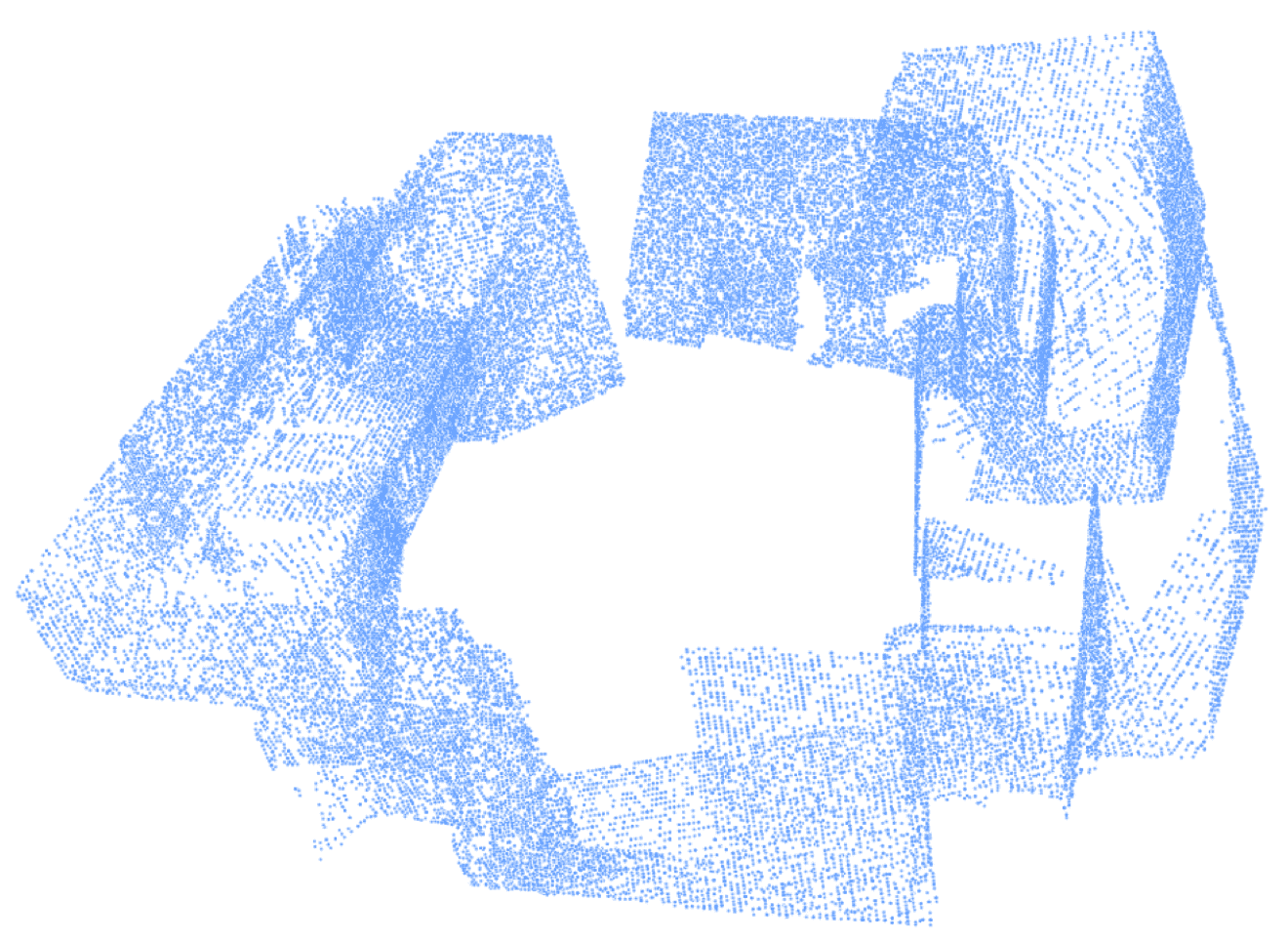}} & 
\T{\includegraphics[width=\imw]      {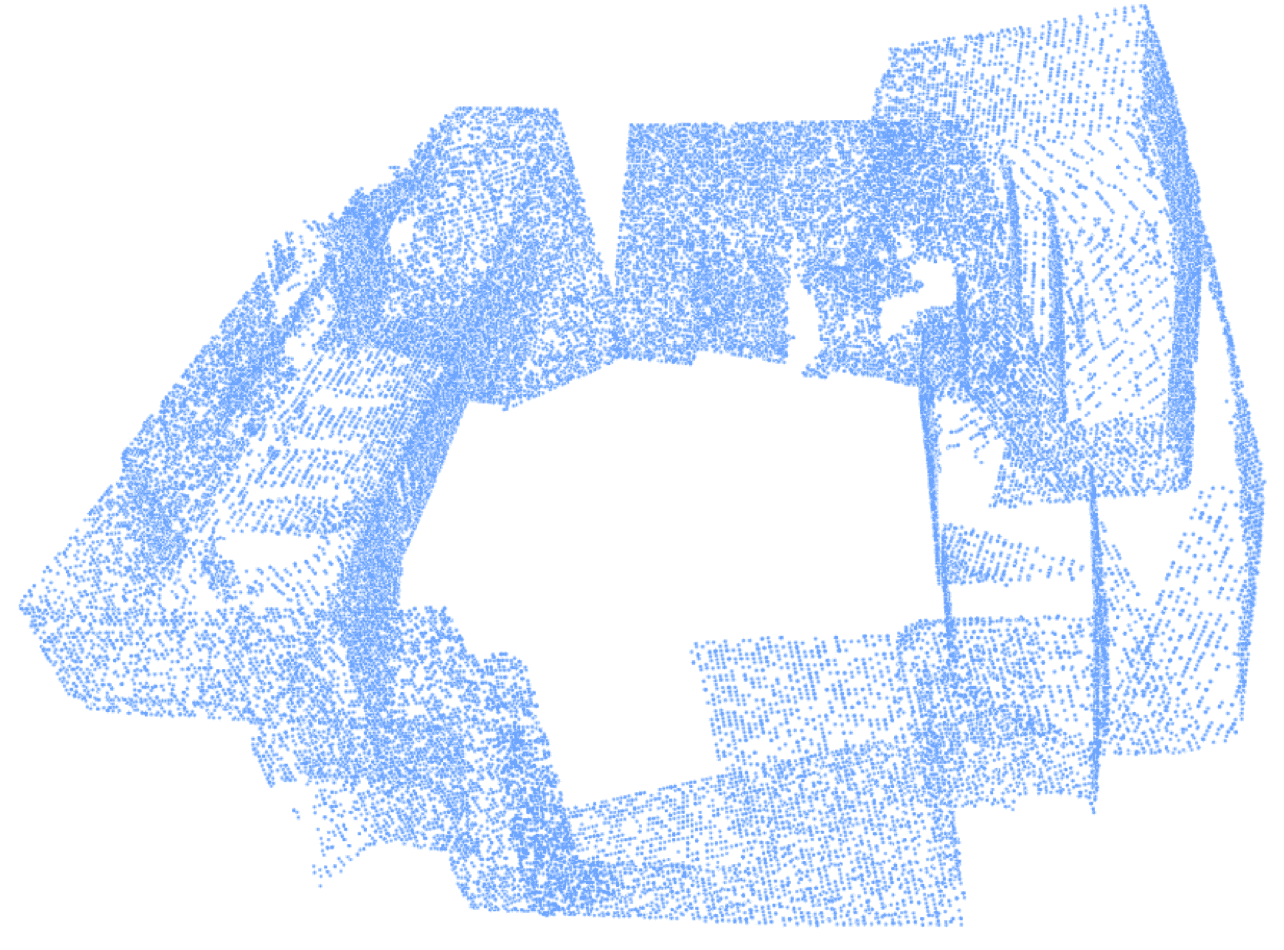}} \\

(a) RobustRecons & (b) Ours & (c) Ground-truth \\
\end{tabular}
\caption{Results of multi-scan alignment under the few-view setting. Each block shows the results from 5 randomly sampled scans of a 3D scene that show minimal overlaps. (a) RobustRecons~\protect\cite{choi2015robust}. (b) Our approach. (c) Ground-truth.}

\label{Fig:Few:View}
\vspace{-0.15in}
\end{figure}

The first aspect focuses on completing the input scans and matching completed scans~\cite{Yang_2019_CVPR}. This strategy learns from data to predict the surrounding regions of each input scan, turning non-overlapping scans into overlapping scans. A critical factor in this context is the data representation for scan completion. This representation not only dictates the extracted features for relative pose estimation but also affects the type of completion networks. Instead of focusing on one data representation, a key contribution of this paper is to combine multiple data representations (i.e., 360-image~\cite{Yang_2019_CVPR}, 2D layout ~\cite{ritchie2019fast}, and planar patches~\cite{liu2018planenet}) for scan completion. Our approach then extracts a subset of consistent features to compute the relative pose. The proposed hybrid representation exhibits two appealing advantages. First, it alleviates the issue of lacking a particular type of feature on the input scans (e.g., planar elements from non-planar scenes). Second, the matched features are adaptive to the input scan type and the relation between the input scans (e.g., dense point-wise features among overlapping regions and primitive features for non-overlapping scans).

The second aspect lies in the type of constraints between features for relative pose estimation. Instead of using a single module, our approach follows a global-2-local procedure. This procedure makes it possible to utilize different constraints at various pipeline stages. Specifically, the global matching module computes initial relative poses by performing spectral matching~\cite{Leordeanu:2005:SM,huang2008non} to align features extracted from the completed scans. The local refinement module, which benefits from having initial relative poses, detects and integrates additional pairwise geometric relations (e.g., parallel planes and distances between planes) for pose refinement. This sequential procedure can be easily modified to output multiple relative poses to model ambiguities in extreme relative pose estimation. 

We have evaluated our approach on scan pairs from three benchmark datasets, including SUNCG~\cite{song2016ssc}, Matterport~\cite{DBLP:journals/corr/abs-1709-06158}, and ScanNet~\cite{DBLP:journals/corr/DaiCSHFN17}. For randomly sampled scan pairs, the mean top-1 rotation/translation errors of our approach are $18.1^{\circ}/0.16m$, $22.3^{\circ}/0.39m$, and $28.6^{\circ}/0.90m$, on SUNCG, Matterport, and ScanNet, respectively. In contrast, the state-of-the-art approach achieved $31.1^{\circ}/0.27m$, $34.2^{\circ}/0.53m$, and $34.1^{\circ}/0.93m$, respectively. These improvements show the advantage of using a hybrid 3D representation for relative pose estimation. Moreover, the mean top-5 errors in rotation/translation dropped to $7.8^{\circ}/0.21m$, $14.3^{\circ}/0.41m$, and $16.8^{\circ}/0.76m$. These performance gains show that ambiguities are abundant in extreme relative pose estimation, and the multiple solutions returned by our approach can effectively address such ambiguities. 

Besides, we demonstrate the usefulness of our approach in multi-scan reconstruction. Experimental results show that combing our method and the pose optimizer of MRF-SFM~\cite{crandall2013pami} enables faithful 3D reconstructions from a few views and outperforms state-of-the-art multi-scan reconstruction approaches (See Figure~\ref{Fig:Few:View}).

\section{Related Works}
\label{Section:Related:Works}

\noindent\textbf{Relative pose estimation via feature matching.} Global relative pose estimation approaches usually fall into two categories (c.f.~\cite{li2015computing,van2011survey}). The first category of approaches leverages global descriptors or performs Fourier transform to align two input scans. These methods are most suitable for matching complete objects. The second category of approaches performs feature matching, e.g., RANSAC~\cite{fischler1981random}, robust regression~\cite{bouaziz2013sparse}, and spectral matching~\cite{huang2008non, leordeanu2005spectral}, to estimate relative poses. However, these approaches still require that the input scans possess considerable overlapping regions. In contrast, this paper focuses on relative pose estimation between potentially non-overlapping scans. 

\noindent\textbf{Relative pose estimation via local optimization.} Besides global matching, another category of relative pose estimation approaches focuses on local pose refinement~\cite{journals/tvcg/TamCLLLMMSR13}. A standard method is geometric alignment~\cite{Besl:1992:MRS,Chen:1992:OMR}, which minimizes point-wise distances between a pair of scans in their overlapping region. However, existing approaches typically require that the input scans are overlapping (c.f.~\cite{journals/tvcg/TamCLLLMMSR13}). In contrast, this paper proposes to solve the local pose refinement problem by learning and enforcing geometric relation (perpendicular, parallel, co-planar), which is suitable for both overlapping and non-overlapping scans.

\begin{figure*}

\includegraphics[width=\textwidth]{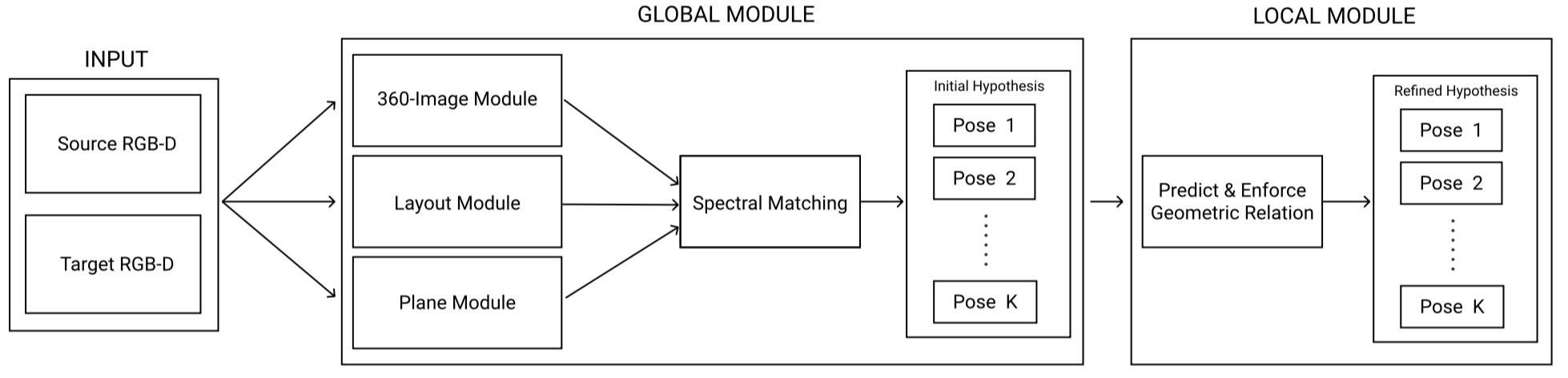}
\caption{\small{Our approach takes two potentially non-overlapping RGB-D scans as input and outputs one or multiple relative poses between them. The proposed pipeline combines a global module and a local module. The global module takes a pair of RGB-D scans as input and outputs multiple relative poses between them. The local module refines each relative pose by enforcing predicted geometric relations.}}
\vspace{-0.15in}
\label{Figure:Overview}
\end{figure*}

\noindent\textbf{Learning based relative pose estimation.} Thanks to advances in deep learning that provide powerful tools to establish maps between different domains, recent works~\cite{melekhov2017relative, Ummenhofer_2017_CVPR,Zhou_2017_CVPR, kim2018recurrent, ranftl2018deep} formulate relative pose estimation as training pose estimation networks. Early works~\cite{melekhov2017relative, Ummenhofer_2017_CVPR,Zhou_2017_CVPR} usually follow the procedure of feature extraction and then relative pose estimation using a correlation module. Recent works~\cite{kim2018recurrent, ranftl2018deep} employ more sophisticated matching techniques such as recurrent neural networks. However, most existing approaches still require that the two input scans possess significant overlaps, with the noticeable exception of \cite{Lin_2019_ICCV}, which first estimate local layout for partial scans then use hand-coded floorplane priors to assembly all pieces. Our approach differs in that we leverage the priors by explicitly learning a hybrid completion components from data.

This work is most relevant to~\cite{Yang_2019_CVPR}, which presents a relative pose estimation approach that first performs scene completion and then computes a single relative pose prediction via geometric matching on the completed scenes. The difference in this work is that we employ a hybrid 3D scene completion in contrast to the 360-image employed in~\cite{Yang_2019_CVPR}. Besides, our approach introduces a novel geometric matching procedure that outputs multiple solutions and a  learned local pose refinement module.  

\noindent\textbf{Scene completion.} Our approach is also motivated by recent advances in inferring complete environments from partial observations~\cite{DBLP:journals/corr/PathakKDDE16,dinesh-lookaround,song2016im2pano3d,DBLP:journals/corr/abs-1709-00505,Zou_2018_CVPR}. However, our approach differs from these approaches in two ways. First, in contrast to returning the completion as the final output~\cite{song2016im2pano3d,Zou_2018_CVPR} or utilizing it for learning feature representations~\cite{DBLP:journals/corr/PathakKDDE16,DBLP:journals/corr/abs-1709-00505} or motion policies~\cite{dinesh-lookaround}, our approach treats completions as an intermediate representation for feature extraction and detecting geometric relations. More importantly, instead of using a single representation for scene completion, our approach utilizes multiple representations and lets the geometric matching module to pick the most suitable representations for relative pose estimation.

\section{Approach}
\label{Section:Approach}

This section describes the technical details of our approach. Section~\ref{Section:Overview} presents an overview of our approach. Section~\ref{Section:Global:Module} to Section~\ref{Subsection:Network:Training} discuss each component in depth. 

\subsection{Problem Statement and Approach Overview}
\label{Section:Overview}

Our approach takes two RGB-D scans $S_1$ and $S_2$ of the same scene as input and outputs multiple rigid transformations $\set{T}_{12} = \{T_{12}\}\subset SE(3)$ that align the two input scans. As illustrated in Figure~\ref{Figure:Overview}, our approach combines a global module, which generates a set of candidate relative poses, and a local module, which refines each candidate relative pose. In doing so, our approach can output multiple relative poses.

\noindent\textbf{Global module.} This module follows the standard pipeline of first extracting features from the input scans and then matching the extracted features. It begins with performing scan completion to generate sufficient shared features between small-overlapping or non-overlapping scans. In particular, our approach utilizes three intermediate representations, i.e., 360-image~\cite{Yang_2019_CVPR}, 2D layout~\cite{ritchie2019fast}, and planar primitives~\cite{liu2018planenet}, for scan completion. Given the completed scans, this module then establishes consistent correspondences between extracted features and fit a rigid transformation to each consistent set. To this end, we apply spectral matching, which extracts consistent subsets of feature correspondences by computing leading eigenvectors of a consistency matrix among candidate correspondences. Each resulting correspondence set generates an initial relative pose by rigid pose regression.

\noindent\textbf{Local module.} In the same spirit as using ICP~\cite{Besl:1992:MRS} to optimize relative poses, our approach employs a local module to refine each output of the global module. The motivation comes from the fact initial relative poses enable additional constraints, such as iteratively computed dense correspondences and geometric relations (i.e., co-planar, parallel planes, and perpendicular planes), for relative pose estimation. Our approach also employs robust norms to remove outliers in dense correspondences and geometric relations. 

\noindent\textbf{Network training.} A fundamental property of our approach is that the local module solves a robust optimization problem. Its optimal solution is generally insensitive to small perturbations to the outputs of the global module. Thus, we introduce two objectives for network training; namely,  the optimal solution of the local module is close to the underlying ground-truth, and the global module provides effective initial solutions. The total objective function for network training consists of three terms. The first term trains the completion networks. The second term jointly trains the completion networks and the spectral matching procedure. The third term is enforced on the output of the local module. 
\subsection{Global Module}
\label{Section:Global:Module}

The global module computes a set of relative poses $\set{T} = \{(R_{12},\bs{t}_{12})\}$ between the input scans $S_1$ with $S_2$. This module consists of two sub-modules. The first sub-module performs scan completion and feature extraction from each input scan $S_i$. The result is summarized into a collection of features $\set{F}_{\theta_g}(S_i) = \{f\}$, where $\theta_g$ denotes network parameters. Each $f =\{ \bs{p}_f,\bs{n}_f,\bs{d}_f\}$ consists of a 3D position $\bs{p}_{f}$, a 3D normal $\bs{n}_f$, and a feature descriptor $\bs{d}_f$. Note that although all extracted features share the same encoding, this sub-module combines the strengths of scan completion networks that are available under multiple representations (e.g., 360-image, 2D-layout, and 3D planar patches). 

The second sub-module computes initial relative poses from correspondences between these extracted features. Since the local module will refine each relative pose, this sub-module uses spectral matching~\cite{Leordeanu:2005:SM,huang2008non}, which is efficient and can output multiple relative poses.

\begin{figure}
\includegraphics[width=\columnwidth]{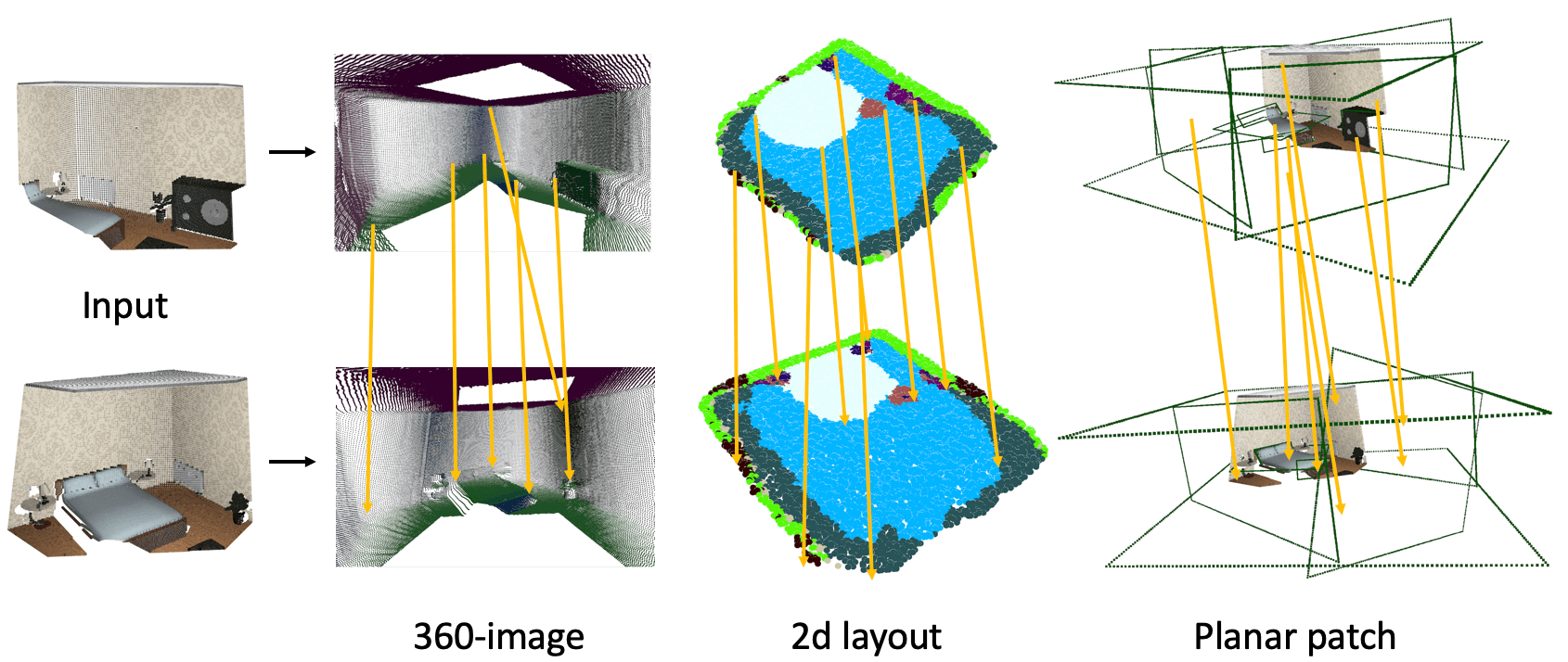}
\caption{\small{Matched correspondences between two scans across different representations. All representations help. Note that to make the visualization uncluttered, we only show a subset of computed correspondences and draw edges of the walls. }}
\label{Figure:Global:Module}
\vspace{-0.15in}
\end{figure}

\noindent\textbf{Scan completion sub-module.} Our approach considers three data representations (See Figure~\ref{Figure:Global:Module}). Since we primarily use state-of-the-art scan completion results, we only highlight the main characteristics of each completion network and defer the details to the supp. material. 

The first representation is 360-image~\cite{song2016im2pano3d,Yang_2019_CVPR}, which represents a given 3D scene as a collection of multi-channel images. The same as~\cite{Yang_2019_CVPR}, we encode for each pixel its position $\bs{p}_f$, normal $\bs{n}_f$. The completion network admits an encoder-decoder architecture and outputs predicted position, normal, and descriptor channels. This representation can generate dense completions. Besides, it is a generic representation that does not place any assumptions about the underlying scene.  However, the quality of the predicted features drops substantially among regions that are far from the visible area of each scan (c.f.~\cite{Yang_2019_CVPR}).

The second representation is 2D layout. We first to predict the plane of the floor of each scan. We then turn each input scan into an top-down view image by projecting points of the scan onto a grid representation of the floor. Similar to \cite{lang2019pointpillars}, the feature descriptor of each pixel collects the averaged feature among 4 vertical bins. We train the feature descriptor jointly with contrastive loss same as ~\cite{Yang_2019_CVPR} and semantic segmentation loss. The completion network again employs an encoder-decoder architecture. 

The third representation is planar patches~\cite{song2016im2pano3d,shi2018planematch,liu2018planenet}. To generate this representation, we perform RANSAC~\cite{Fischler:1981:RSC} to extract planar patches from each input scan and the corresponding complete 3D scene. We then perform PCA on the points associated with each patch to extract its patch center. The position $\bs{p}_f$ and normal $\bs{n}_f$ of this planar feature are simply the patch center and plane normal, respectively. Its feature descriptor $\bs{d}_f$ is the mean feature descriptor among the points associated with this planar patch. The completion network employs a variant of PointNet~\cite{qi2017pointnet} by treating each plane as a generalized point. Experimentally, we found that this planar patch representation leads to better generalization performance than the first two representations, but it only involves planar patches.

\noindent\textbf{Spectral matching sub-module.} The output of the completion sub-module is a feature set $\set{F}_{\theta_g}(S_i)$ for each scan $S_i$ (that include dense pixels and planar features). Let $\overline{\set{M}}_{\theta_g, p} \subset \set{F}_{\theta_c}(S_1)\times \set{F}_{\theta_g}(S_2)$ collect all correspondences between the same type of features for the scan pair $p = (S_1, S_2)$. For representations of 360-image and 2D layout, we match SIFT keypoint among the visible region of one scan to the closest point on the other scan (in terms of feature descriptors) to reduce the number of feature correspondences. As the global module only provides initial solutions for the local module, we found that increasing the number of feature matches has minor impacts on the results of the local module. 

The spectral matching sub-module extracts consistent correspondence sets $\set{M}_k \subset \overline{\set{M}}_{\theta_g, p}, 1\leq k \leq K$ ($K = 5$ in our experiments). Each correspondence set generates one initial relative pose. This step follows~\cite{Leordeanu:2005:SM,Yang_2019_CVPR}, which construct a consistency matrix $C_{\theta_g,\alpha_{g}, p} \in \R^{|\overline{\set{M}}_{\theta_g, p}|\times |\overline{\set{M}}_{\theta_g, p}|}$. The value of each element 
$$
C_{\theta_g,\alpha_{g}, p}(c,c') = s_{c} s_{c'} g_{c,c'}
$$
where c and c' represent a pair of correspondences, $s_c$ models the descriptor similarity of $c$, $g_{c,c'}$ measures the preservation of distances and angles between $c$ and $c'$, and $\alpha_{g}$ represent hyper-parameters of $s_c$ and $g_{c,c'}$. Due to the space constraint, please refer to the supp. material for details. 

Given $C_{\theta_g,\alpha_{g}, p}$, we follow~\cite{Leordeanu:2005:SM,Yang_2019_CVPR} to extract top K matches, each of which is given by an indicator vector $\bs{u}_k \in [0,1]^{|\overline{\set{M}}_{\theta_g, p}|}$. We employ the rigid regression formulation in~\cite{Horn:1987:CFSQ} to obtain the initial relative pose $R_k^{\star}, \bs{t}_k^{\star}$:
\begin{align}
\underset{R,\bs{t}}{\min}\ \sum\limits_{c=(f_1,f_2)} & u_k(c) \|R\bs{p}_{f_1} + \bs{t} - \bs{p}_{f_2}\|^2  
\label{Eq:robust:regression}
\end{align}

\subsection{Local Module}
\label{Subsection:Local:Module}

The local module refines each initial relative pose obtained from the global module. Benefited from having initial relative poses, this module employs different sets of features for refinement. Specifically, it performs robust regression on iteratively updated closest point pairs and predicted geometric relations (e.g., point pairs that lie on the same plane) between the completed scans. In the following, we first introduce how to predict geometric relations. We then describe how to perform robust regression.

\noindent\textbf{Geometric relations.} We consider three types of geometric relations that frequently occur in man-made scenes:
point pairs that share the same plane, point pairs that lie on perpendicular planes, and point pairs that lie on parallel planes. Consider a down-sampled set of features $\set{F}^{g}_{\theta_g}(S_i)\subset \set{F}_{\theta_g}(S_i)$ for each scan $S_i$. Our implementation uses 6000 random samples for each scan pair. 

Given all feature pairs $\hat{\set{M}}^l_{\theta_g,p}$ between these features, the local module trains a variant of PointNet~\cite{qi2017pointnet} to classify each pair $c \in \set{M}^{l}_{\theta_g, p}$ into either no relation or one of the relations described above, i.e., four categories in total. In the following, we denote the probability of $c$ belonging to the categories of co-planar, parallel, and perpendicular as $w_{\theta_{l}}^{\coplane}(c)\in [0,1]$, $w_{\theta_{l}}^{\parall}(c)\in [0,1]$, and $w_{\theta_{l}}^{\perped}(c)\in [0,1]$, respectively, where $\theta_l$ denotes the network parameters of the relation prediction network.

\noindent\textbf{Robust regression.} We begin with defining consistency scores between a predicted geometric relation $c = (f_1, f_2)$ and the current rigid transformation $(R,\bs{t})$:
\begin{align}
f^{\coplane}_{R,\bs{t}}(c):= & \ ((R\bs{p}_{f_1}+\bs{t}-\bs{p}_{f_2})^T\bs{n}_{f_2})^2 \nonumber \\
&\ + ((R\bs{p}_{f_1}+\bs{t}-\bs{p}_{f_2})^TR\bs{n}_{f_1})^2 \\
f^{\parall}_{R}(c) := & \ 1 - ((R\bs{n}_{f_1})^T\bs{n}_{f_2})^2\\
f^{\perped}_{R}(c) := & \ ((R\bs{n}_{f_1})^T\bs{n}_{f_2})^2
\end{align}
In other words, $f_{\coplane}$, $f_{\parall}$, and $f_{\perped}$ force a point pair to satisfy the co-planar, parallel, and perpendicular relations, respectively. In addition, $f_{\coplane}$ also serves as a symmetric point-to-plane measure for matching closest point pairs (c.f~\cite{Rusinkiewicz:2019:ASO}). 

The robust regression formulation generalizes the standard approach of using iteratively reweighted non-linear least squares for rigid alignment~\cite{journals/tvcg/TamCLLLMMSR13}. Specifically, at each iteration, it solves the following non-linear least squares:
\begin{align}
\min\limits_{R, \bs{t}}  \sum\limits_{c\in \set{M}_{\theta_c}^{nn}}w_c^0 f^{\coplane}_{R,\bs{t}}(c) + \alpha_r \big(\sum\limits_{c\in \set{M}_{\theta_c}^{\coplane}}w_c^{1} w_{\theta_{g}}^{\coplane}(c) f^{\coplane}_{R,\bs{t}}(c)  \nonumber \\
+ \sum\limits_{c\in \set{M}_{\theta_c}^{\parall}}w_c^{2} w_{\theta_{g}}^{\parall}(c) f^{\parall}_{R}(c)+ \sum\limits_{c\in \set{M}_{\theta_c}^{\perped}}w_c^{3} w_{\theta_{g}}^{\perped}(c) f^{\perped}_{R}(c)\big)
\label{Eq:Total:Loss1}
\end{align}
where $\set{M}_{\theta_c}^{nn}$ collects nearest neighbor pairs from features of $\set{F}_{\theta_c}(S_1)$ to $\set{F}_{\theta_c}(S_2)$, and vice-versa, based on the relative pose at the previous iteration. $w_c^0$, $w_c^1$, $w_c^2$, and $w_c^3$ are updated adpatively based on the current relative pose $R, \bs{t}$:
\begin{align}
w_{c}^0 := \alpha_0^2/(\alpha_0^2 + f^{\coplane}_{R,\bs{t}}(c)), \ w_{c}^1 := \alpha_1^2/(\alpha_1^2 + f^{\coplane}_{R,\bs{t}}(c))\nonumber \\
w_{c}^2 := \alpha_2^2/(\alpha_2^2 + f^{\parall}_{R}(c)), \ w_{c}^3 := \alpha_3^2/(\alpha_3^2 + f^{\perped}_{R}(c))\label{Eq:Weight:Update}
\end{align}
where $\alpha_{l} = \{\alpha_r, \alpha_0,\alpha_1,\alpha_2,\alpha_3\}$ are hyper-parameters, which will be trained together with  $\theta_g$ and $\theta_l$.

With this setup, the robust regression procedure first applies (\ref{Eq:Weight:Update}) to set the term weights using the initial relative pose. It then alternates among computing the nearest neighbors, applying the Gauss-Newton method to solve (\ref{Eq:Total:Loss1}), and updating the term weights. In our implementation, we run 3 alternating iterations. 
\subsection{Network Training}
\label{Subsection:Network:Training}

\begin{figure*}
\centering
\footnotesize
\def\imh{0.073\textwidth}
\def\imw{0.14\textwidth}
\newcommand{\T}[1]{\raisebox{-0.5\height}{#1}}
\setlength{\tabcolsep}{1pt}
\begin{tabular}{ccccccc}

\T{\includegraphics[width=\imw]     {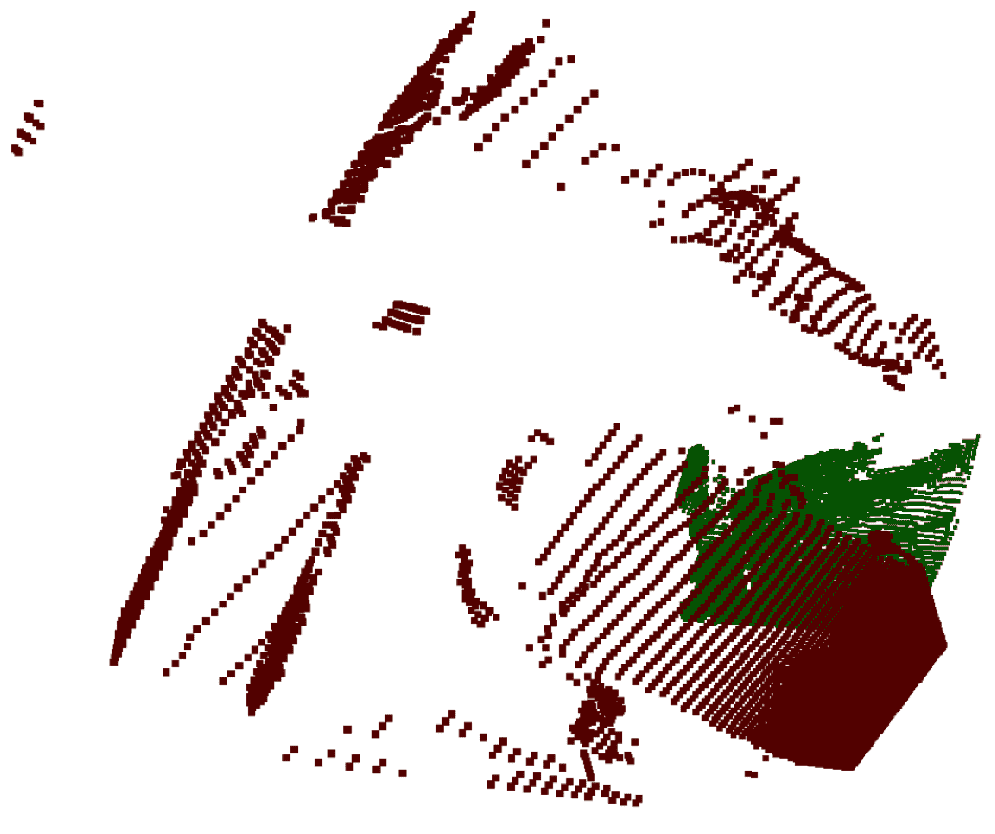}} & 
\T{\includegraphics[width=\imw]   {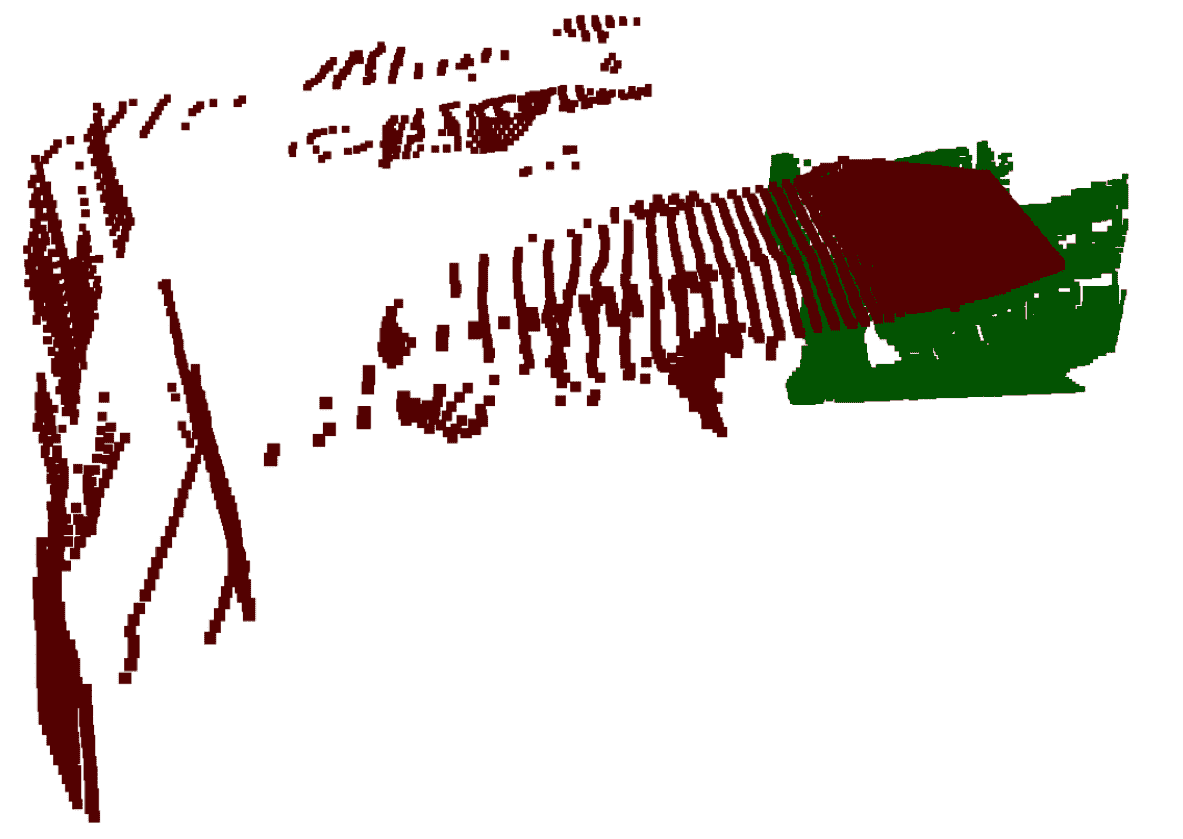}} & 
\T{\includegraphics[width=\imw]      {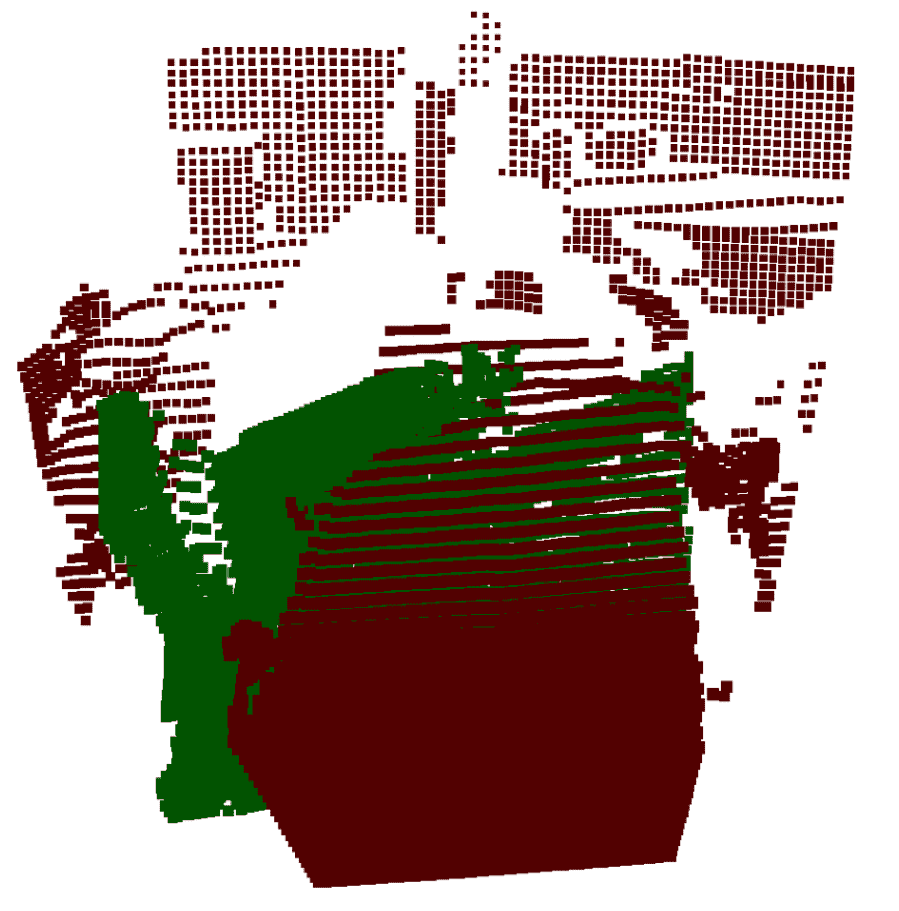}} & 
\T{\includegraphics[width=\imw]    {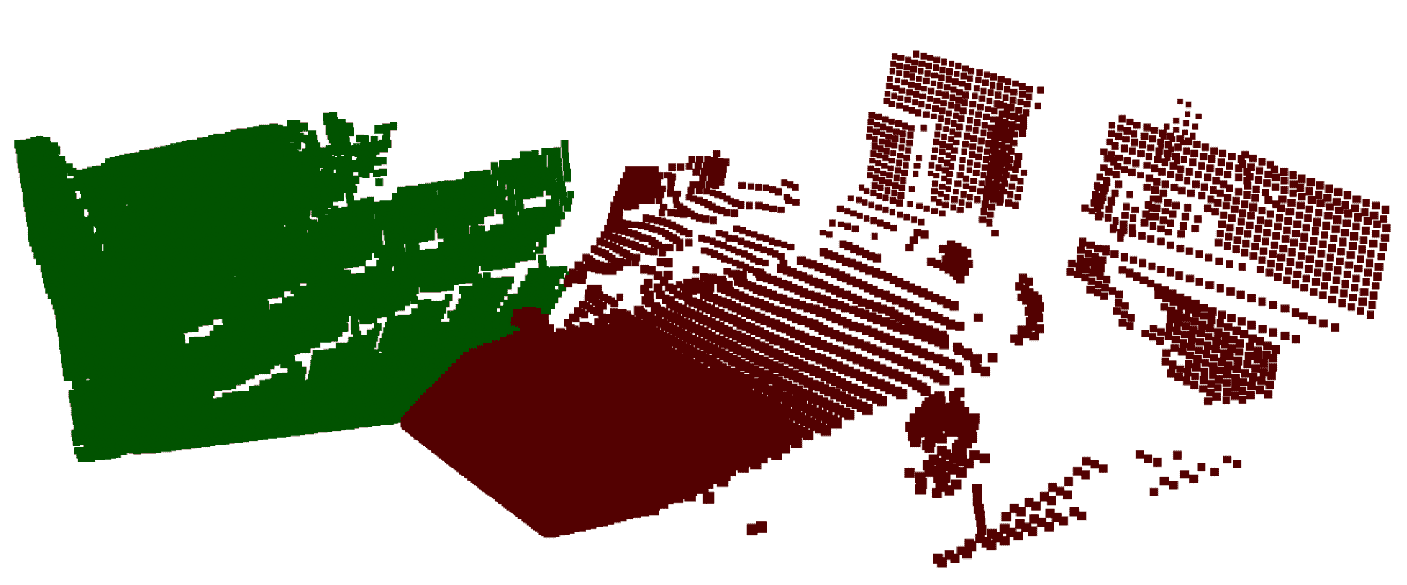}} & 
\T{\includegraphics[width=\imw]      {figures/fig5/0169_top1.png}} & 
\T{\includegraphics[width=\imw]     {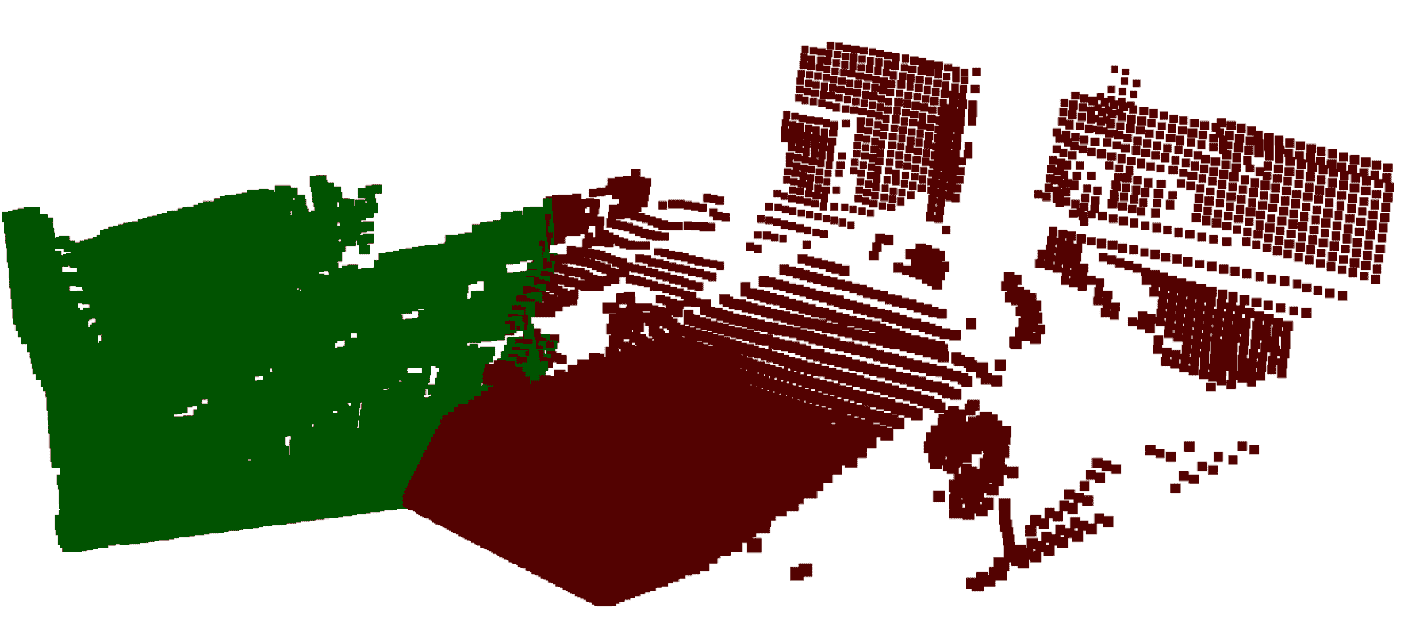}} & 
\T{\includegraphics[width=\imw]    {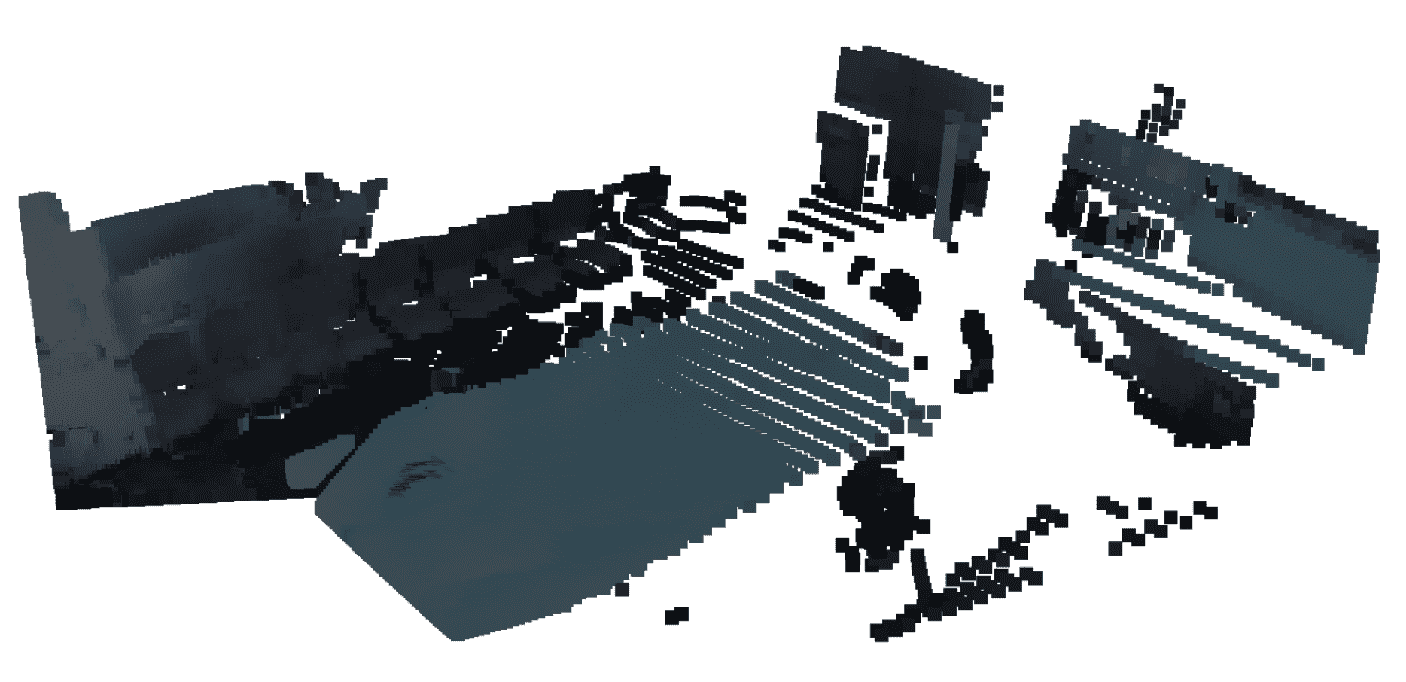}} \\
[+21pt]

\T{\includegraphics[width=\imw]     {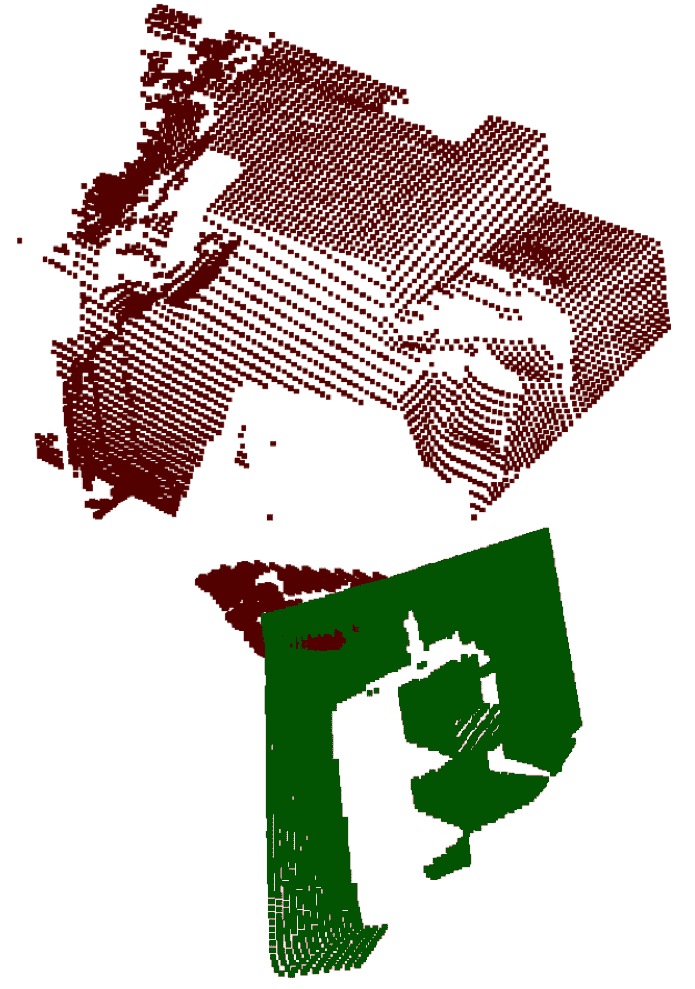}} & 
\T{\includegraphics[width=\imw]   {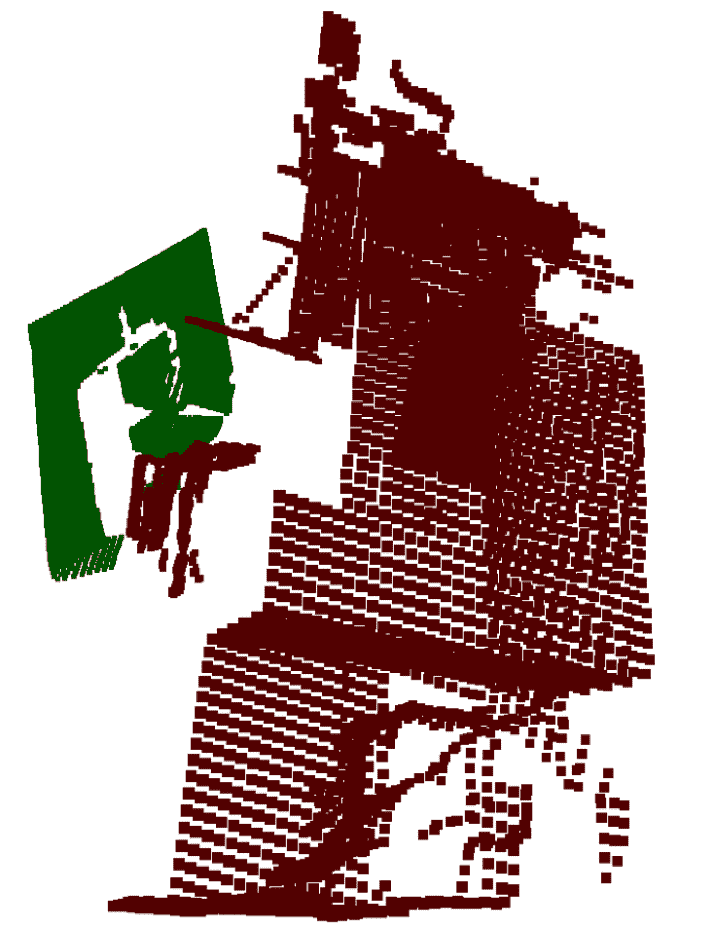}} & 
\T{\includegraphics[width=\imw]      {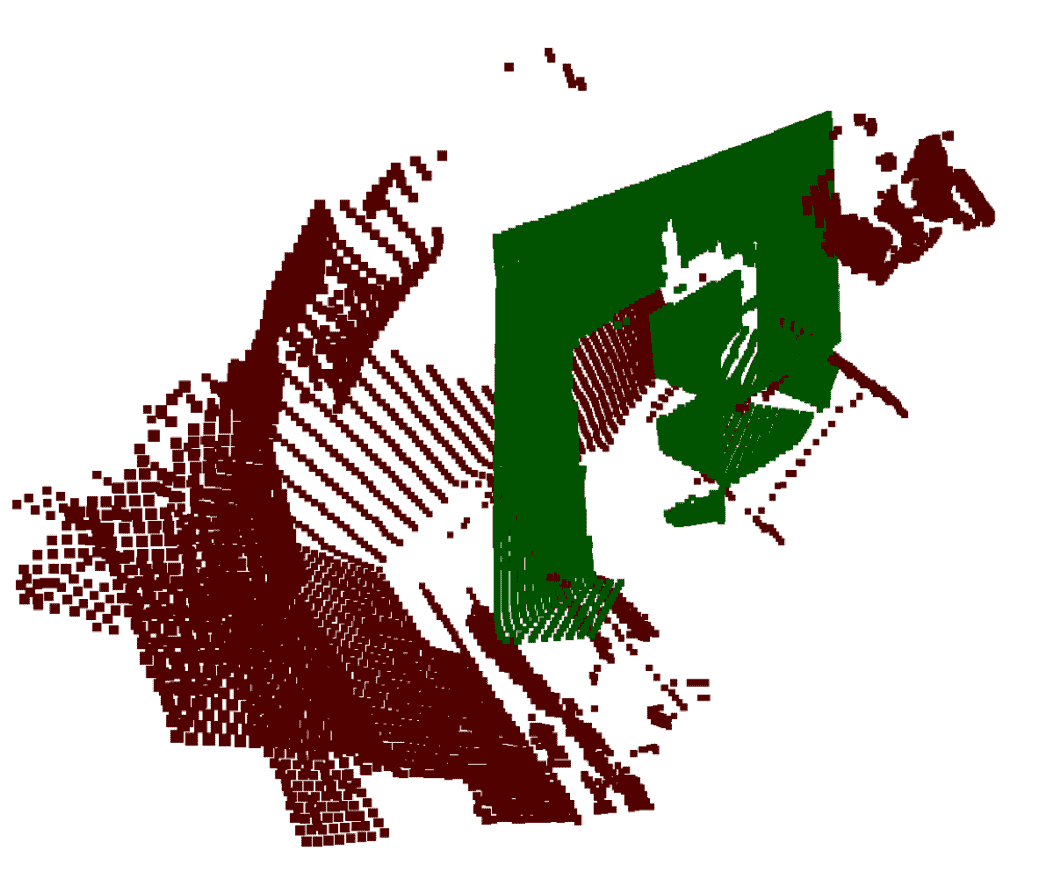}} & 
\T{\includegraphics[width=\imw]    {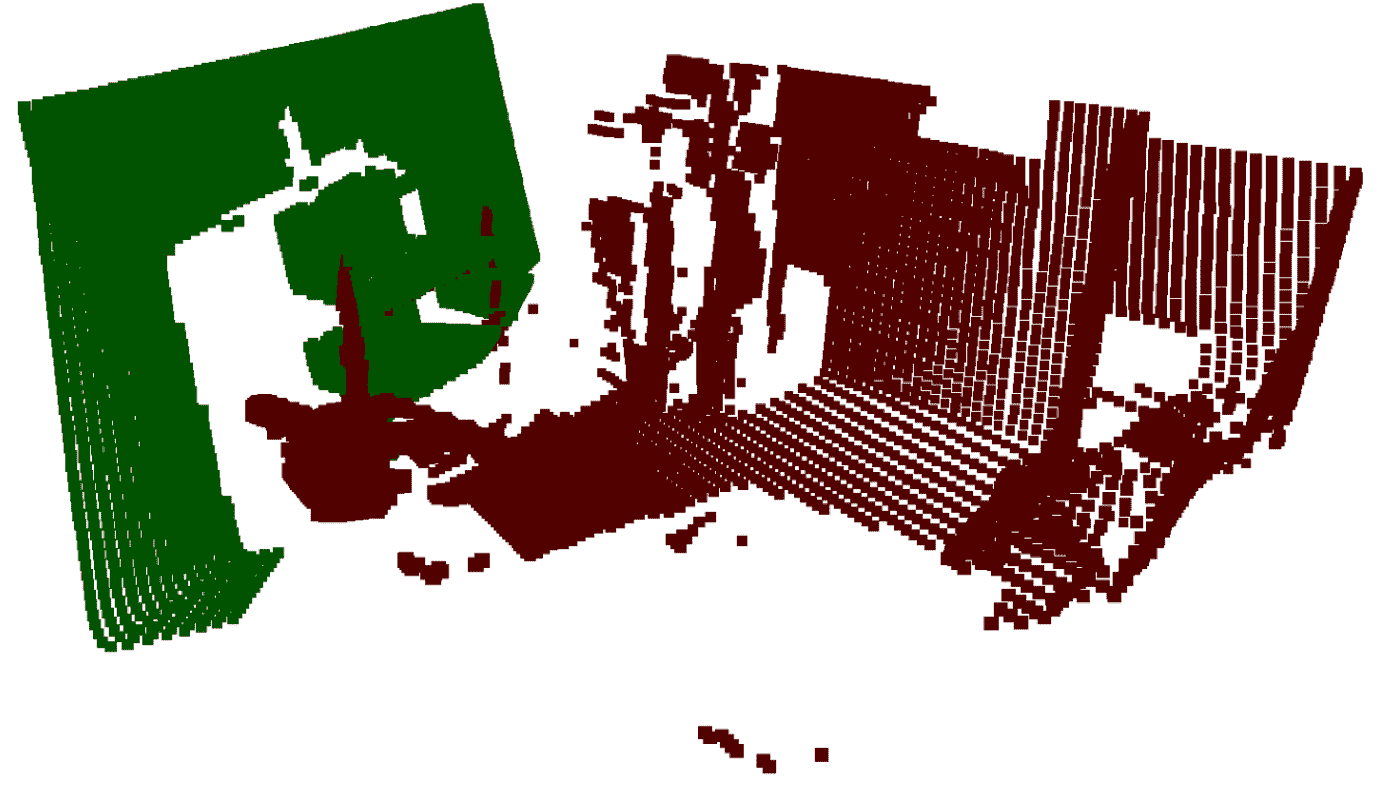}} & 
\T{\includegraphics[width=\imw]      {figures/fig5/0591_top1.png}} & 
\T{\includegraphics[width=\imw]     {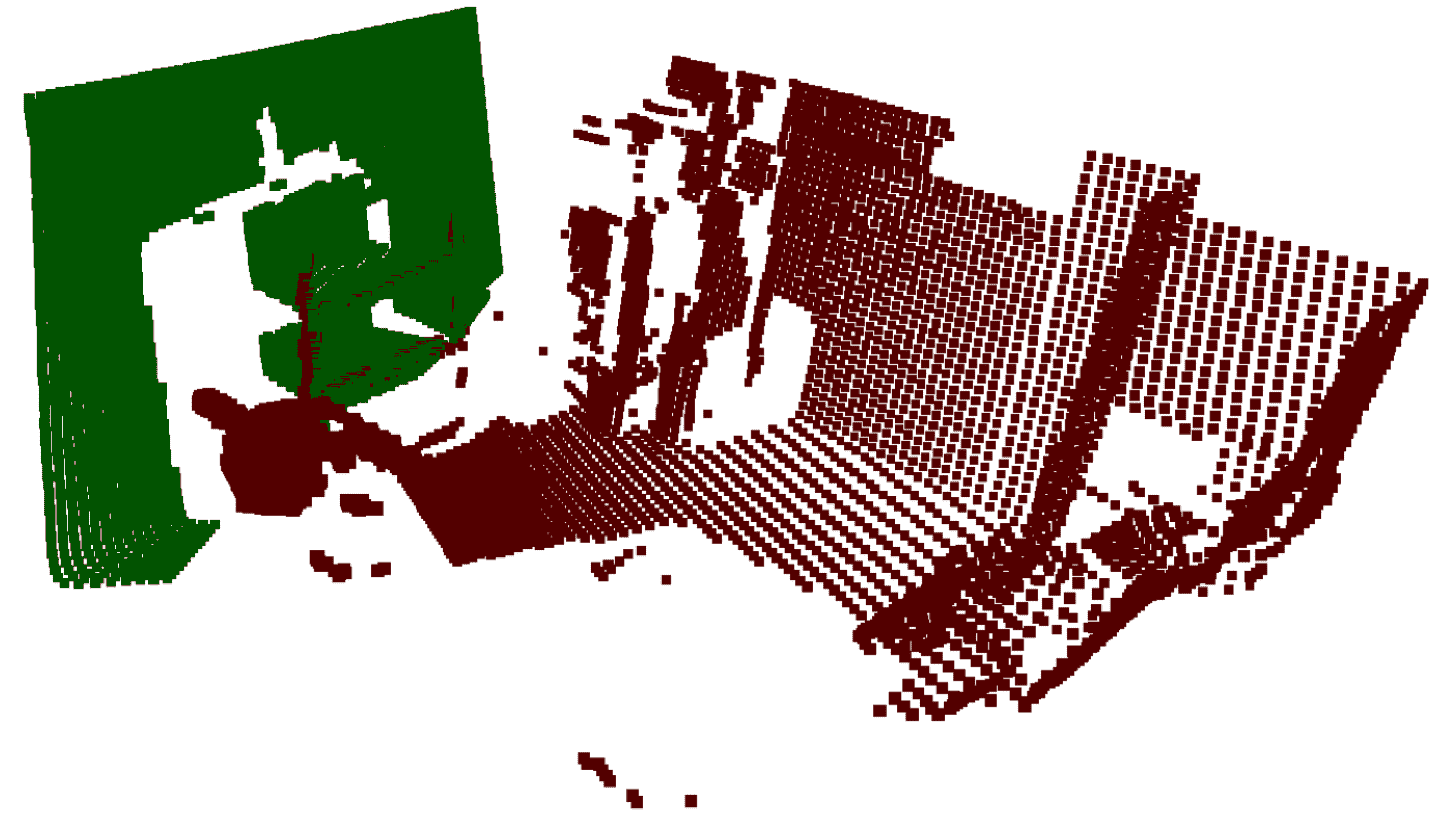}} & 
\T{\includegraphics[width=\imw]    {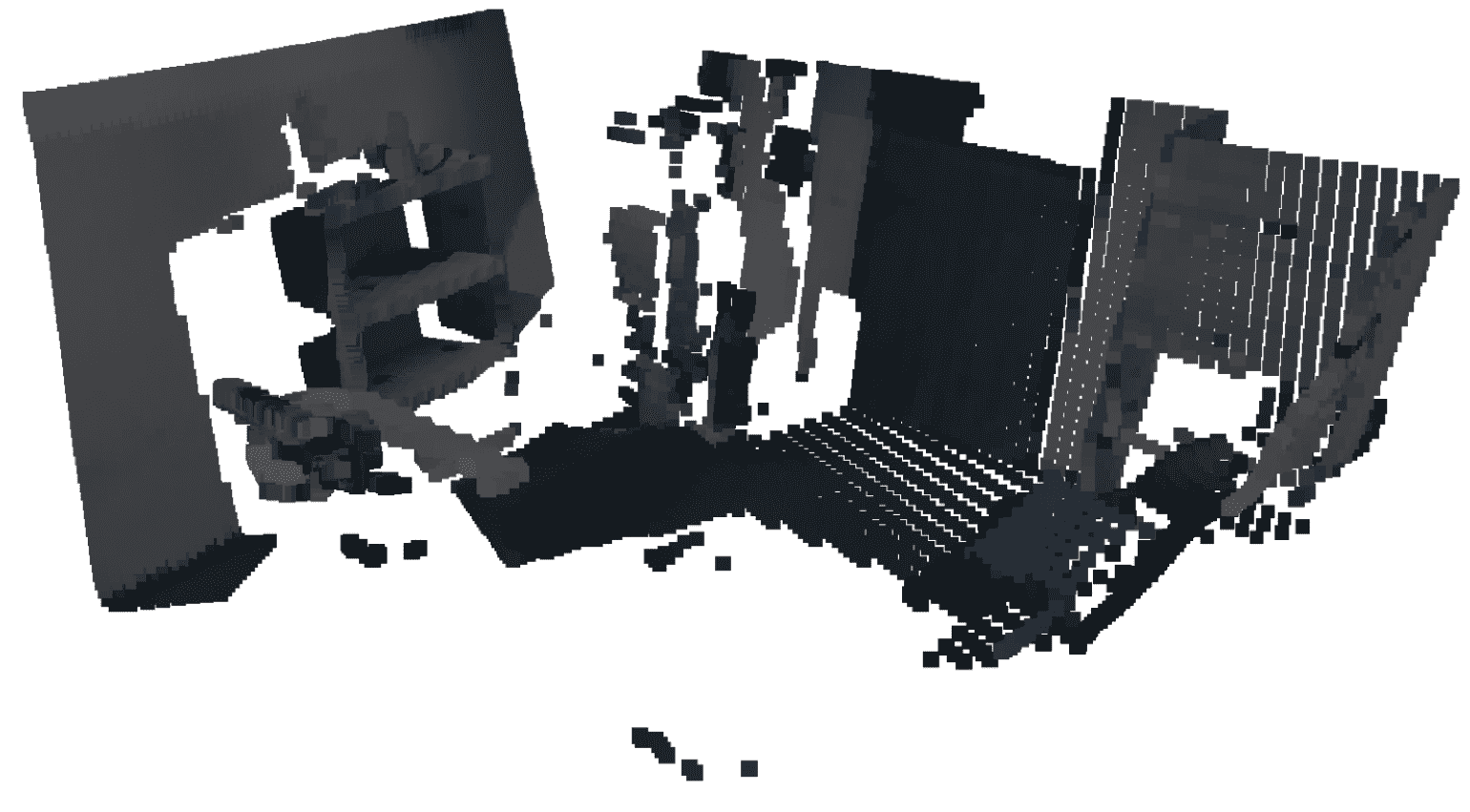}} \\
[+21pt]
 
\T{\includegraphics[width=\imw]   {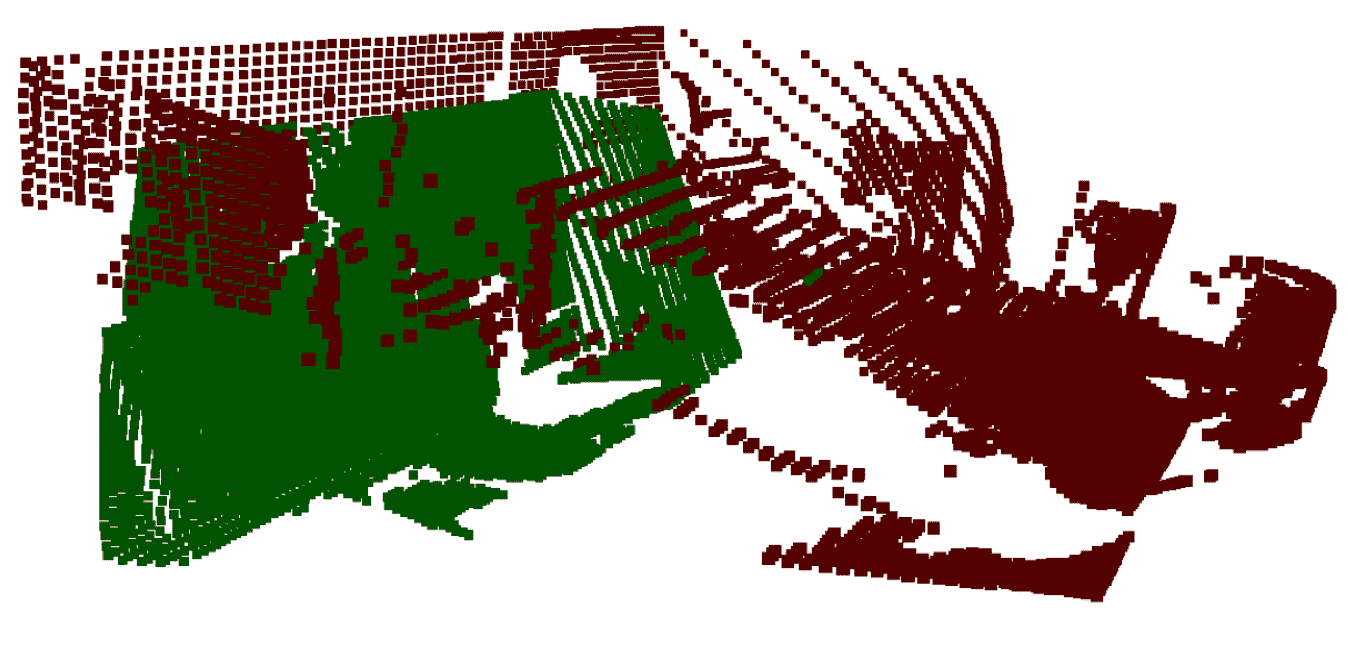}} & 
\T{\includegraphics[width=\imw]      {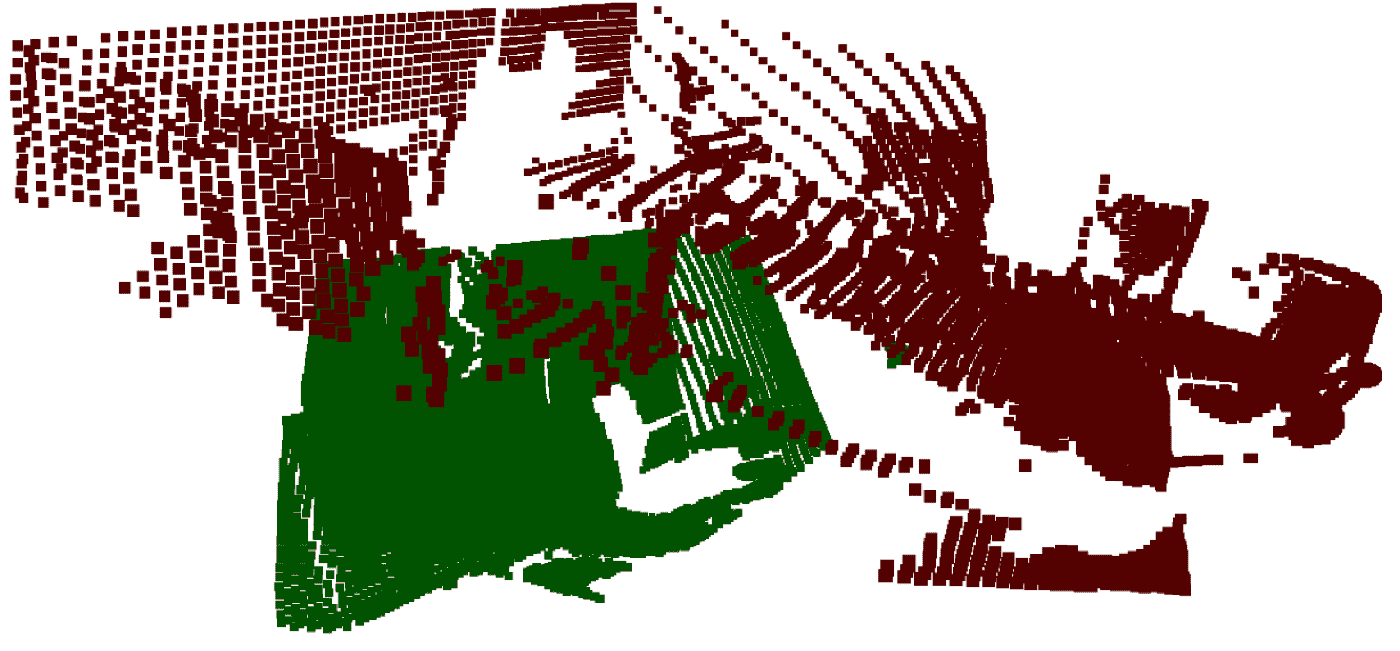}} & 
\T{\includegraphics[width=\imw]    {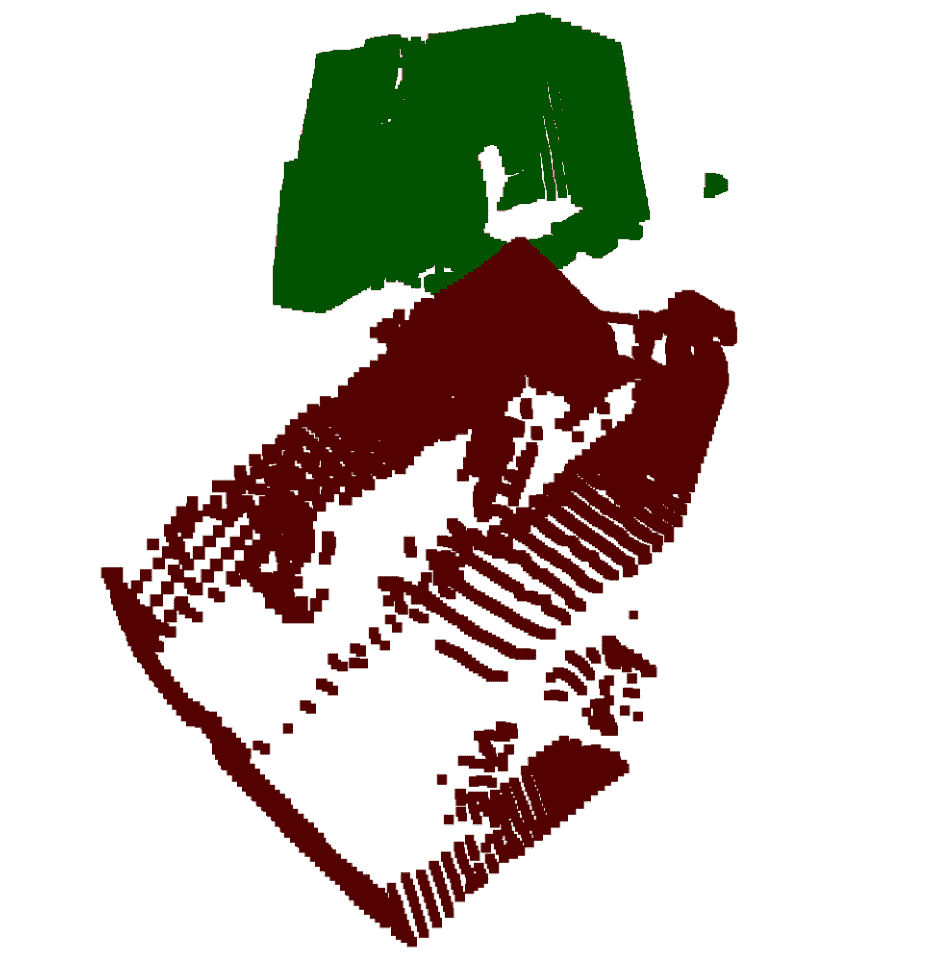}} & 
\T{\includegraphics[width=\imw]      {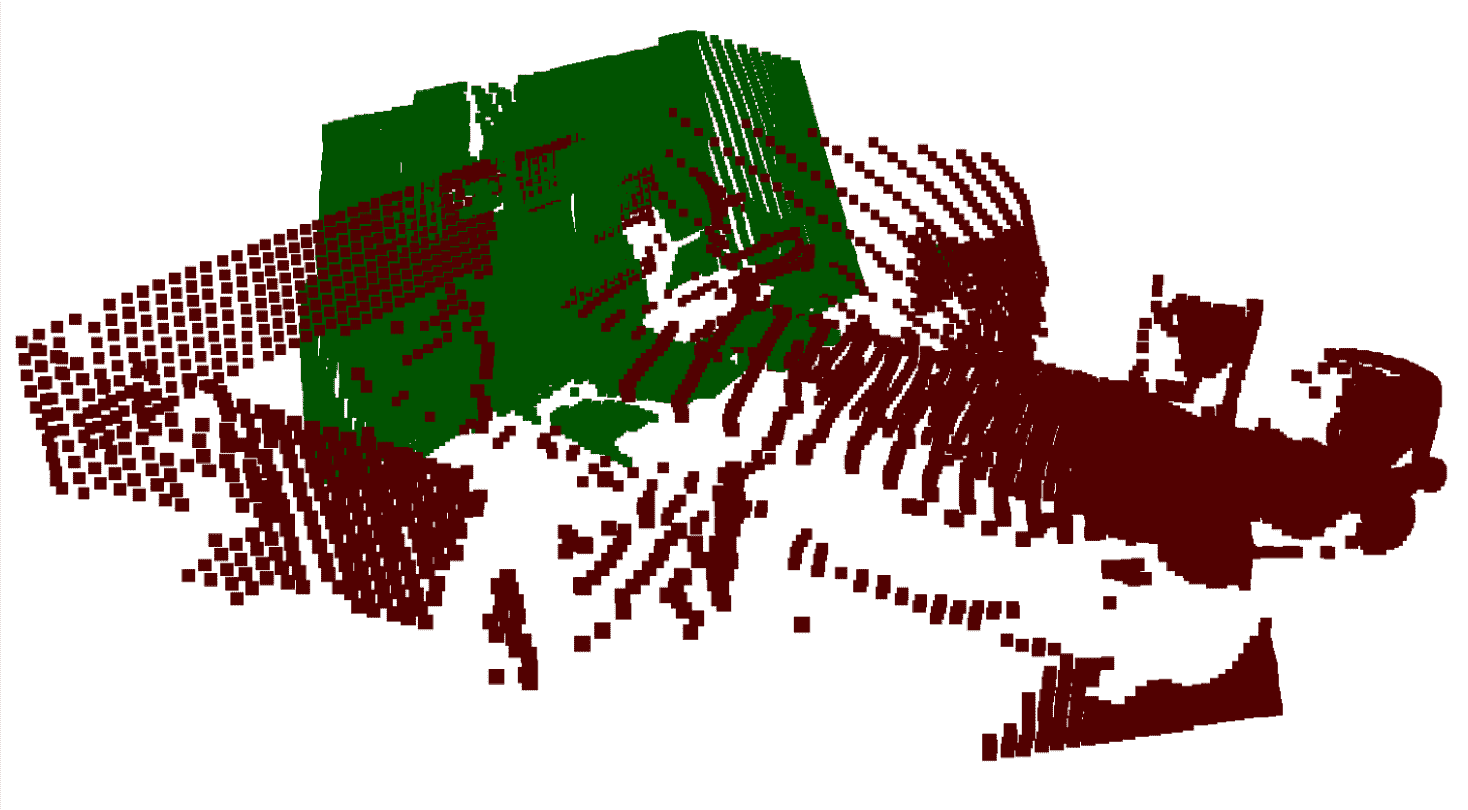}} & 
\T{\includegraphics[width=\imw]     {figures/fig5/0575_top1.png}} & 
\T{\includegraphics[width=\imw]    {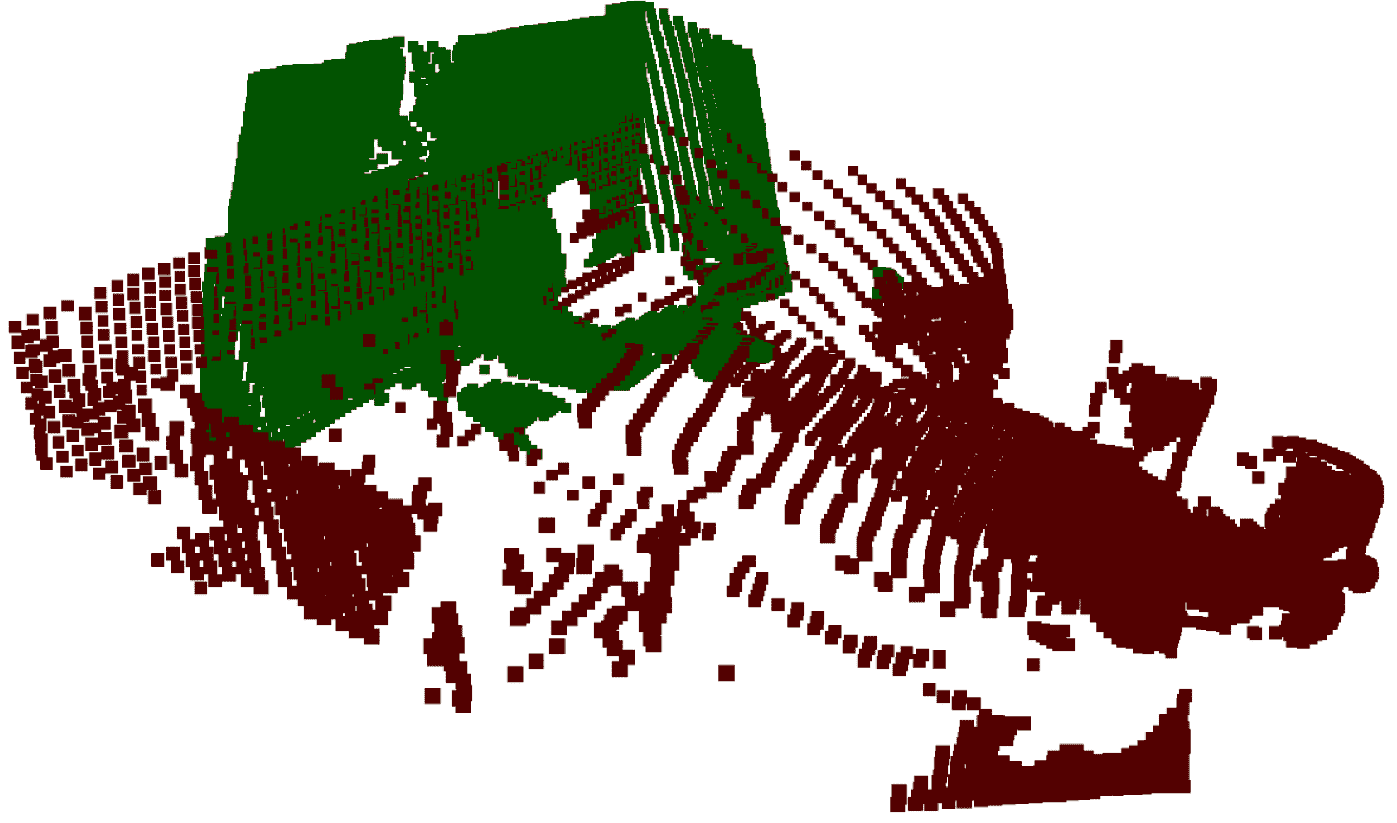}} & 
\T{\includegraphics[width=\imw]       {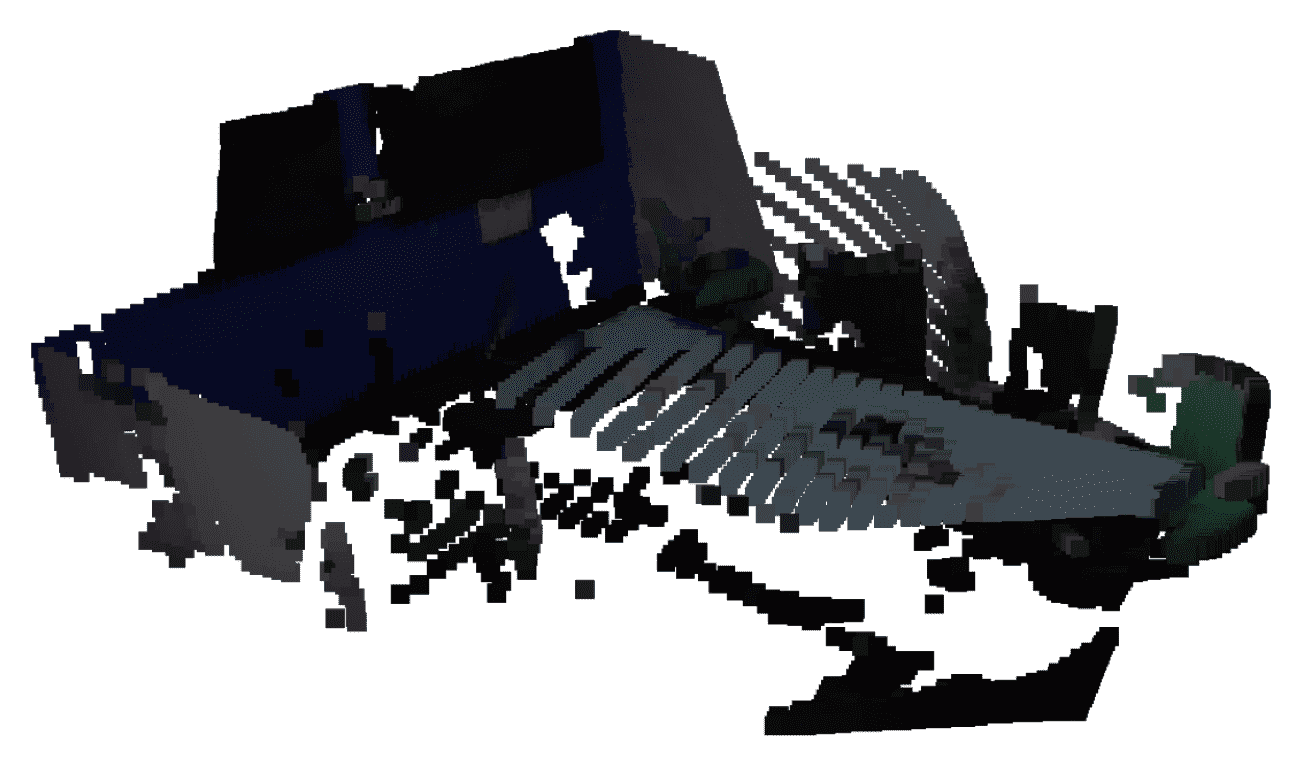}} \\
[+21pt]

\T{\includegraphics[width=\imw]   {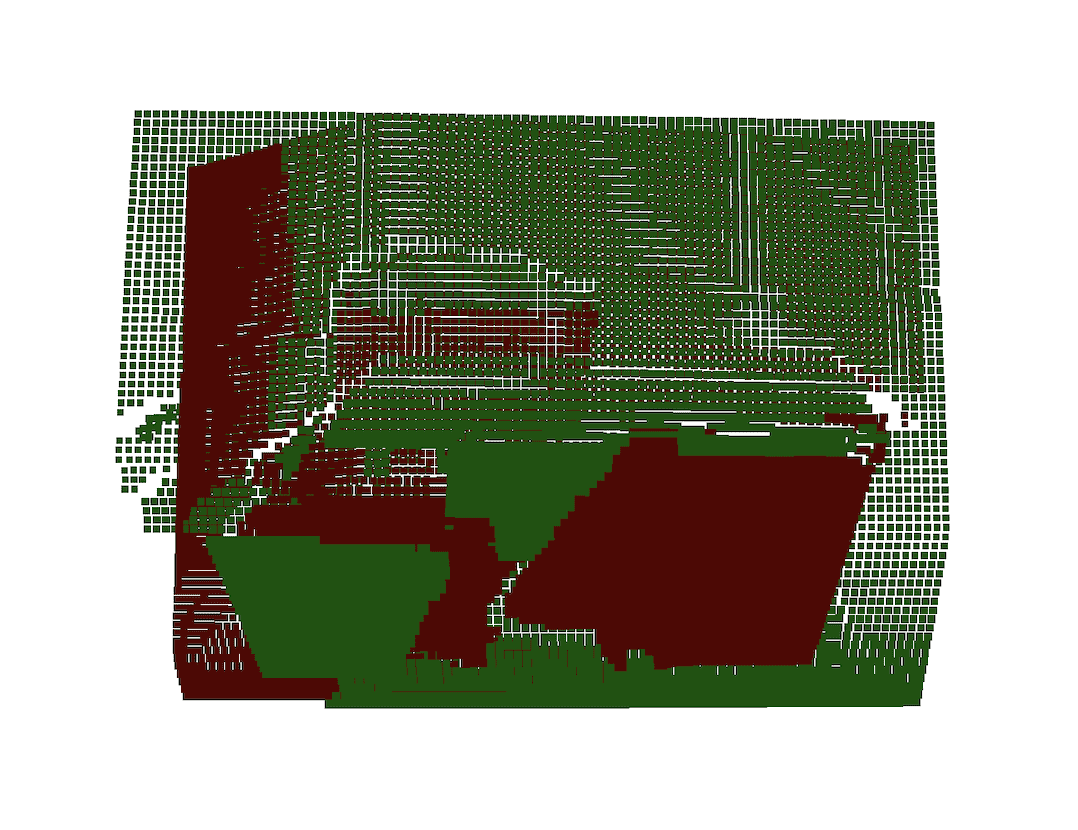}} & 
\T{\includegraphics[width=\imw]      {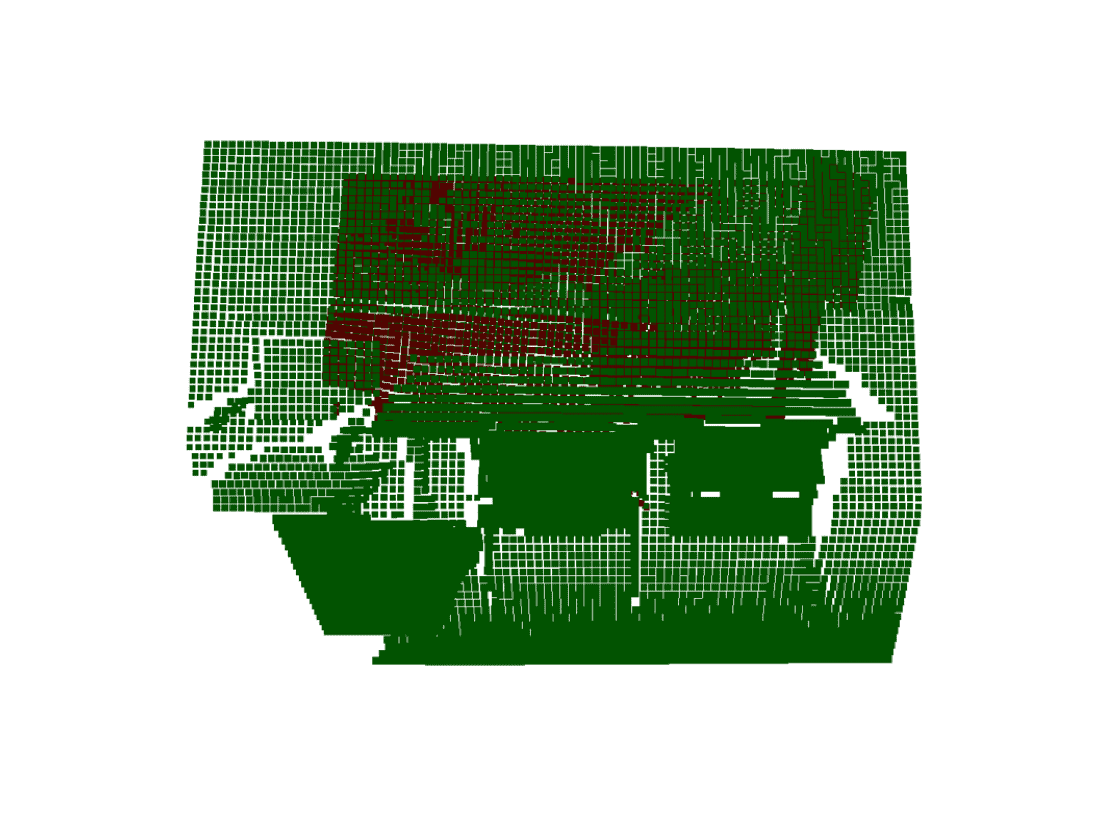}} & 
\T{\includegraphics[width=\imw]    {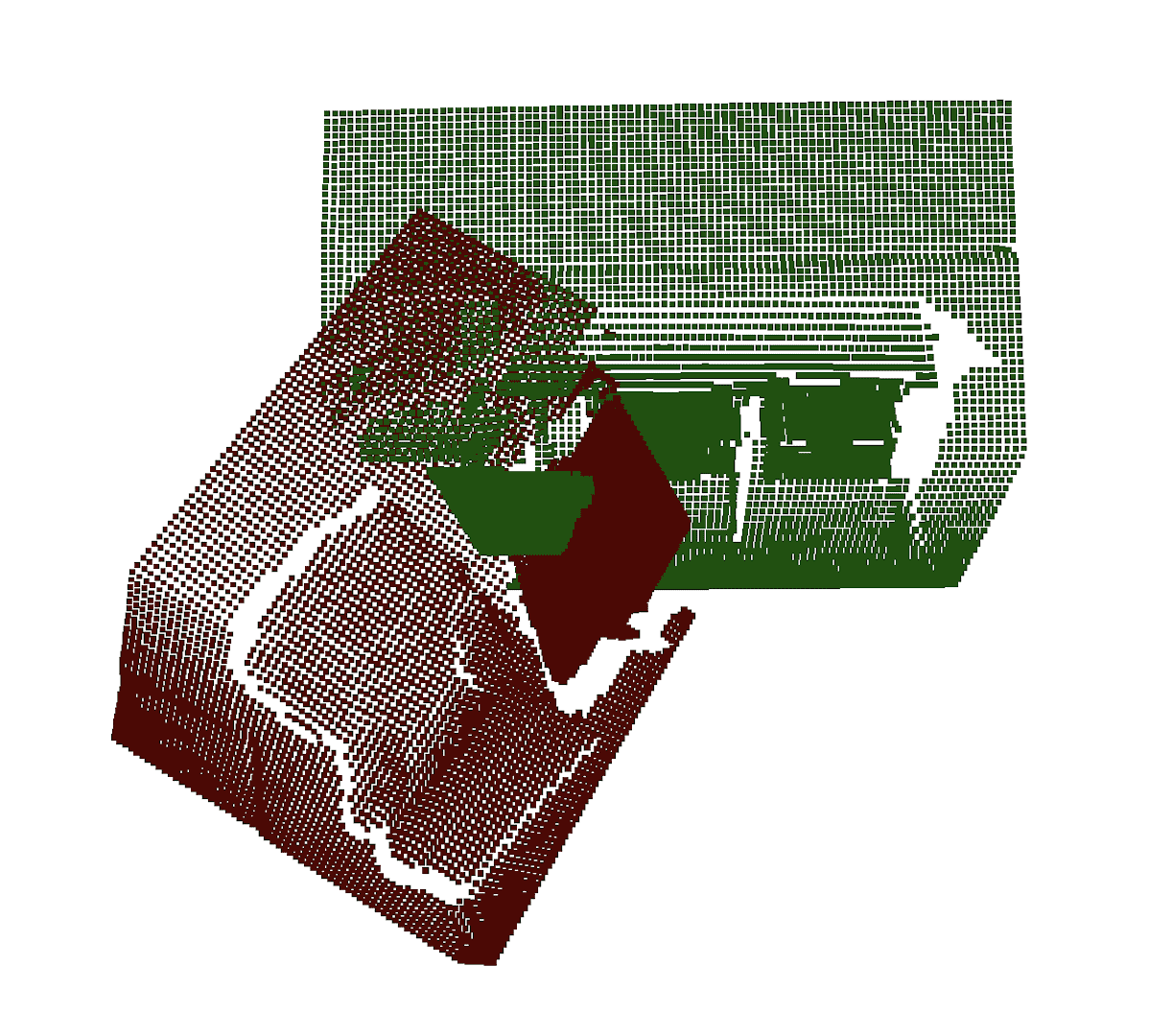}} & 
\T{\includegraphics[width=\imw]      {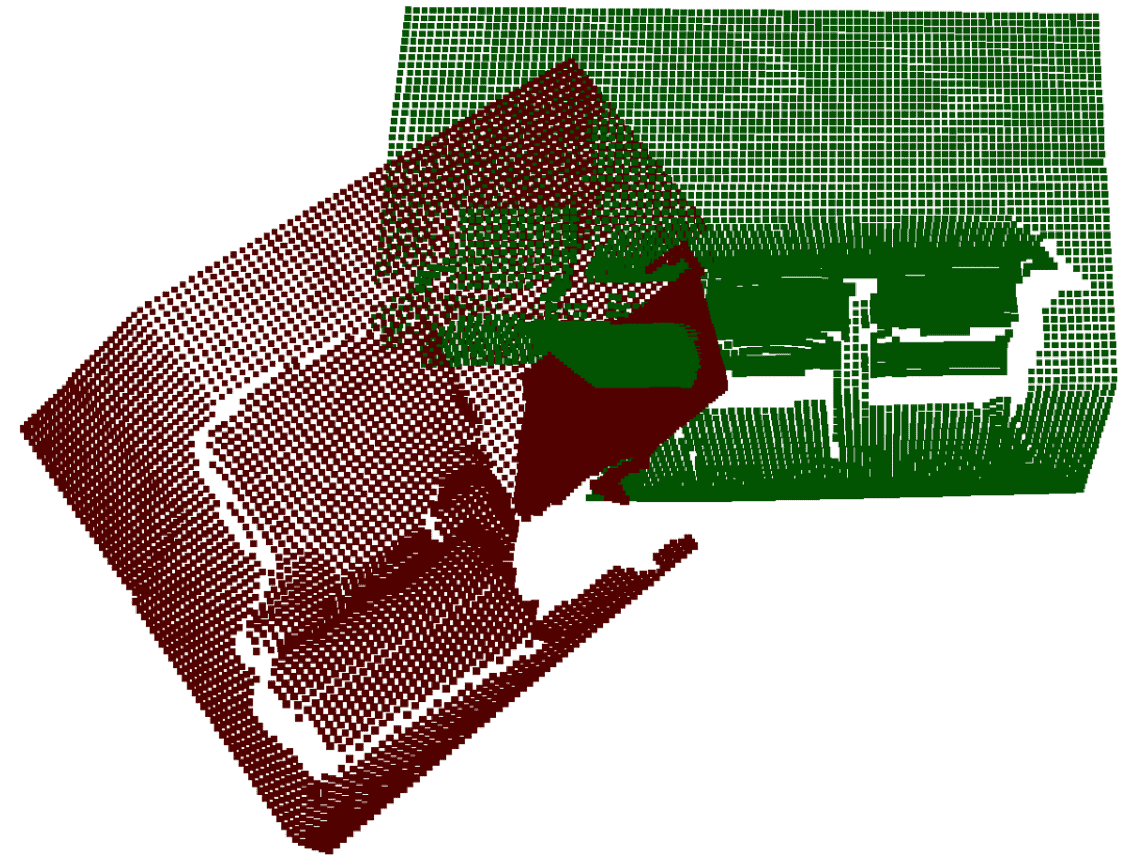}} & 
\T{\includegraphics[width=\imw]     {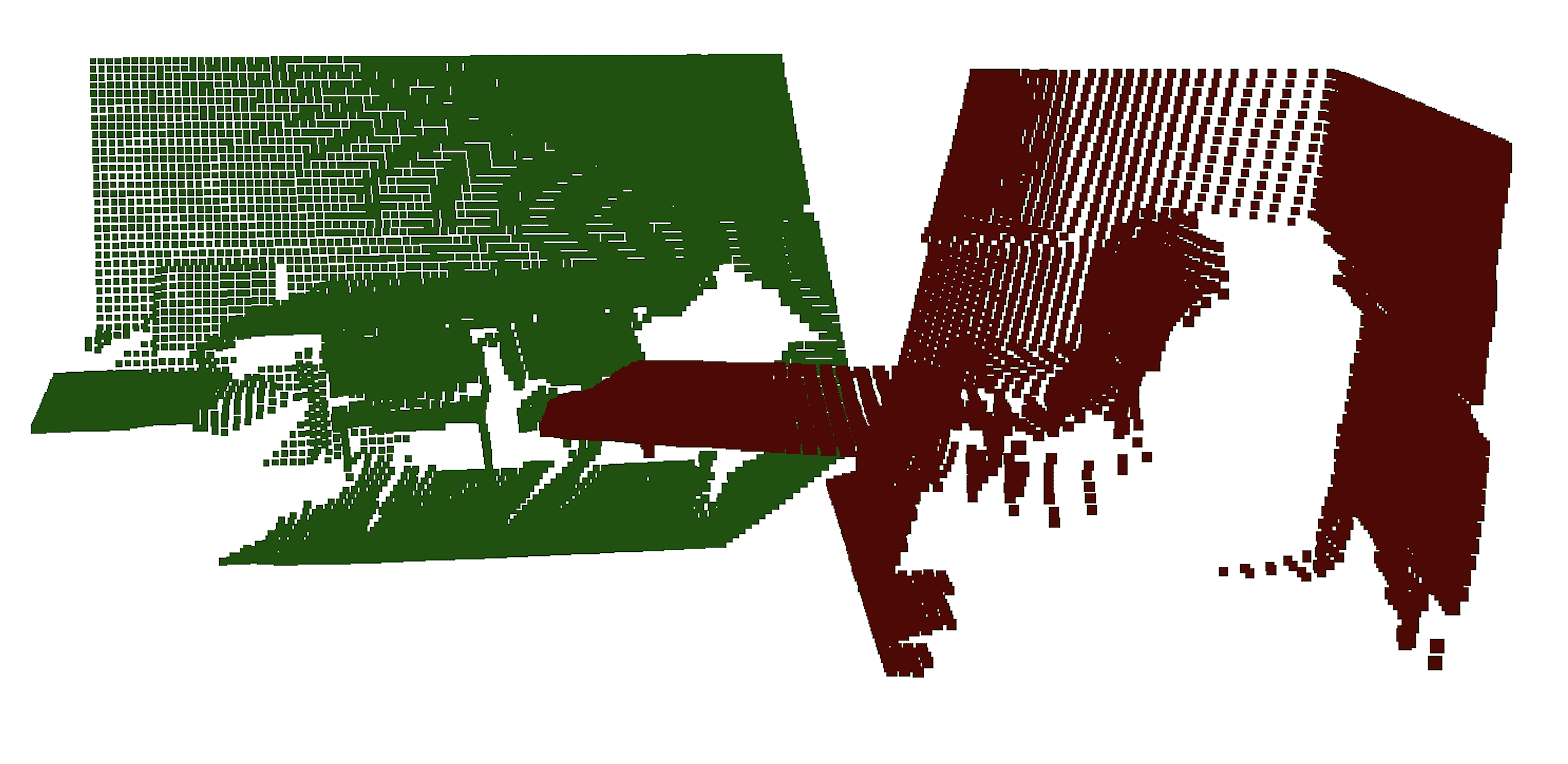}} & 
\T{\includegraphics[width=\imw]    {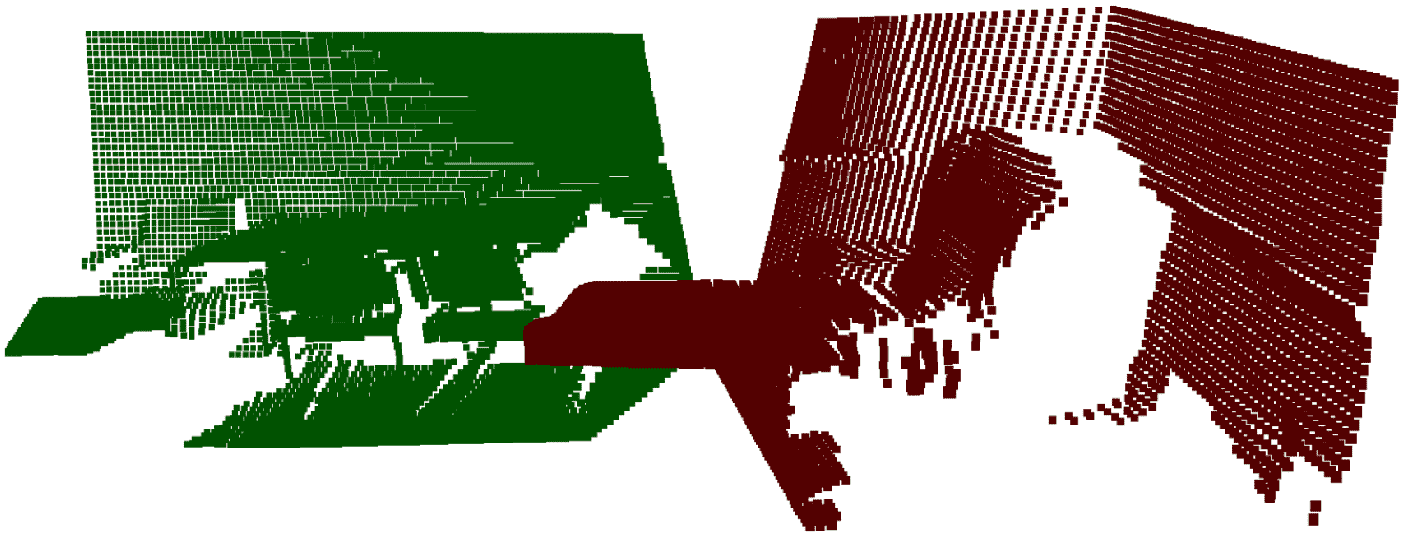}} & 
\T{\includegraphics[width=\imw]       {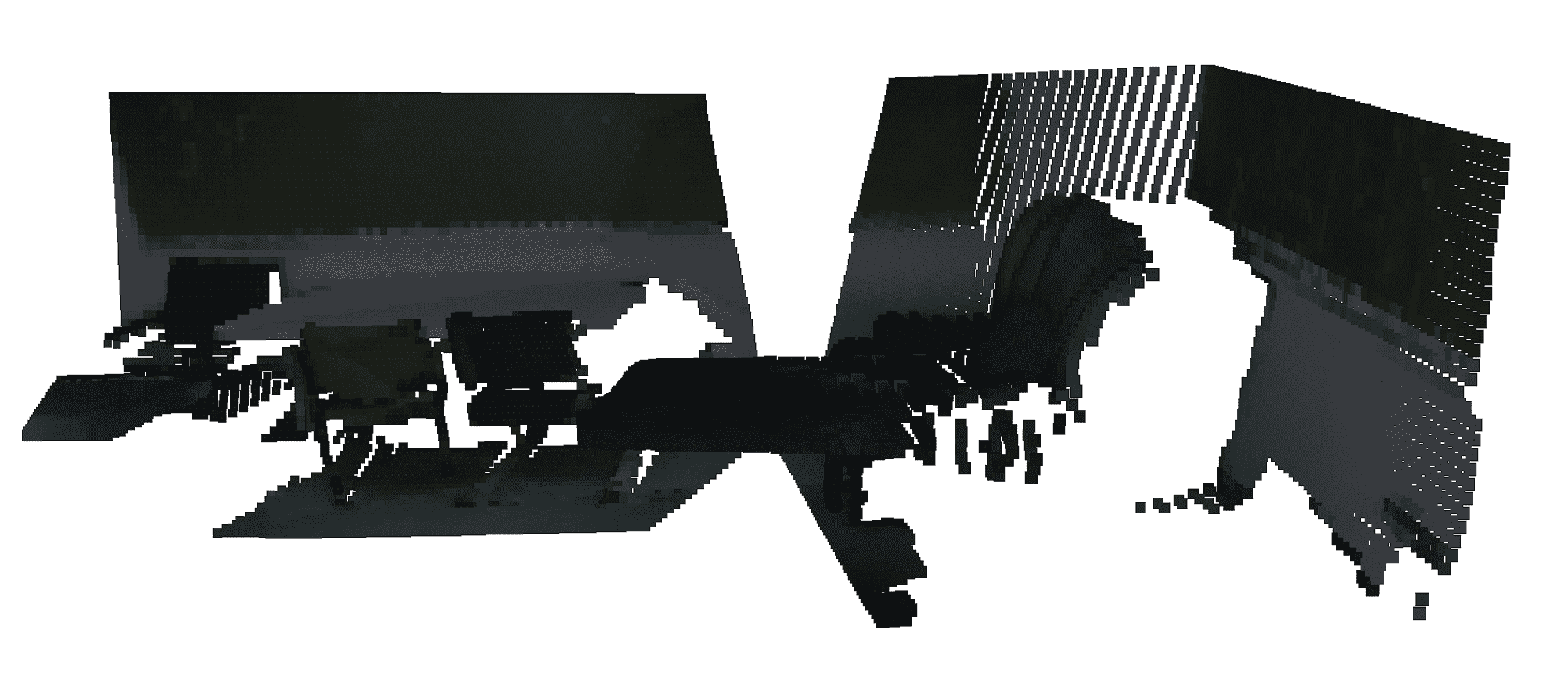}} \\
[+21pt]

\T{\includegraphics[width=\imw]   {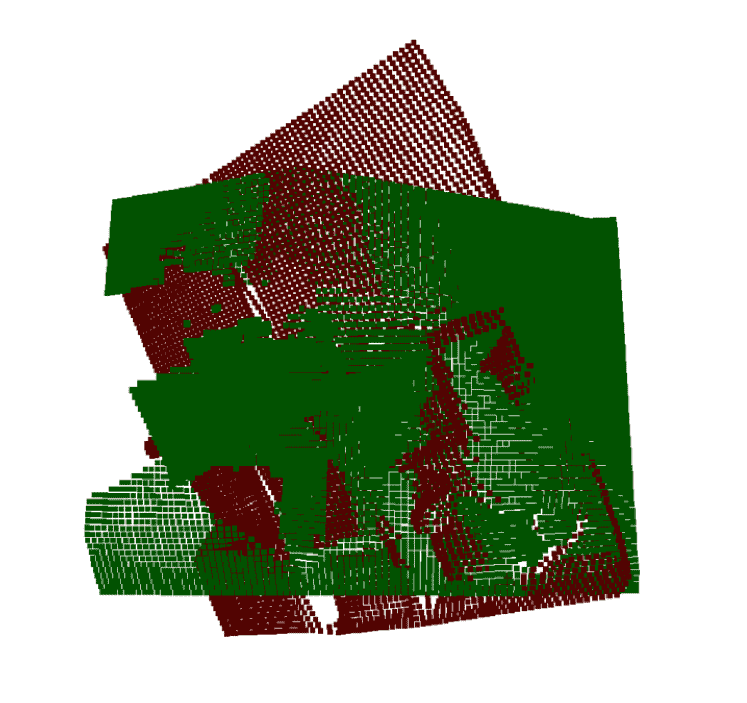}} & 
\T{\includegraphics[width=\imw]      {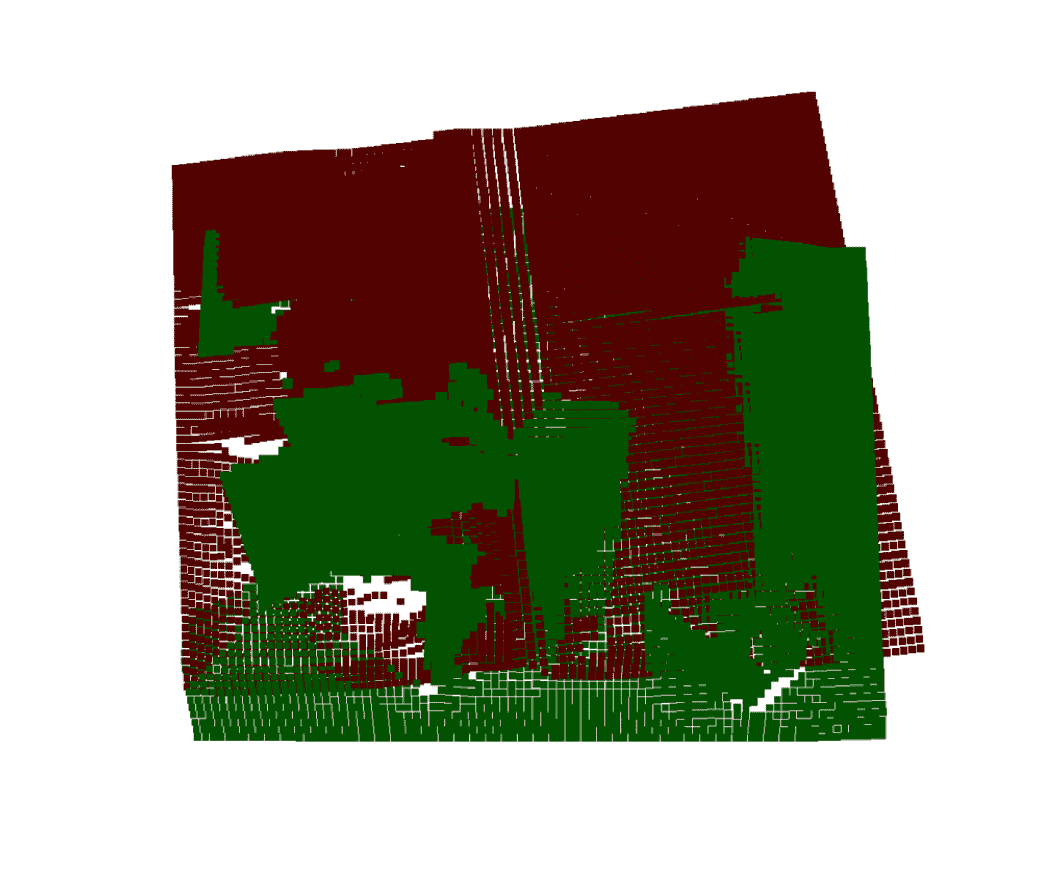}} & 
\T{\includegraphics[width=\imw]    {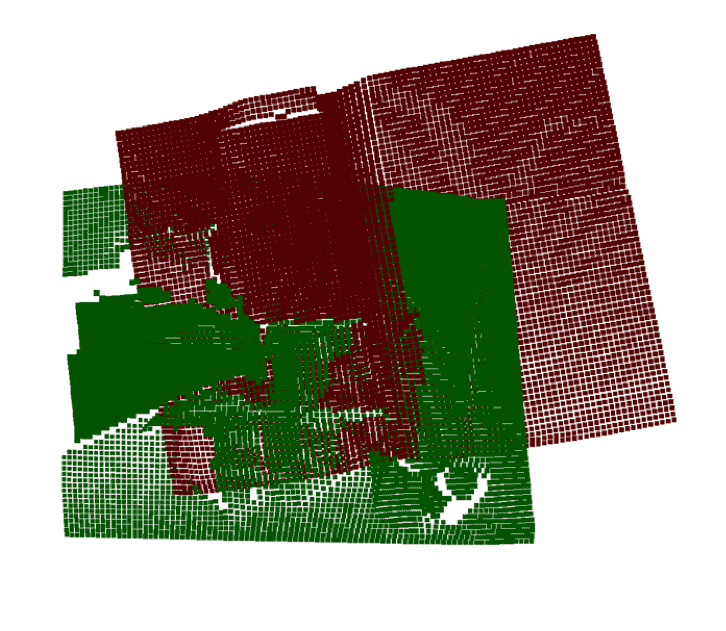}} & 
\T{\includegraphics[width=\imw]      {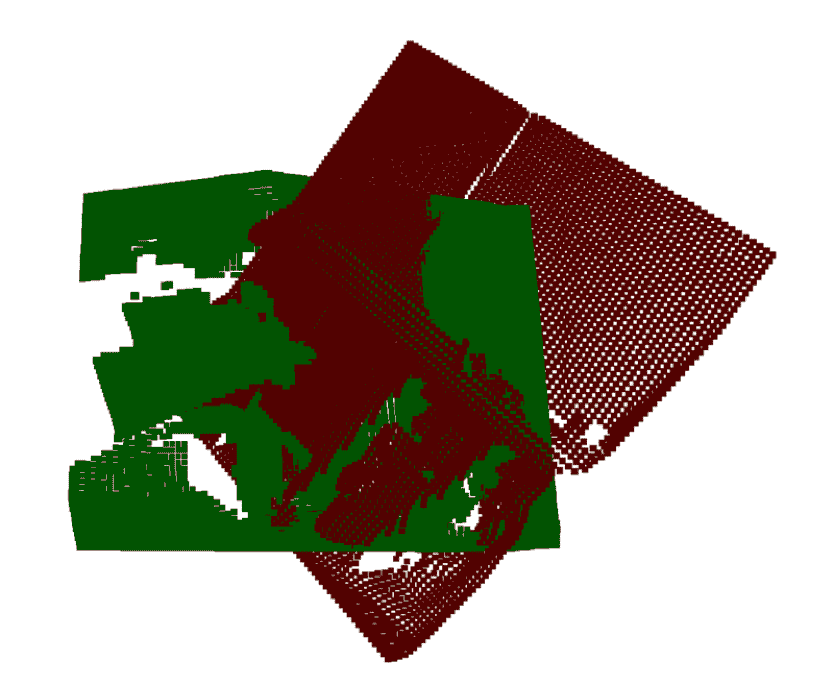}} & 
\T{\includegraphics[width=\imw]     {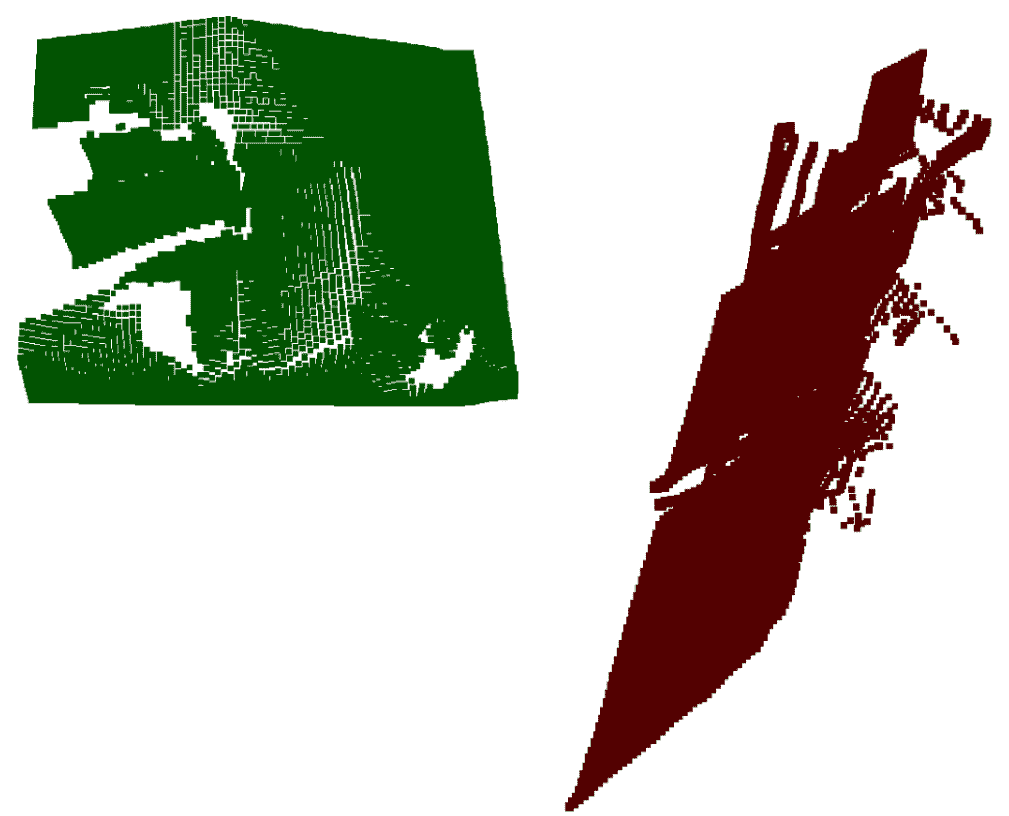}} & 
\T{\includegraphics[width=\imw]    {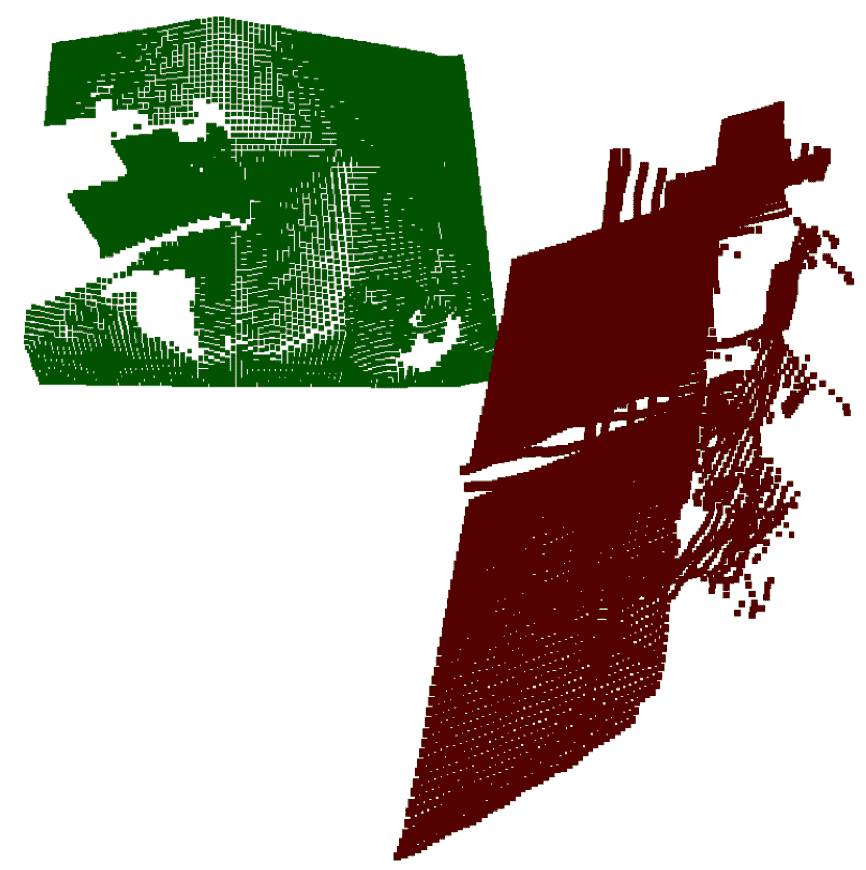}} & 
\T{\includegraphics[width=\imw]       {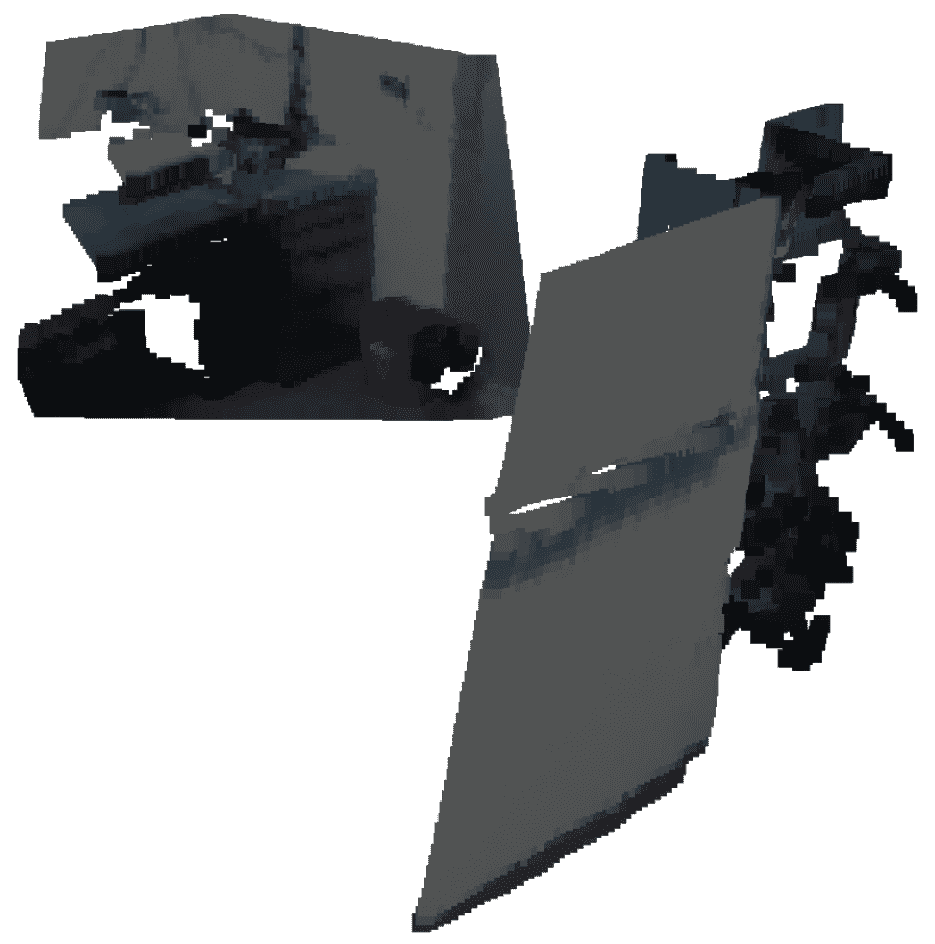}} \\
[+21pt]

\textbf{Super4PCS} & \textbf{RobustGR} & \textbf{ScanComplete} & \textbf{Ours Top1} & \textbf{Ours Top3} & \textbf{G.T. Scene} & \textbf{G.T. Color} \\
\end{tabular}

\caption{\small{Qualitative comparison between our approach and baseline approaches. From left to right, we show the results of Super4PCS~\protect\cite{CGF:CGF12446},RobustGR~\protect\cite{DBLP:conf/eccv/ZhouPK16},~ScanComplete~\protect\cite{Yang_2019_CVPR},Ours-Top1, Ours-Top3, and Ground Truth figures.}}
\label{fig:figure_baseline}
\vspace{-0.15in}
\end{figure*}

Since the local module solves a robust optimization problem, its optimal solution is insensitive to small perturbations to the outputs of the global module. Thus, instead of training the combination of the local module and the global module end-to-end, we utilize a decoupled approach in this paper. Specifically, our approach optimizes all variables $\Theta  =\{\theta_{g}, \alpha_{g}, \theta_{l}, \alpha_{l}\}$ by minimizing the following loss term:

\begin{equation}
\min\limits_{\Theta} \ l_{1}(\theta_g) + \lambda_2 l_2(\theta_l) + \lambda_3 l_3(\alpha_{g}) + \lambda_4 l_4(\alpha_{l})
\label{Eq:Total:Loss}
\end{equation}
where $l_1$, $l_2$, $l_3$, and $l_4$ train scan completion networks, the geometric relation network,  the spectral matching sub-module, and the local module, respectively. $\lambda_i$ are hyper-parameters determined via 10-fold cross validation.

The scan completion loss $l_1(\theta_g)$ combines the losses for training the scan completion network under each data representation. The geometric relation loss $l_2(\theta_l)$ utilizes initial relative poses that randomly perturbed from the underlying ground-truth. Since the formulations of these two losses are standard for network training, we leave the details to the supp. material. In the following, we focus on the loss terms for spectral matching sub-module and the local module.

\noindent\textbf{Loss term for spectral matching.} Consider the consistency matrix $C_{p}$ for a scan pair $p = (S_1, S_2)$. Our goal for training the spectral matching sub-module is to enforce that the normalized indicator vector $\bs{u}_{p}^{gt}$ of ground-truth correspondences is a top eigenvector of $C_{p}$. Note that $C_{p}$ is dependent on $\alpha_g$, We omit them to make the notations uncluttered.

To avoid training a recurrent neural network converted from power iteration for eigen-vector computation (c.f.~\cite{Pascanu:2013:DTR}), our approach utilizes three properties of a top eigen-vector, i.e., a normalized vector $\bs{u}$ is a top eigen-vector of a matrix $C$ if (1) the corresponding eigenvalue $\lambda = \bs{u}^T C\bs{u}$, (2) $C\bs{u} = \lambda \bs{u}$, and (3) $\lambda$ is maximized. Given a collection of scan pairs $\set{P} = \{p\}$ as, we define $l_3(\alpha_g):=$
\begin{align*}
\sum\limits_{p\in \set{P}}& \big(\|C_{p}\bs{u}_{p}^{gt}-({\bs{u}_{p}^{gt}}^TC_{p}\bs{u}_{p}^{gt})\bs{u}_{p}^{gt}\|^2 - {\bs{u}_{p}^{gt}}^TC_{p}\bs{u}_{p}^{gt}\big).
\end{align*}

\noindent\textbf{Loss term for the local module.} Define $\set{I}_{p}(R,\bs{t})\in \R^{3\times 4}$ as one iteration of the IRNLS procedure from the current pose $(R,\bs{t})$, i.e., optimization and reweighting. Similar to $l_{3}$, we avoid training recurrent networks of the robust regression procedure by directly penalizing the difference between the ground-truth relative pose $R_p^{gt},\bs{t}_p^{gt}$ and $\set{I}_{p}(R^{gt},\bs{t}^{gt})$:
\begin{equation}
l_4(\alpha_l) = \sum\limits_{p \in \set{P}}\ \|(R_p^{gt},\bs{t}_p^{gt}) - \set{I}_{p}(R^{gt},\bs{t}^{gt})\|_{\set{F}}^2
\label{Eq:l:4}
\end{equation}
where $\|\cdot\|_{\set{F}}$ is the matrix Frobenius-norm. One intuition of (\ref{Eq:l:4}) is that when $(R_p^{gt},\bs{t}_p^{gt}) = \set{I}_{p}(R^{gt},\bs{t}^{gt})$, then $(R_p^{gt},\bs{t}_p^{gt})$ becomes the stationary point of the IRNLS procedure.

Minimizing $l_4(\alpha_l)$ requires computing the derivatives of $\set{I}_{p}$. Since $\set{I}_p$ is based on a local optimum of (\ref{Eq:Total:Loss1}), we use the implicit function theorem~\cite{IFTBook} to back-propagate the gradients from the optimized poses to the predicted features and geometric relations. Again, we defer the details to the supp. material for the interest of space.

\noindent\textbf{Training procedure.} We train (\ref{Eq:Total:Loss}) through three phases. The first phase trains each prediction network in isolation. The second phase fixes the prediction networks and trains the hyper-parameters of each module in isolation. The final phase fine-tunes all the variables together by solving (\ref{Eq:Total:Loss}). Network training involves the ADAM optimizer~\cite{KingmaB14}, and please refer to the supp. material for details. 

\section{Experimental Evaluation}
\label{Section:Experimental:Evaluation}

This section presents an experimental evaluation of the proposed approach. Section~\ref{Subsection:Experimental:Setup} describes the experimental setup. Section~\ref{Subsection:Analysis:of:Results} analyzes the experimental results and compares the proposed approach with baseline approaches. Section~\ref{Subsection:Global:Network} and Section~\ref{Subsection:Local:Network} analyze the global module and the local module, respectively. 

\subsection{Experimental Setup}
\label{Subsection:Experimental:Setup}

\noindent\textbf{Datasets.} We perform experimental evaluation using three datasets:\textsl{SUNCG}~\cite{song2016ssc}, \textsl{Matterport}~\cite{DBLP:journals/corr/abs-1709-06158}, and \textsl{ScanNet}~\cite{DBLP:journals/corr/DaiCSHFN17}, where SUNCG provides synthetic data, Matterport and ScanNet provide real data. For each room in each dataset, we randomly sample 25 camera locations. For real data sets such as ScanNet, the camera locations are sampled from the recording sequences. For each dataset, we collect around 50k scan pairs for training, and 1k scan pairs for testing. Note that the scenes of the training and testing scans do not overlap.

\setlength\tabcolsep{3.6pt}
\begin{table*}
  \footnotesize
  \centering
  \begin{tabular}{l|c|c|c|c|c|c|c|c|c|c|c|c|c|c|c|c|c|c}
  \toprule
  & \multicolumn{6}{c|}{SUNCG} & \multicolumn{6}{c|}{Matterport} &  \multicolumn{6}{c}{ScanNet}\\
  \hline
  & \multicolumn{3}{|c|}{Rotation} & \multicolumn{3}{|c|}{Trans} & \multicolumn{3}{|c|}{Rotation} & \multicolumn{3}{|c|}{Trans} &  \multicolumn{3}{|c|}{Rotation} & \multicolumn{3}{|c}{Trans}\\
  \hline
  & top1&top3&top5 & top1&top3&top5 & top1&top3&top5 & top1&top3&top5 & top1&top3&top5 & top1&top3&top5\\
  \hline
  360-image ($\geq 10\%$) & 9.8 & 4.5 & 3.5 & 0.17
  & 0.16 & 0.16 & 19.9 & 10.4 & 8.5 & 0.39 & 0.38 & 0.38 & 18.4 & 11.0 & 8.7 & 0.71 & 0.54 & 0.50\\
  Plane ($\geq 10\%$) & 100.3 & 65.9 & 45.8 & 0.58 & 0.70 & 0.59 & 123.6 & 92.4 & 78.6 & 0.79 & 0.99 & 0.99 & 123.6 & 91.4 & 79.1 & 3.03 & 2.63 & 2.65\\
  2D-layout ($\geq 10\%$)& 52.0 & 23.1 & 16.4 & 0.85 & 0.50 & 0.52 & 73.1 & 42.2 & 33.8 & 1.57 & 1.38 & 1.36 & 45.5 & 25.2 & 18.3 & 1.31 & 1.07 & 1.01\\
  Ours ($\geq 10\%$) & \textbf{8.0} & \textbf{3.6} & \textbf{3.3} & \textbf{0.16} & \textbf{0.14} & \textbf{0.14} & \textbf{13.9} & \textbf{9.7} & \textbf{7.8} & \textbf{0.32} & \textbf{0.31} & \textbf{0.31}  & \textbf{16.7} & \textbf{10.5} & \textbf{8.7}  & \textbf{0.69} & \textbf{0.53} & \textbf{0.50}\\
  \hline
  360-image ($\leq 10\%$)& 79.7 & 30.4 & 19.7 & 0.49 & 0.44 & 0.40 & 70.0 & 42.6 & 34.1  & 0.88 & 0.83 & 0.80 & 66.4 & 46.9 & 36.1  & 1.37 & 1.34 & 1.35\\
  Plane($\leq 10\%$)  & 123.4 & 80.4 & 57.8 & 0.65 & 0.86 & 0.72 & 135.8 & 77.8 & 68.9 & 0.71 & 0.96 & 0.74  & 119.5 & 86.4 & 81.5 & 2.56 & 2.60 & 2.67\\
  2D-layout($\leq 10\%$)  & 99.9 & 55.6 & 41.4 & 1.18 & 0.76 & 0.82 & 107.0 & 63.8 & 52.5 & 1.42 & 1.69 & 1.74 & 73.7 & 46.3 & 34.2 & 1.94 & 1.90 & 1.89\\
  Ours ($\leq 10\%$) & \textbf{62.8} & \textbf{25.6} & \textbf{18.1} & \textbf{0.40} & \textbf{0.38} & \textbf{0.38} & \textbf{58.9} & \textbf{39.9} & \textbf{32.3}  & \textbf{0.69} & \textbf{0.68} & \textbf{0.68} & \textbf{59.3} & \textbf{40.4} & \textbf{33.5}  & \textbf{1.36} & \textbf{1.33} & \textbf{1.33}\\
  \hline
  360-image (all)& 31.1 & 12.4 & 8.5 & 0.27 & 0.24 & 0.23 & 34.2 & 19.0 & 15.3 & 0.53 & 0.50 & 0.49 & 34.1 & 22.7 & 17.7 & 0.93 & 0.79 & 0.78 \\
  Plane(all)  & 107.3 & 70.3 & 49.4 & 0.60 & 0.75 & 0.63 & 126.4 & 89.1 & 76.4 & 0.78 & 0.99 & 0.95 & 122.4 & 89.9 & 79.8 & 2.89 & 2.60 & 2.65\\
  2D-layout(all)  & 66.6 & 33.0 & 24.0 & 0.95 & 0.58 & 0.61 & 82.1 & 47.9 & 38.8 & 1.53 & 1.46 & 1.46 & 56.3 & 32.1 & 23.5 & 1.50 & 1.33 & 1.30\\
  Ours (all) & \textbf{24.7} & \textbf{10.3} & \textbf{7.8} & \textbf{0.23} & \textbf{0.21} & \textbf{0.21} & \textbf{26.0} & \textbf{17.8} & \textbf{14.3}  & \textbf{0.41} & \textbf{0.41} & \textbf{0.41} & \textbf{30.7} & \textbf{20.3} & \textbf{16.8}  & \textbf{0.91} & \textbf{0.76} & \textbf{0.76}\\
  \bottomrule
  \end{tabular}
{%
 \caption{Ablation study on global module. We show the comparison between our hybrid representation module and the single representation module. The $10\%$ refers to the overlap ratio. For each module , we also show the top1/top3/top5 error for rotation and translation.}
 \label{tab:global_ablation}
}
\vspace{-0.1in}
\end{table*}

\setlength\tabcolsep{1.5pt}
\begin{table}
  \footnotesize
  \centering
  \begin{tabular}{l|c|c|c|c|c|c}
  \toprule
  & \multicolumn{2}{c|}{SUNCG} & \multicolumn{2}{c|}{Matterport} & \multicolumn{2}{c}{ScanNet}\\
  \hline
  & Rotation & Trans & Rotation & Trans & Rotation & Trans\\
  \hline
  4PCS-Overlap ($\geq 10\%$)& 45.2 & 0.22 & 49.7 & 0.53 & 31.3 & 0.94 \\
  RobustGR ($\geq 10\%$)) & 64.9 & 0.57 & 38.1 & 0.69 & 45.4 & 1.20 \\
  ScanComp. ($\geq 10\%$) & 9.8 & 0.17 & 19.9 & 0.39 & 18.4 & 0.71\\
  Ours-Local ($\geq 10\%$) & 36.0 & 0.22 & 40.8 & 0.49 & 25.4 & 0.93\\
  Ours-Global ($\geq 10\%$) & 8.0 & 0.16 & 13.9 & 0.31 & 16.7 & 0.69\\
  Ours ($\geq 10\%$) & \textbf{5.0} & \textbf{0.11} & \textbf{10.7} & \textbf{0.29} & \textbf{14.1} & \textbf{0.67}\\
  \hline
  ScanComp.($\leq 10\%$) & 79.7 & 0.49 & 70.0 & 0.88 & 66.4 & 1.37\\
  Ours-Local ($\leq 10\%$) & 105.0 & \textbf{0.31} & 115.6 & \textbf{0.49} & 67.2 & \textbf{1.34}\\
  Ours-Global ($\leq 10\%$) & 62.8 & 0.40 & 59.3 & 0.69 & 59.3 & 1.36\\
  Ours ($\leq 10\%$) & \textbf{60.9} &  0.41 & \textbf{54.1} & 0.68 & \textbf{58.4} & 1.36\\  
  \hline
  ScanComp.(all) & 31.1 & 0.27 & 34.2 & 0.53 & 34.1 & 0.93\\
  Ours-Local (all) & 56.9 & 0.24 & 60.7 & 0.49 & 36.6 & 0.97\\
  Ours-Global (all) & 24.7 & 0.23 & 26.0 & 0.41 & 30.7 & 0.91\\
  Ours (all) & \textbf{18.1} &  \textbf{0.16} & \textbf{22.3} & \textbf{0.39} & \textbf{28.6} & \textbf{0.90} \\ 
  \bottomrule
  \end{tabular}
{%
 \caption{Benchmark evaluation on our approach and baseline approaches. Ours-Local and Ours-Global stand for our method with global module and local module removed, respectively. We show the mean error for rotation and translation components for overlapping scan pairs ($\geq 10\%$) and non-overlapping scan pairs ($\leq 10\%$), respectively.}
 \label{tab:baseline-comparison}
}

\vspace{-0.15in}
\end{table}

\noindent\textbf{Baseline approaches.} 
We consider the five baseline approaches:~\textsl{Super4PCS}~\cite{CGF:CGF12446},~\textsl{RobustGR}~\cite{DBLP:conf/eccv/ZhouPK16},~\textsl{ScanComplete}~\cite{Yang_2019_CVPR},~\textsl{Ours-Global}, and \textsl{Ours-Local}. Here, Super4PCS is a widely used global scan matching method; RobustGR is a state-of-the-art global scan matching method that is based on robust regression; ScanComplete is a state-of-the-art approach for indoor RGB-D registration that performs scan completion using the 360-image representation. Moreover, Ours-Global is our approach without the local module. Ours-Local is our approach without the global module, i.e., applying the local module directly on the completed scans.

\noindent\textbf{Evaluation protocol.} We follow the standard protocol of reporting the relative angular error 
$\|\log(R^{\star}{R^{gt}}^{T})\|/\sqrt{2}$
and the relative translation error $\|\bs{t}^{gt}-\bs{t}^{\star}\|$. Here $(R^{gt},\bs{t}^{gt})$ and $(R^{\star},\bs{t}^{\star})$ denote the ground-truth and the output of a method, respectively. In the presence of multiple outputs, we define the best match as the one that minimizes the rotation error. In this section, we report the rotation and the translation errors among the top 1, 3, and 5 matches. The order of the matches are given by the eigenvalues of the consistency matrix $C_{\theta_g,\alpha_{g}, p}$ and the final scores of (\ref{Eq:Total:Loss1}), respectively.  
In addition, we divide scan pairs into two categories: overlapping and non-overlapping. The category of overlapping scans consists of scans pairs that overlap by at least 10\% points (with respect to the scan with fewer number of points). The category of non-overlapping contains the remaining scans. 

\subsection{Analysis of Results}
\label{Subsection:Analysis:of:Results}

Table~\ref{tab:baseline-comparison} and Figure~\ref{fig:figure_baseline} provide quantitative and qualitative results of our approach and baseline methods. Overall, our approach outperforms baseline approaches considerably. The mean rotation/translation errors of the best-match of our approach are 18.1$^{\circ}$/0.16m, 22.3$^{\circ}$/0.39m, and 28.6$^{\circ}$/0.90m on SUNCG, Matterport, and ScanNet, respectively. In contrast, the state of the art method only achieved 31.1$^{\circ}$/0.27m, 34.2$^{\circ}$/0.53m, and 34.1$^{\circ}$/0.93m, respectively. 

\noindent\textbf{Overlapping v.s. non-overlapping.} In the overlapping regime, the relative improvements of mean rotation/translation errors are 40.8\%/35.3\%, 46.2\%/25.6\%, and 55.0\%/5.6\% on the three datasets described above, respectively. Meanwhile, the relative improvements in the non-overlapping regime are 23.6\%/16.3\%, 22.9\%/22.7\%, and 10.5\%/0.7\%, respectively. These statistics show the consistency of our approach across different overlapping rates.

\subsection{Analysis of the Global Module}
\label{Subsection:Global:Network}

We focus on analyzing two important aspects of the global module, namely the effects of using hybrid representation and multiple outputs. 

\noindent\textbf{Single representation versus hybrid representation.} As shown in Table~\ref{tab:global_ablation}, using hybrid representations leads to noticeable performance gains when compared to using a single representation. For mean rotation/translation errors, the relative improvements from the top performing single representation (\textit{ScanComplete} in this case) are 20.4\%/14.8\%, 24.0\%/22.6\%, and 11.7\%/2\% on SUNCG, Matterport, and ScanNet, respectively. It should be noted that the 2D layout and Plane representations individually do not provide sufficient features for relative pose estimation. However, they provide complementary features to the 360-image representation that boost the overall performance of our approach. This is also consistent with  Figure~\ref{Figure:Global:Module}, which shows that the matched features consist of a mixture of planar features and point features from 360-image and 2D layout. 

\noindent\textbf{Best match versus K-best.} With multiple outputs, the top-5 rotation/translation errors of the global module drop significantly. The relative improvements on the three datasets are 68.4\%/8.7\%, 45.0\%/0.1\%, and 45.2\%/16.5\%, respectively. Besides, the relative improvements on non-overlapping scans are significantly bigger than those on overlapping scans. An explanation is that indoor scenes exhibit approximate discrete symmetries, e.g., rotational symmetries of a box-shape, and such symmetries incur ambiguities for non-overlapping scans (i.e., they lack detailed geometric features to lock their relative poses). On the other hand, the orders of these symmetries are relatively small, and top matches tend to capture them.

In addition, the probability distribution of the rank of the matches with the best rotational error are 42\%, 17\%, 17\%, 10\%, and 14\%, respectively. In other words, 58\% of the best matches are not derived from the first eigen-vector. This indicates the importance of leveraging multiple outputs. In Section~\ref{Section:Few:View:ReconstructionAdd}, we show that using multiple outputs for multi-scan reconstruction is superior to using a single output.

\subsection{Analysis of the Local Module}
\label{Subsection:Local:Network}
\setlength\tabcolsep{3.6pt}
\begin{table}
\centering
\footnotesize
\begin{tabular}{l|c|c|c|c|c|c}
  \toprule
  & \multicolumn{2}{c|}{SUNCG} & \multicolumn{2}{c|}{Matterport} &  \multicolumn{2}{c}{ScanNet}\\
  \hline
  & Rot & Trans & Rot & Trans  & Rot & Trans\\
  \hline
  Base-Global+Base-Local & 31.9 & 0.45 & 33.6 & 0.52 & 34.1 & 0.93 \\
  Base-Global+Ours-Local & 27.8 & 0.24& 30.1 & 0.51 & 31.2 & 0.92 \\
  Ours-Global+Base-Local & 25.9 & 0.44 & 25.4 & 0.40 & 30.4 & 0.91 \\
    Ours-Global+Ours-Local & \textbf{18.1} & \textbf{0.16} & \textbf{22.3} & \textbf{0.34} & \textbf{28.6} & \textbf{0.90} \\
  \bottomrule
  \end{tabular}
 \caption{Benchmark evaluation on the different combination of global initialization and local refinement. We use the strongest baseline \protect\cite{Yang_2019_CVPR} to serve as global module, and SparseICP~\protect\cite{Bouaziz:2013:SIC} as baseline local module. We show the mean rotation and translation errors on each dataset.}
 \label{tab:local}

\vspace{-0.15in}
\end{table}

Overall, the local module leads to noticeable performance gains from the outputs of the global module. As shown in Table~\ref{tab:baseline-comparison}, the mean rotation/translation errors of the top 1 match reduced from 24.7$^{\circ}$/0.23m to 18.1$^{\circ}$/0.16m on SUNCG, from 26.0$^{\circ}$/0.41m to 22.3$^{\circ}$/0.39m on Matterport, and from 30.7$^{\circ}$/0.91m to 28.6$^{\circ}$/0.90m on ScanNet, respectively. These statistics indicate the robustness of our local module. 

\noindent\textbf{Baseline comparison.} Our approach outperforms the robust version of ICP~\cite{Bouaziz:2013:SIC} when applied on the output of the global module (See Table~\ref{tab:local}). The mean rotation/translation errors reduce from 25.9$^{\circ}$/0.44m, 25.4$^{\circ}$/0.40m, and 30.4$^{\circ}$/0.91m to 18.1$^{\circ}$/0.16m, 22.3$^{\circ}$/0.34m, and 28.6$^{\circ}$/0.90m. Such improvements mainly come from the predicted geometric relations. Our approach also outperforms the baseline that combines the output of ScanComplete and our local module. The relative improvements are salient across all the datasets. These statistics indicate that having good initial poses for local refinement is key to the success of relative pose estimation.

\noindent\textbf{The quality of geometric relations prediction.} The average accuracy of geometric relations across all three datasets are 46\%, 69\%, 75\%, and 93\% for no-relation, perpendicular, parallel, co-planar respectively. The statistics are collected from scan pairs perturbed from the underlying ground-truth by a random rotation with degree in $[0,30^{\circ}]$ and a random translation in $[-0.5m, 0.5m]^3$. Although the predictions are imperfect, the success of our approach lies in using robust norms to filter out wrong predictions automatically. Due to space constraints, we leave a detailed analysis of the predicted geometric relations to the supp. material. 

\subsection{Application in Few-View Reconstruction}
\label{Section:Few:View:ReconstructionAdd}

We have studied the impacts of our relative pose estimation approach in multi-scan alignment. To this end, we pair our approach with the multi-view pose optimizer of MRF-SFM~\cite{crandall2013pami}. The goal is to find the most consistent subset of relative poses among the outputs of our approach. To analyze the benefits of having multiple relative pose estimations between each scan pair, we report the performance of using top-1, top-3, top-5 outputs of our approach, respectively. We performed an experimental study on ScanNet, using 5/10/15/20  randomly sampled scans.
Regarding the evaluation protocol, we report absolute errors in rotations and translations of aligned scans using the same metric for evaluating relative poses. Since MRF-SFM extracts consistent relative transformations, the mean rotation/translation errors of the resulting absolute transformations are significantly reduced.  The baseline approaches include RobustRecons~\cite{choi2015robust} and Fine-to-Coarse~\cite{halber2017fine}, which are two state-of-the-art multi-scan alignment approaches. Fine-to-Coarse additionally utilizes the scanning order among the input scans.

\setlength\tabcolsep{3.6pt}
\begin{table}
\centering
\footnotesize
\begin{tabular}{l|c|c|c|c|c|c|c|c}
  \toprule
  & \multicolumn{2}{c|}{5 Scans} & \multicolumn{2}{c|}{10 Scans} &  \multicolumn{2}{c|}{15 Scans} &  \multicolumn{2}{c}{20 Scans}\\
  \hline
  & Rot & Trans & Rot & Trans  & Rot & Trans & Rot & Trans\\
  \hline
Fine2C.\cite{halber2017fine} & 96.5 & 1.41 & 62.4 & 0.93 & 69.6 & 1.57  & 68.6 & 1.43\\
RobustRec\cite{choi2015robust} & 33.1 & 0.78& 25.1 & 0.52 & 16.5 & 0.33 & 12.3 & 0.26  \\
Our-top-1 & 8.39 & 0.20 & 3.55 & 0.15 & 2.97 & 0.134  & 2.31 & 0.12\\
Our-top-3 & 5.77 & \textbf{0.18} & 3.12 & \textbf{0.12} & 2.26  & 0.112 & 1.88 & 0.10\\
Our-top-5 & \textbf{5.13} & \textbf{0.18} & \textbf{2.73} & \textbf{0.12} & \textbf{2.13} & \textbf{0.10} & \textbf{1.81} & \textbf{0.09}\\
  \bottomrule
  \end{tabular}
\caption{Comparions on multi-scan alignment using our approach + MRF-SFM\cite{crandall2013pami} and two baseline approaches. Our approach shows a clear advantage in the sparse setting, e.g., using 5 scans. }
\label{tab:few:view}
\vspace{-0.15in}
\end{table}

As shown in Figure~\ref{Fig:Few:View} and Table~\ref{tab:few:view}, our approach significantly outperforms baseline approaches. The improvements are salient in the sparse setting, where our approach shows high-quality outputs. In such a sparse setting, RobustRecons also outperforms Fine-to-Coarse, as the latter has to utilize dense inputs to detect initial planes. Utilizing top-3 and top-5 matches also improves the performance. One explanation is that there are more correct inputs for multi-scan alignment when utilizing top-3 and top-5 matches.

\section{Conclusions and Future Work}
\label{Section:Conclusions:Future:Work}

In this paper, we have introduced an approach for estimating the relative pose between two potentially non-overlapping RGB-D scans. Our approach combines a global module for computing candidate relative poses and a local module for refining each candidate pose. Our method can output multiple relative poses, and the resulting pose quality is considerably better than state-of-the-art approaches for both overlapping and non-overlapping scans.

{\small
\bibliographystyle{ieee_fullname}
\bibliography{egbib}
}

\clearpage

\appendix
\section{Supplementary Material}

\subsection{360-image Completion Module}
\begin{figure*}
\centering
\footnotesize
\def\imh{0.073\textwidth}
\def\imw{0.15\textwidth}
\newcommand{\T}[1]{\raisebox{-0.5\height}{#1}}
\setlength{\tabcolsep}{1pt}
\begin{tabular}{cccccccc}

\rotatebox[origin=c]{90}{Ours} &
\T{\includegraphics[width=\imw]     {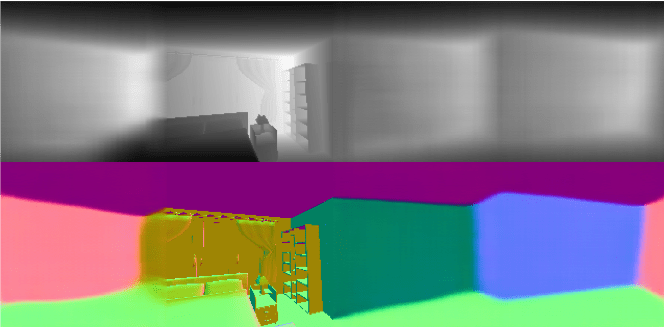}} &
\T{\includegraphics[width=\imw]   
{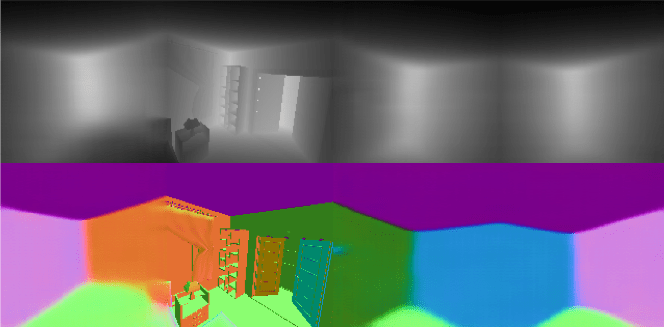}} & 
\T{\includegraphics[width=\imw]      {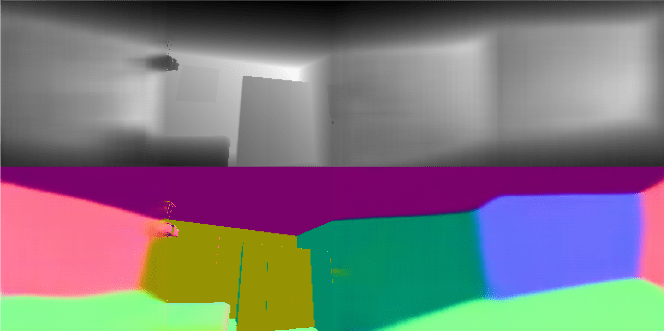}} & 
\T{\includegraphics[width=\imw]    {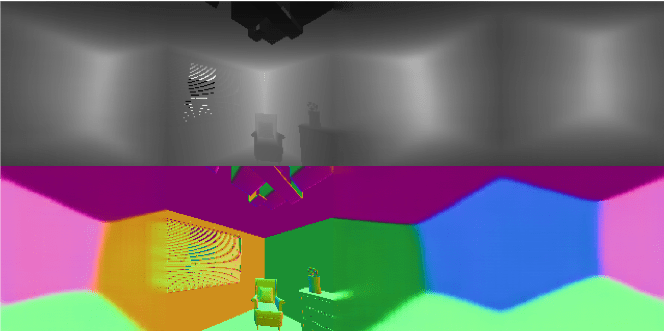}} &
\T{\includegraphics[width=\imw]    {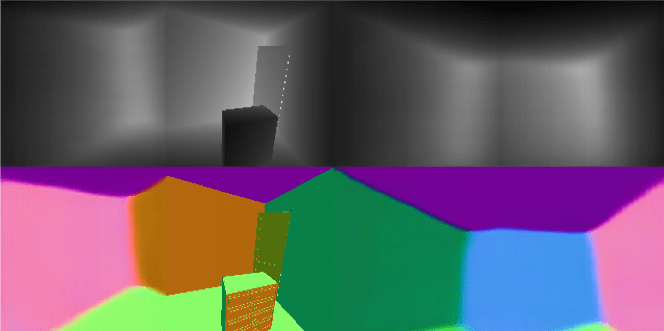}} &
\T{\includegraphics[width=\imw]    {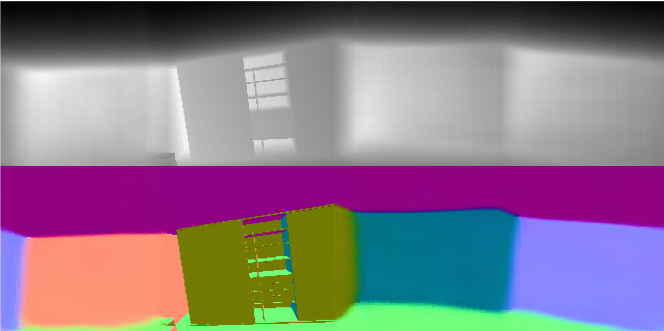}} &
\\ 
[+21pt]
\rotatebox[origin=c]{90}{G. T} &
\T{\includegraphics[width=\imw]     {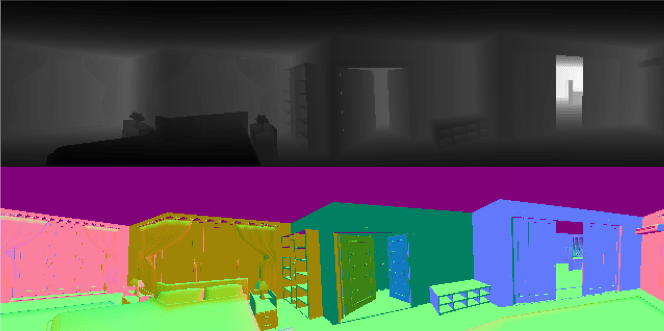}}&
\T{\includegraphics[width=\imw]   
{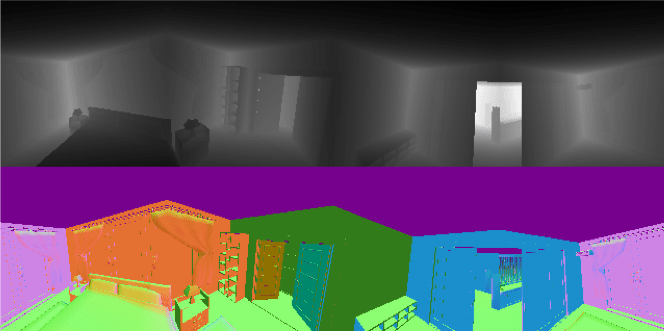}} & 
\T{\includegraphics[width=\imw]      {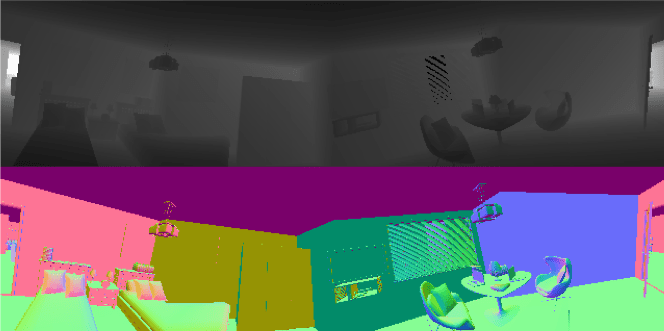}} & 
\T{\includegraphics[width=\imw]    {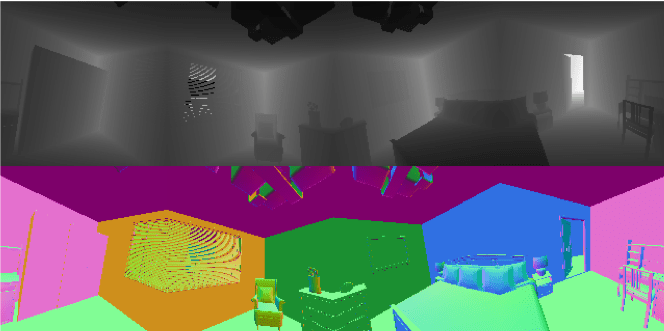}} &
\T{\includegraphics[width=\imw]    {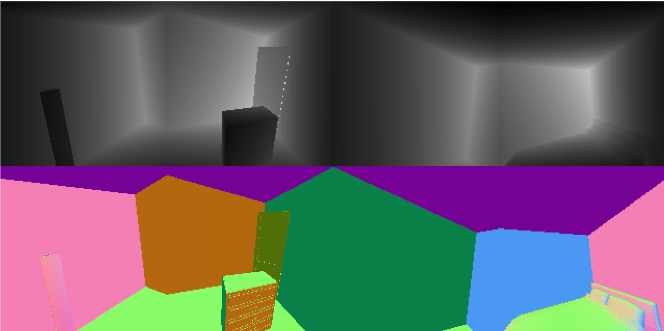}} &
\T{\includegraphics[width=\imw]    {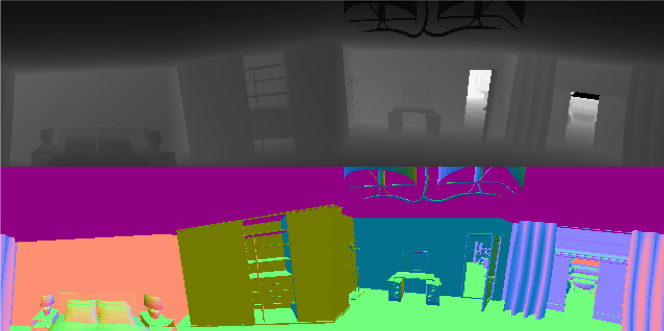}} \\
[+21pt]

\end{tabular}

\caption{360 image completion results. First row is completed normal/depth, second row is ground truth.}
\label{fig:360_example}
\end{figure*}

\noindent\textbf{Network Architecture}
We use the same network architecture as in \cite{Yang_2019_CVPR}. For simplicity, we remove the image warping process and the recurrent module. \\
\noindent\textbf{Qualitative Results} We show qualitative results in Figure~\ref{fig:360_example}.

\subsection{2D Layout Completion Module}
\noindent\textbf{Network Architecture}
The 2D layout module takes in partial scan and output its 2D layout completion. This module consists of two parts, feature extraction module and layout completion module. The feature extraction module is responsible to convert observed partial scans into a partial 2D feature grid. The layout completion module then takes in such partial feature grid and hallucinate a complete layout. The feature extraction module consist of two network works together. The first is a PointNet-style network that operate on point cloud input directly, the second network is a convolutional image backbone for which we use ResNet18 with ImageNet pre-trained weights. The feature from these two networks are combined together to assign a feature for each point. In order to project the 3D points into 2D grid, we simultaneously predict a floor plane (parameterized as plane equation) using max pooled per-point feature. Using the predicted floor plane equation, we can determine which 2D cell one 3D point should rest in. In order to accommodate the variation in height dimension, we further bin the height axis into 4 bins at height $h< -0.1m$, $-0.1m\leq h \leq 0.7m$, $0.7m\leq h \leq 1.5m$, $1.5m<h$ respectively. We average pool the features inside each bin to get a 2D feature map where only cell that has 3D points above it has feature (others have all zero feature vector). Then we pass this feature map through a convolutional encoder-decoder structure to get final 2D layout prediction. Please refer to Figure \ref{fig:layout_arch} for details.

\begin{figure*}
\centering
\footnotesize
\def\imh{0.25\textwidth}
\def\imw{0.25\textwidth}
\newcommand{\T}[1]{\raisebox{-0.5\height}{#1}}
\setlength{\tabcolsep}{1pt}
\begin{tabular}{cccccccc}

\T{\includegraphics[width=\imw]   {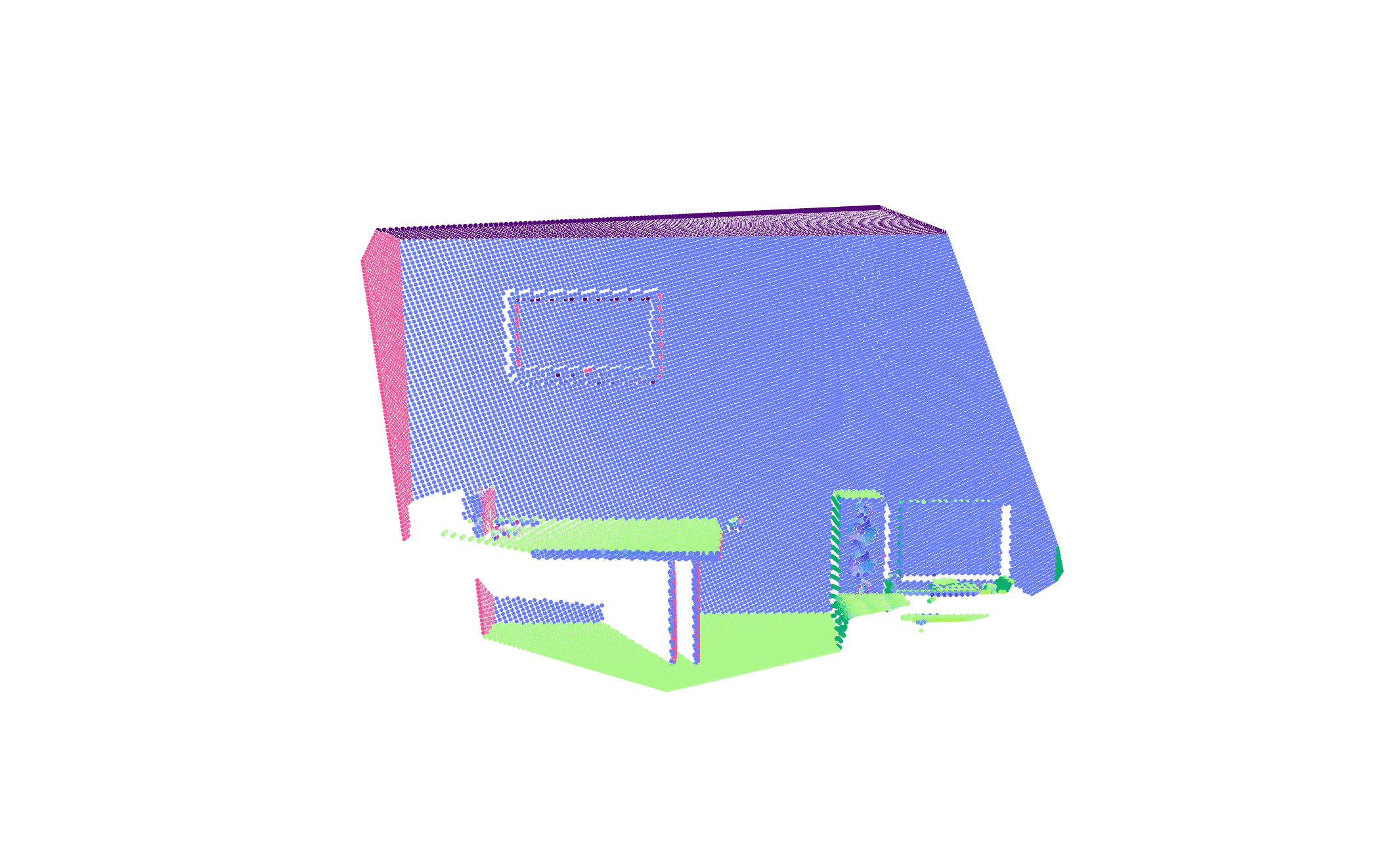}} & 
\T{\includegraphics[width=\imw]      {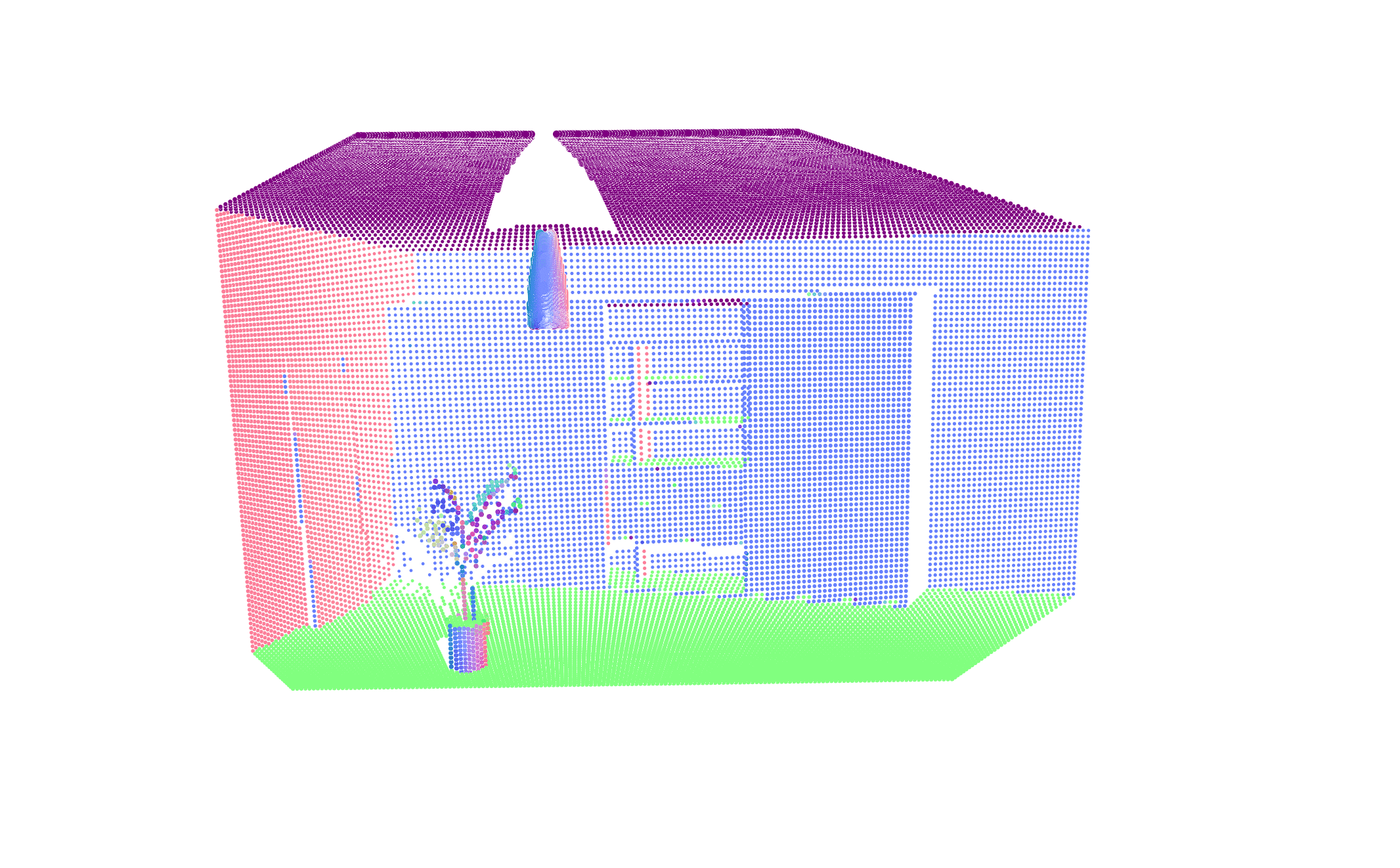}} & 
\T{\includegraphics[width=\imw]    {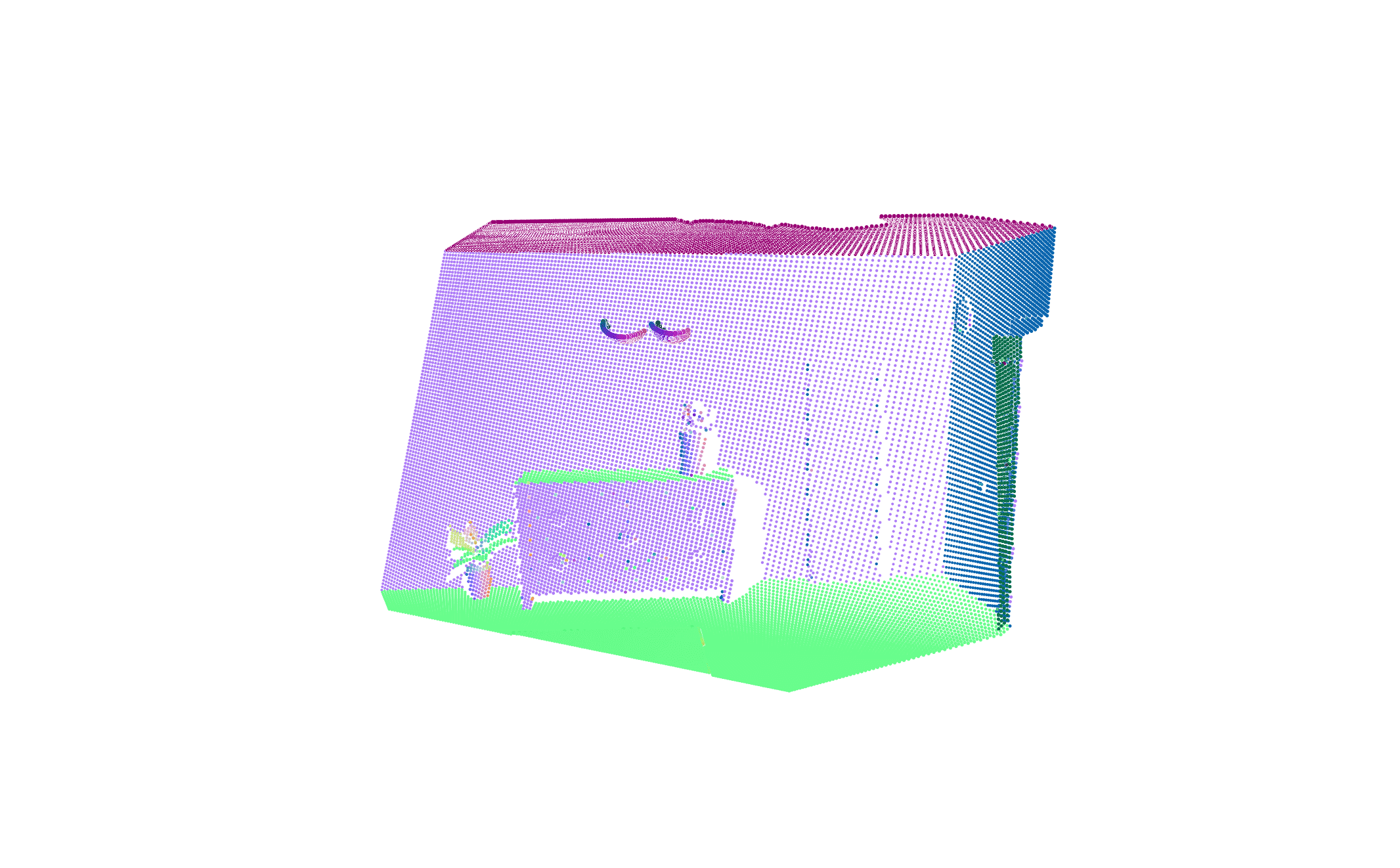}} & 
\T{\includegraphics[width=\imw]     {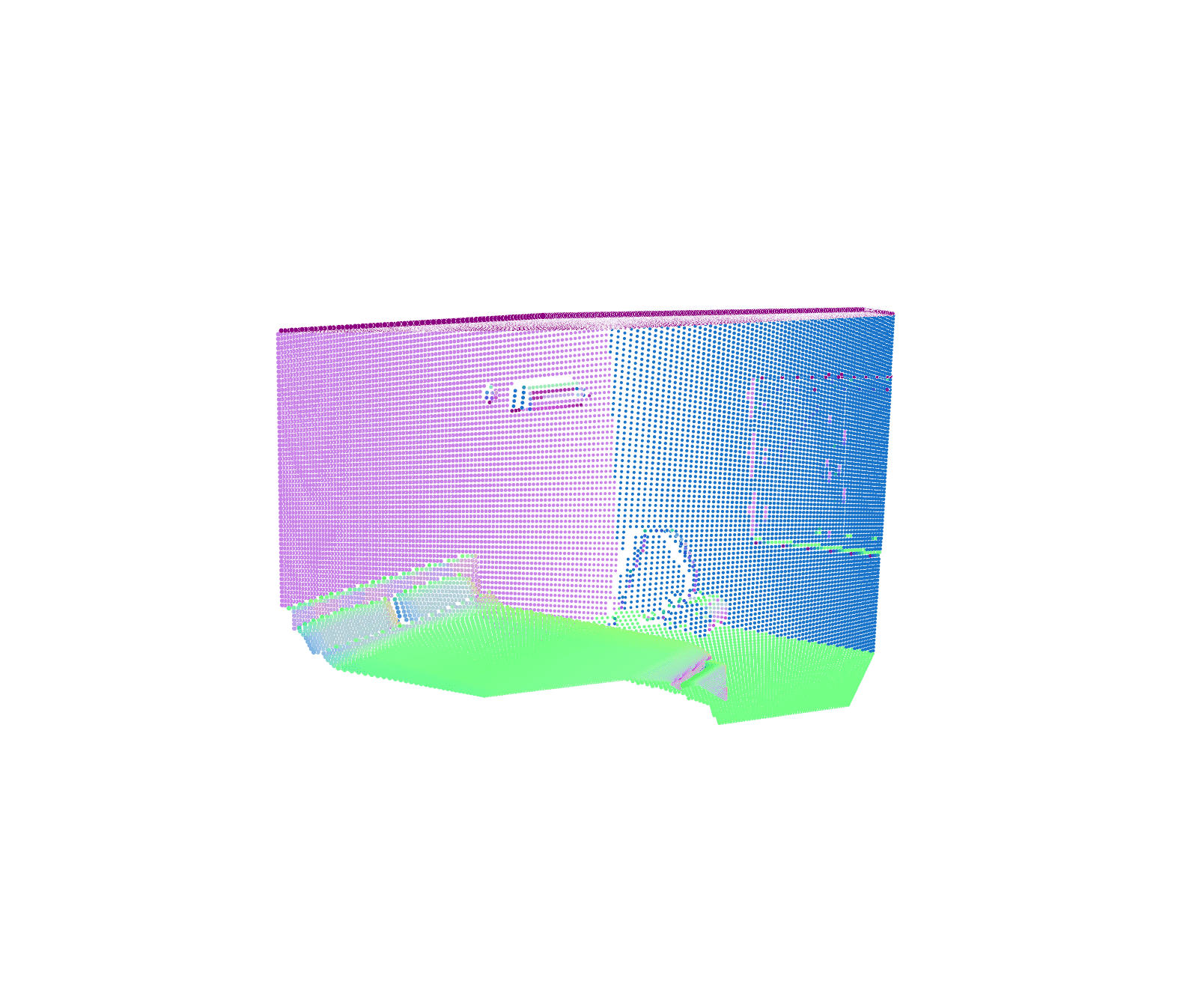}} \\ [+21pt]

\T{\includegraphics[width=\imw]   {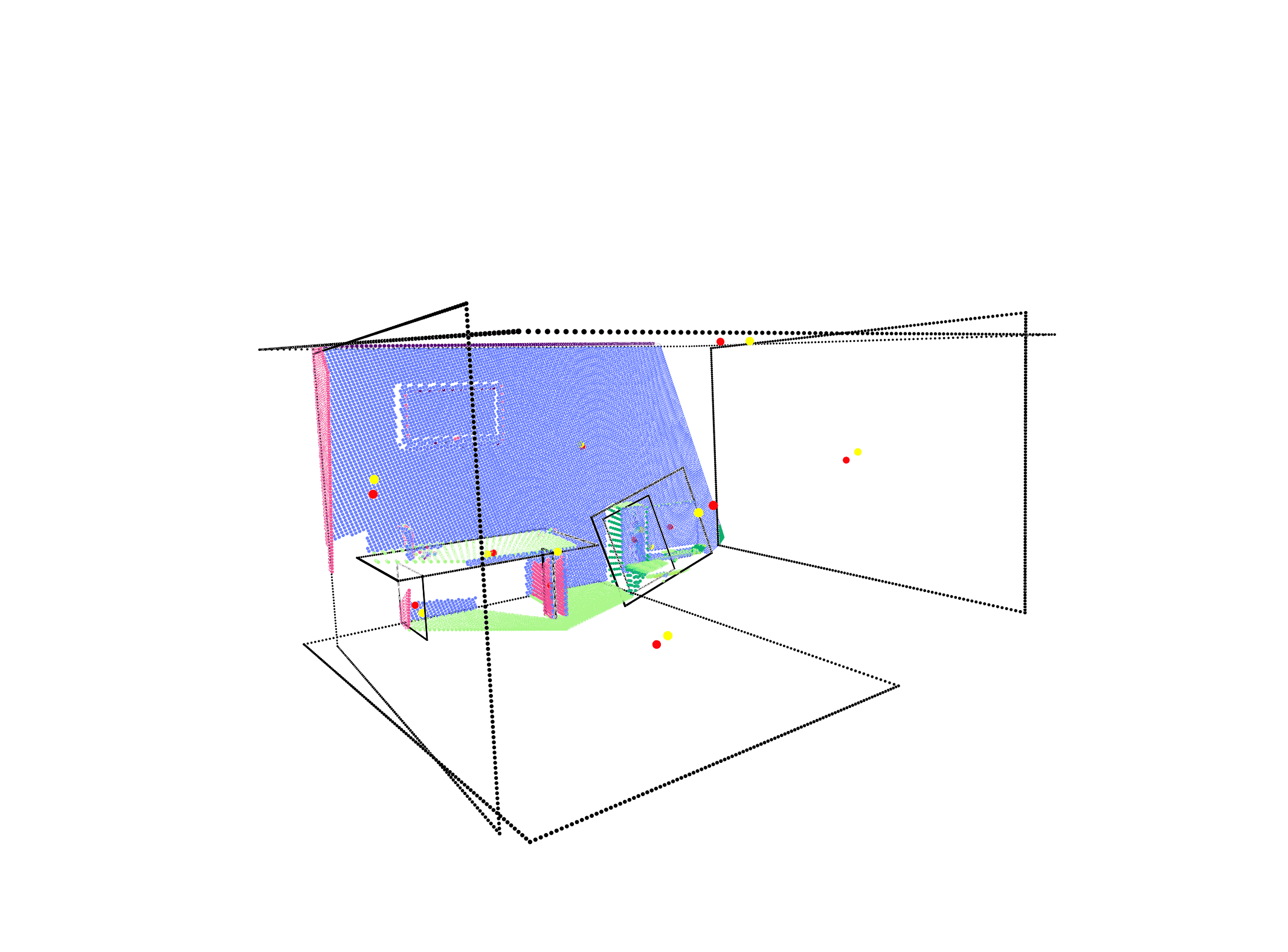}} & 
\T{\includegraphics[width=\imw]      {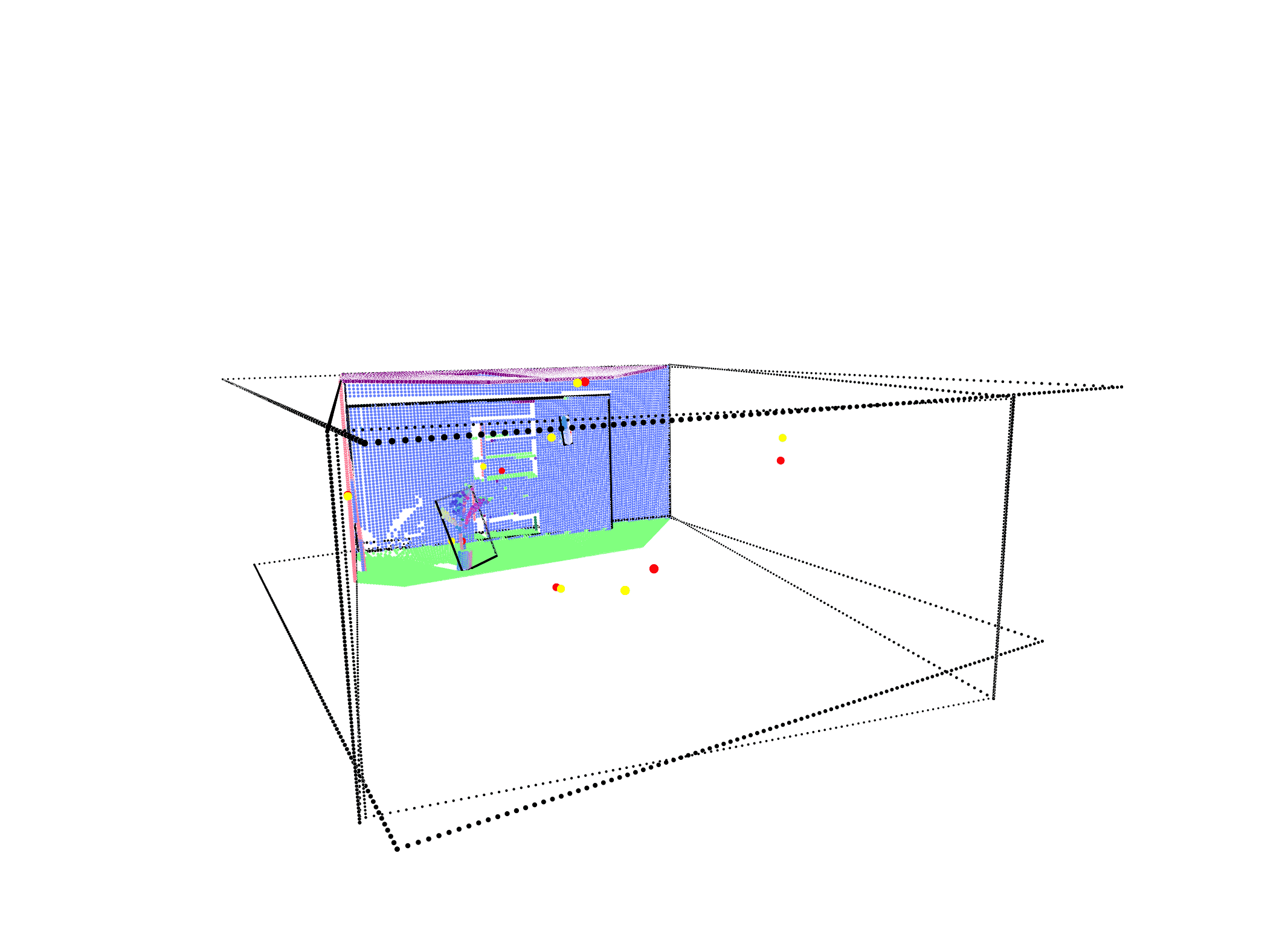}} & 
\T{\includegraphics[width=\imw]    {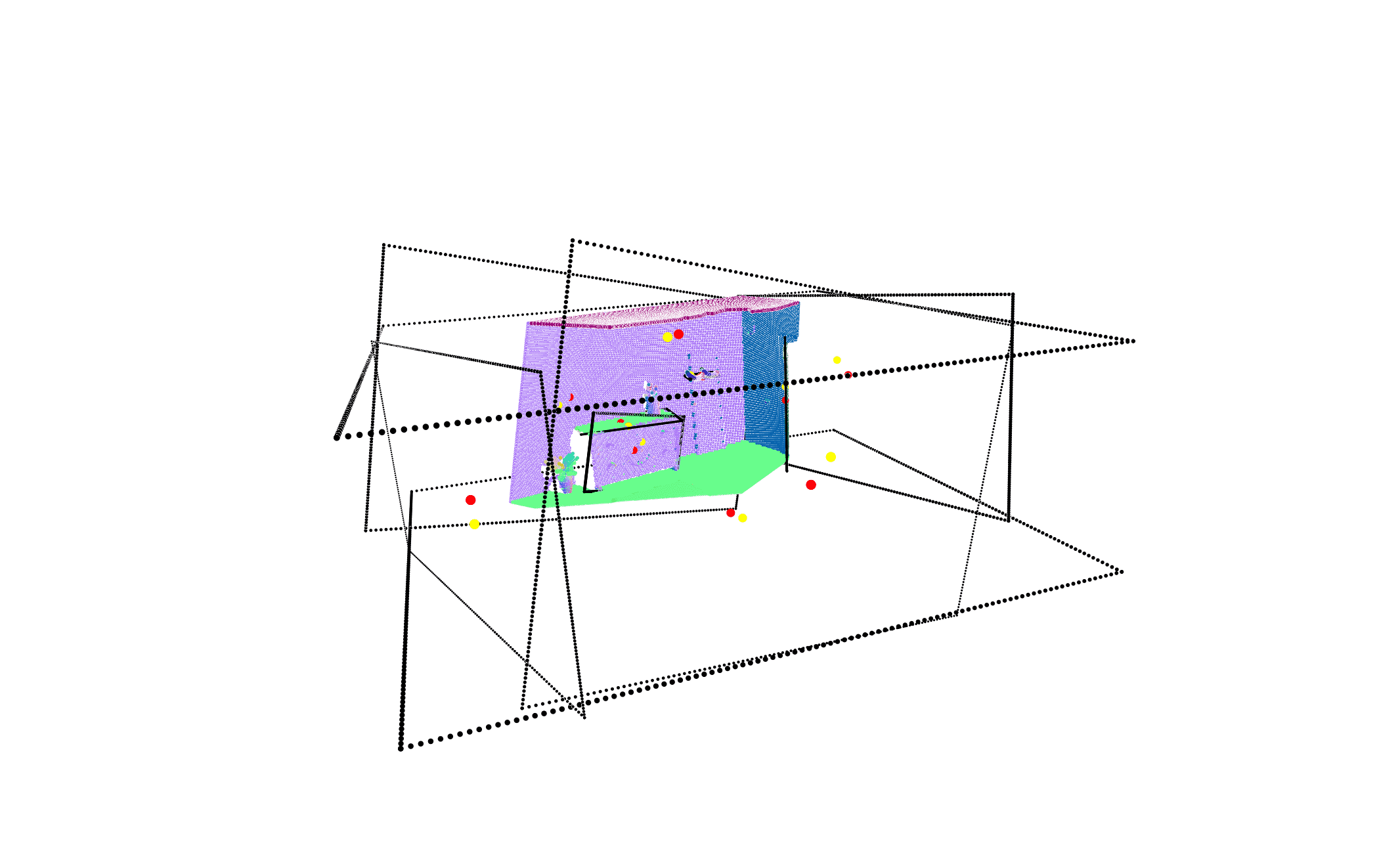}} & 
\T{\includegraphics[width=\imw]     {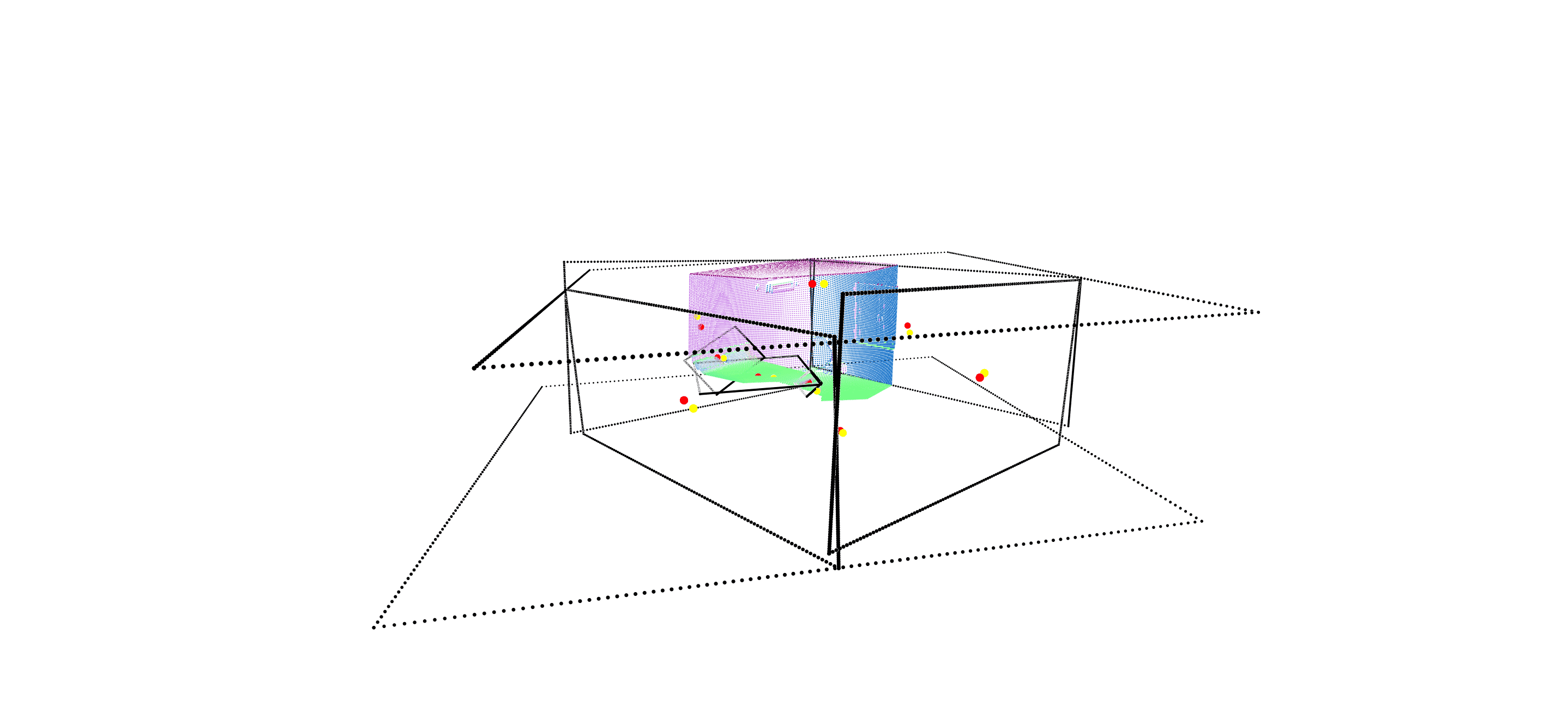}} \\ [+21pt]

\T{\includegraphics[width=\imw]   {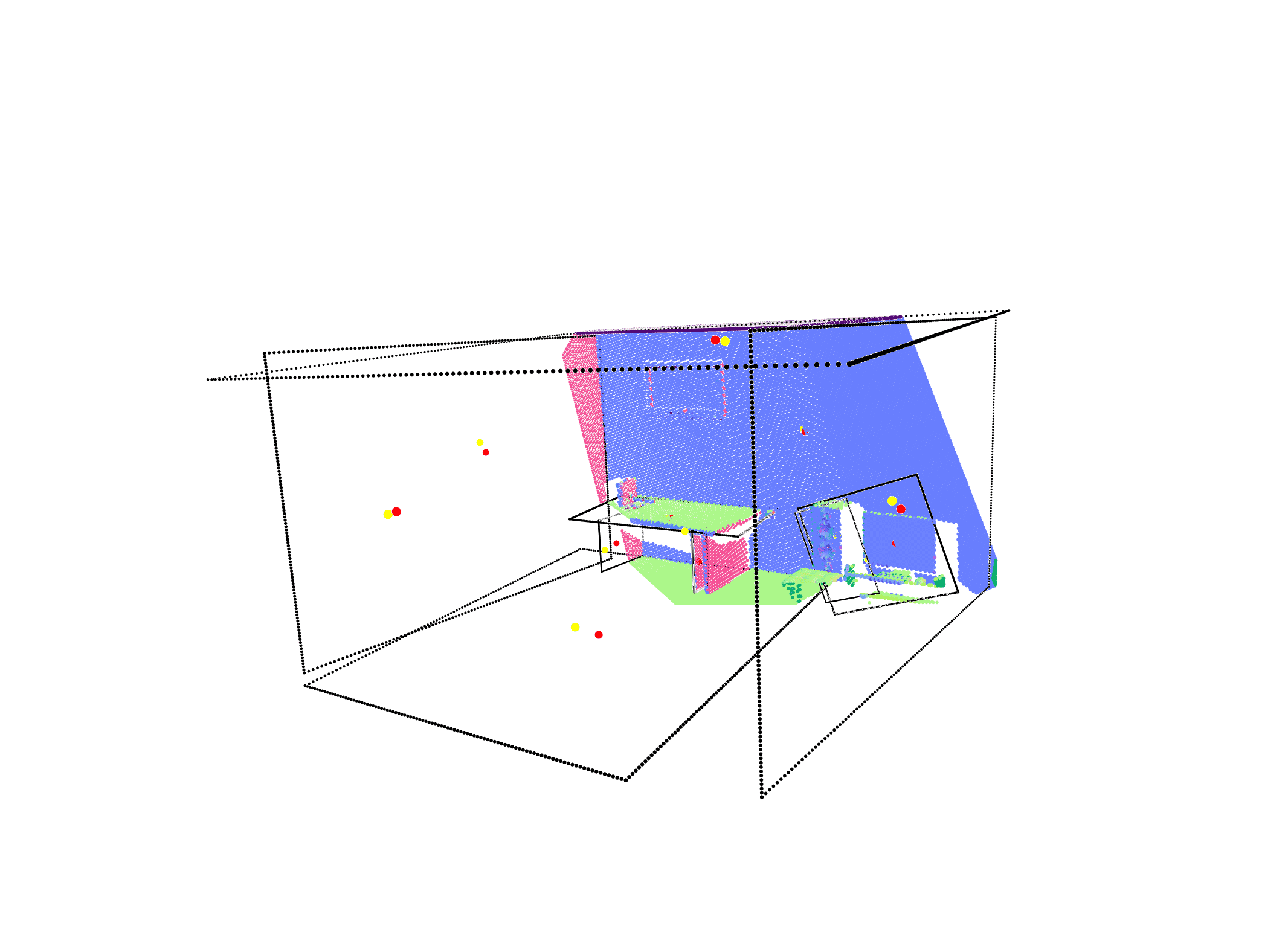}} & 
\T{\includegraphics[width=\imw]      {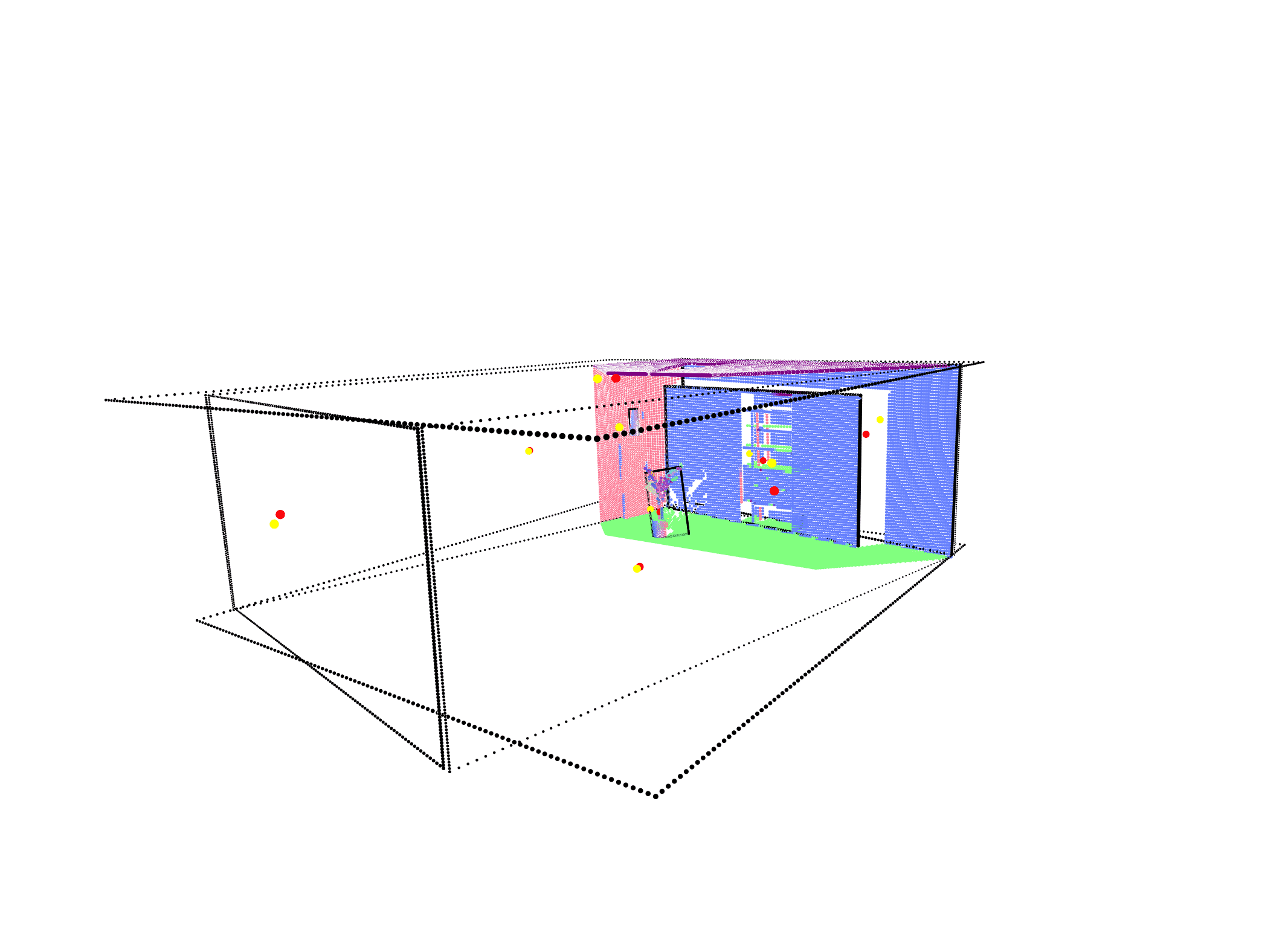}} & 
\T{\includegraphics[width=\imw]    {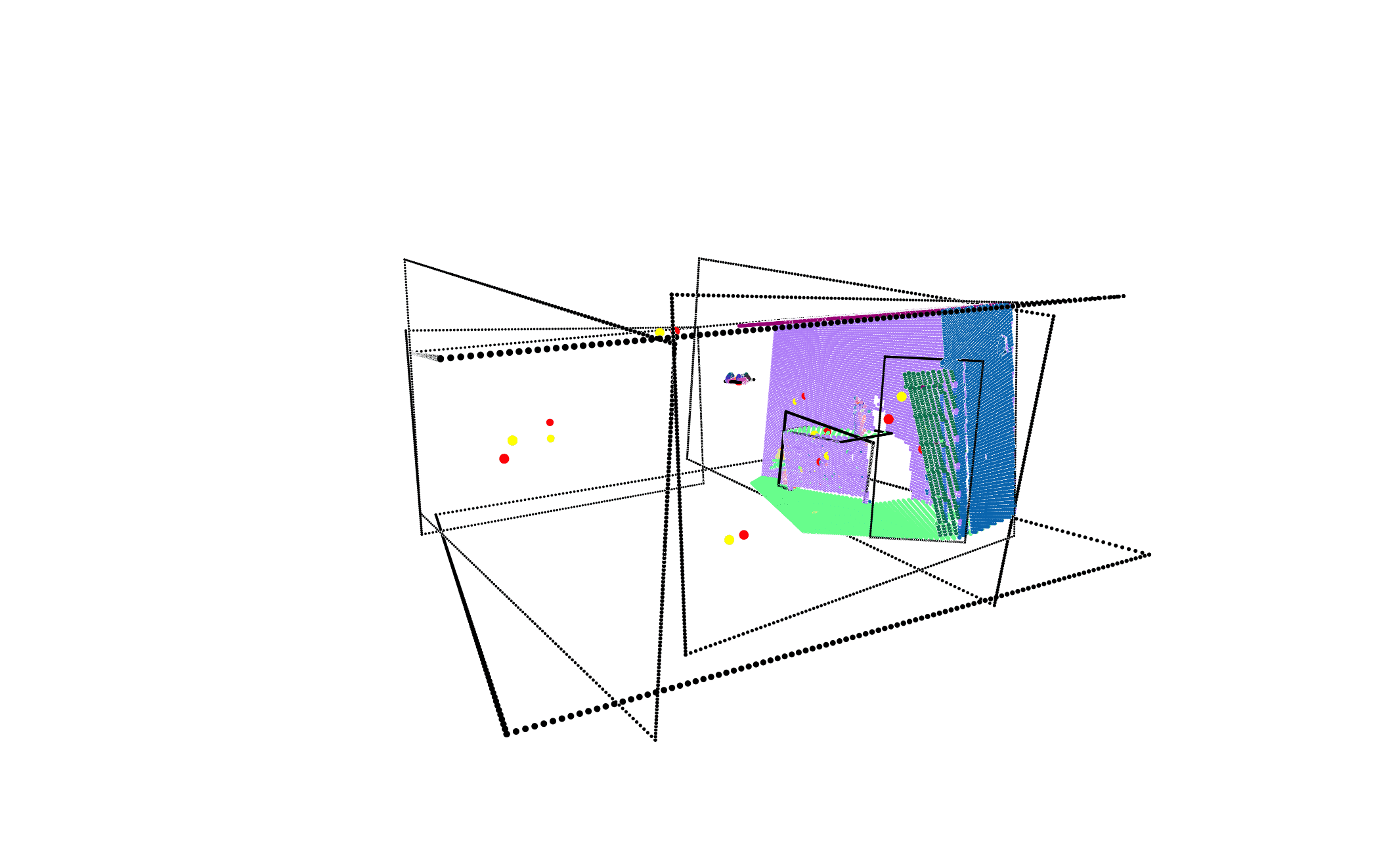}} & 
\T{\includegraphics[width=\imw]     {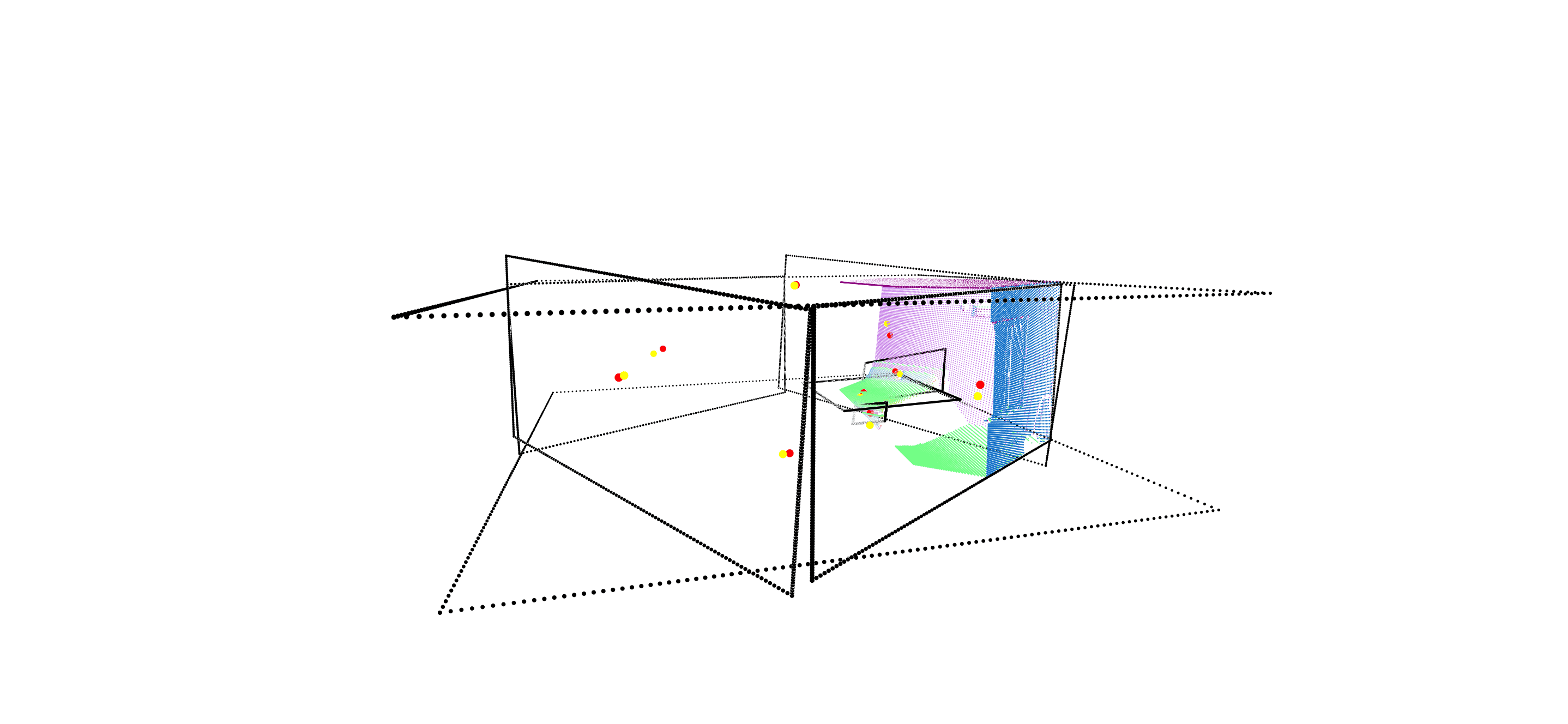}} \\ [+21pt]

\end{tabular}

\caption{Planar patch completion results. Each column shows one example. The first row is the input point cloud of network. The second and third rows are two views of planar patch completion results. The red points in the figure show the ground truth of plane center and the yellow points show the prediction of our network.}
\label{fig:supp_2}
\vspace{-0.15in}
\end{figure*}

\begin{figure}
\centering
\footnotesize
\def\imh{0.073\textwidth}
\def\imw{0.12\textwidth}
\newcommand{\T}[1]{\raisebox{-0.5\height}{#1}}
\setlength{\tabcolsep}{1pt}
\begin{tabular}{ccc}

\T{\includegraphics[width=\imw]     {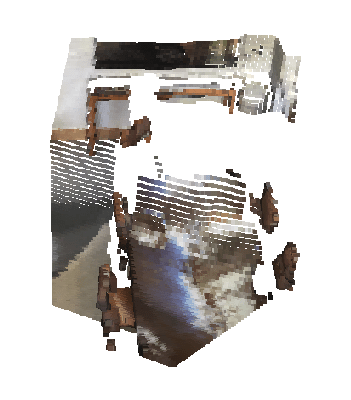}}&
\T{\includegraphics[width=\imw]   {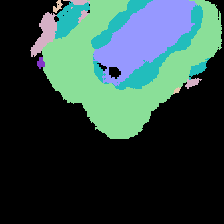}} &
\T{\includegraphics[width=\imw]     {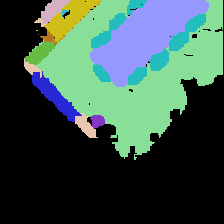}}  \\ [-3pt]

\T{\includegraphics[width=\imw]     {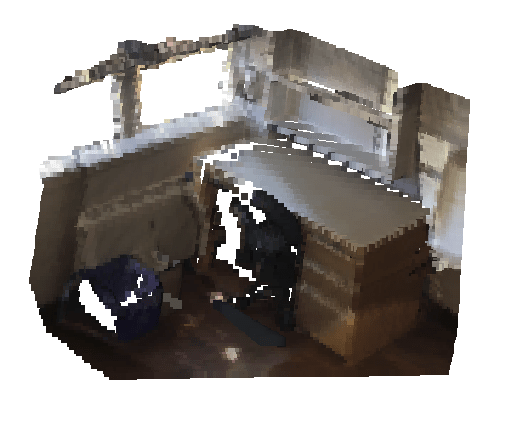}}&
\T{\includegraphics[width=\imw]   {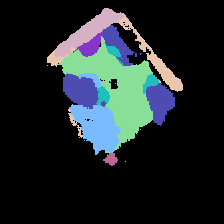}} &
\T{\includegraphics[width=\imw]     {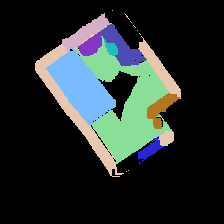}}  \\ [-3pt]

\T{\includegraphics[width=\imw]     {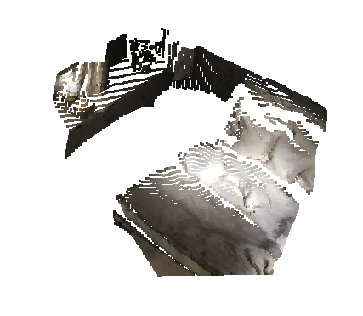}}&
\T{\includegraphics[width=\imw]   {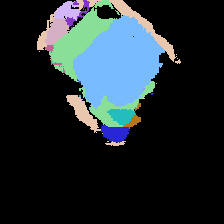}} &
\T{\includegraphics[width=\imw]     {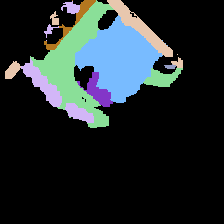}}  \\ [-3pt]

\T{\includegraphics[width=\imw]     {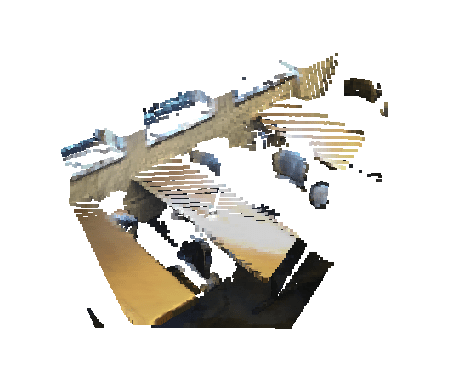}}&
\T{\includegraphics[width=\imw]   {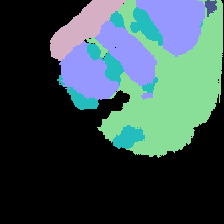}} &
\T{\includegraphics[width=\imw]     {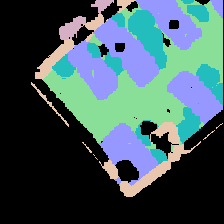}}  \\ [-3pt]

\\\\
\end{tabular}

\caption{Layout completion results. The left column is the input scan, middle column is inferred layout completion. The right column is ground truth. }
\label{fig:supp_4}
\vspace{-0.15in}
\end{figure}

\noindent\textbf{Qualitative Results}
We show qualitative results in Figure ~\ref{fig:supp_4}.

\subsection{Planar Patch Completion Module}

\begin{figure}
\centering
\footnotesize
\def\imh{0.22\textwidth}
\def\imw{0.22\textwidth}
\newcommand{\T}[1]{\raisebox{-0.5\height}{#1}}
\setlength{\tabcolsep}{1pt}
\begin{tabular}{cc}

\T{\includegraphics[width=\imw]     {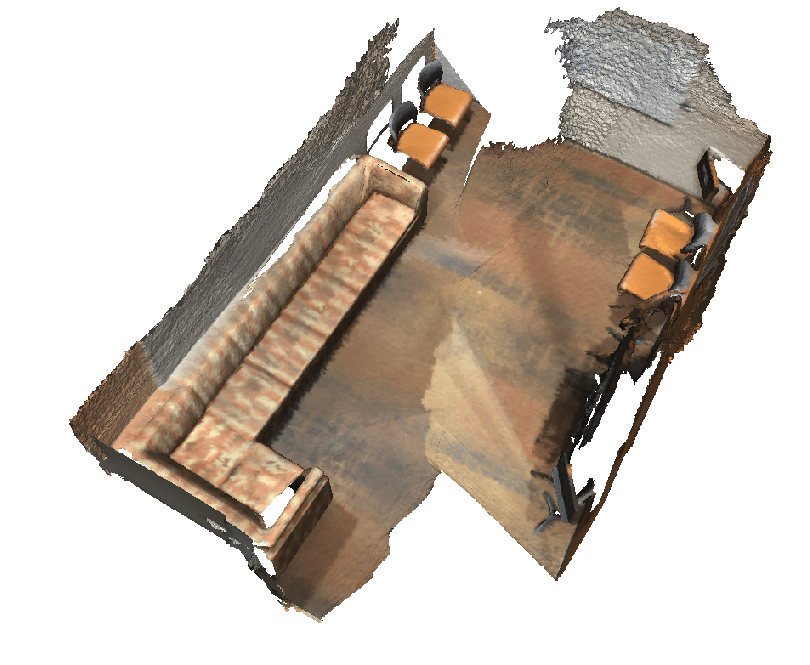}}&
\T{\includegraphics[width=\imw]       {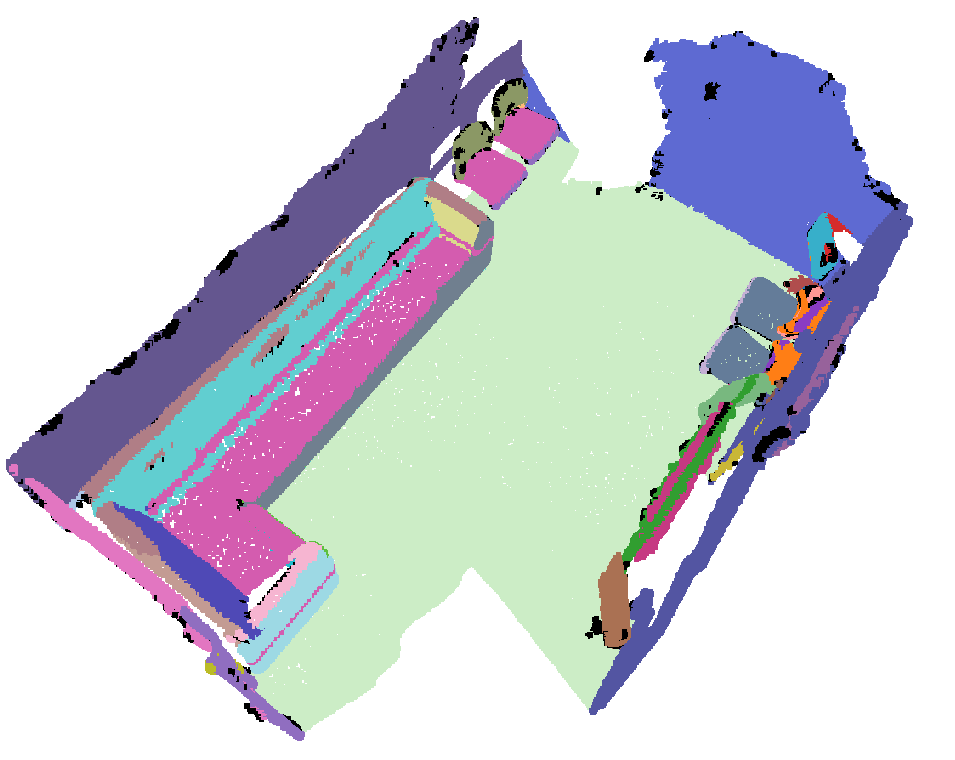}} \\ [-3pt]

\T{\includegraphics[width=\imw]     {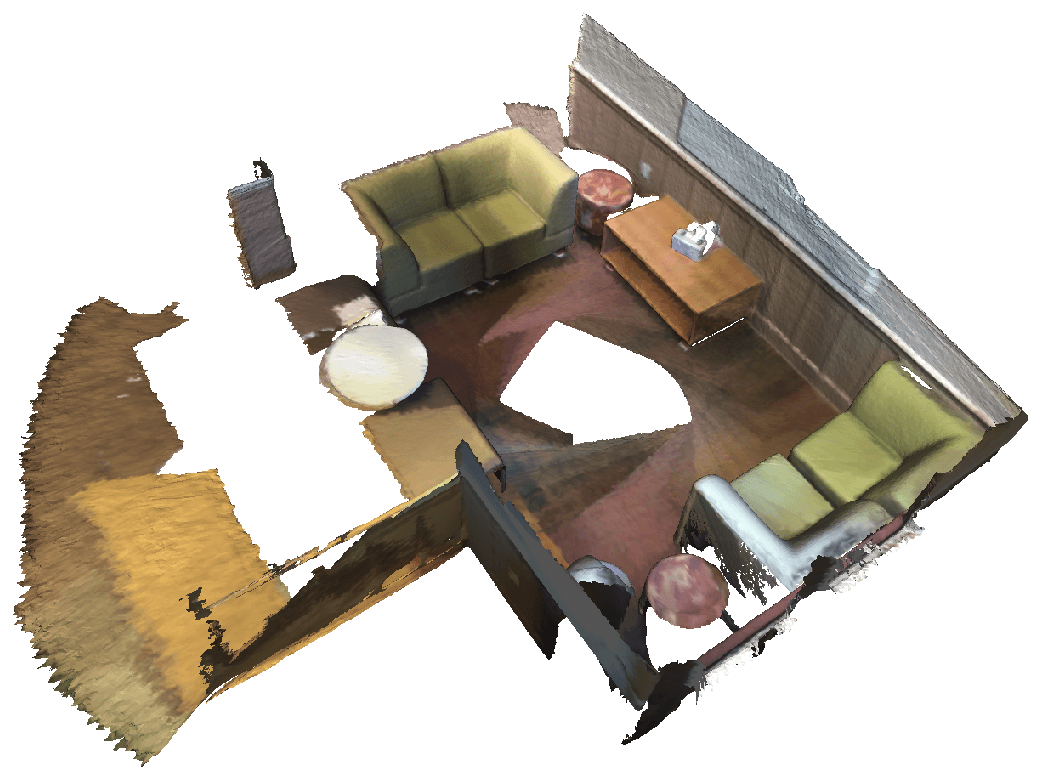}}&
\T{\includegraphics[width=\imw]       {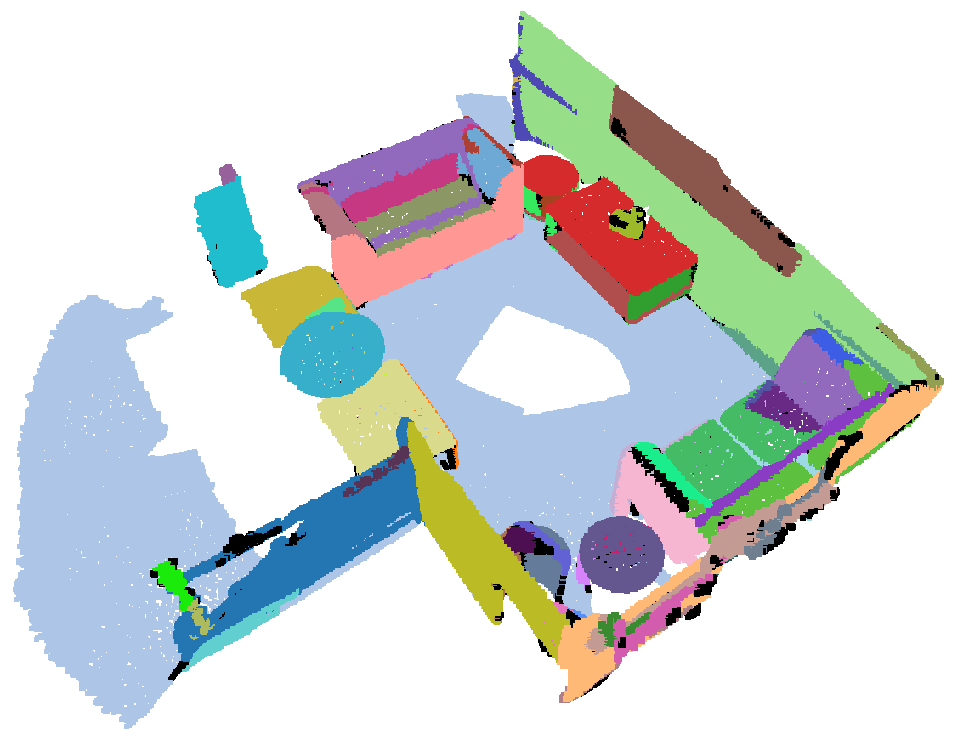}} \\ [-3pt]
\T{\includegraphics[width=\imw]     {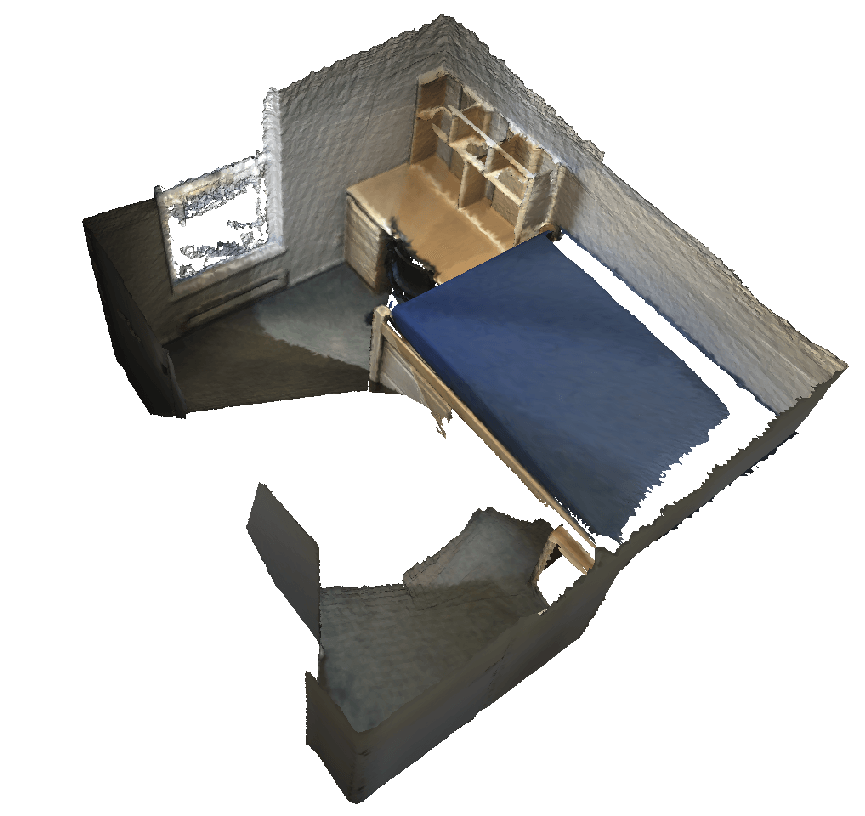}}&
\T{\includegraphics[width=\imw]       {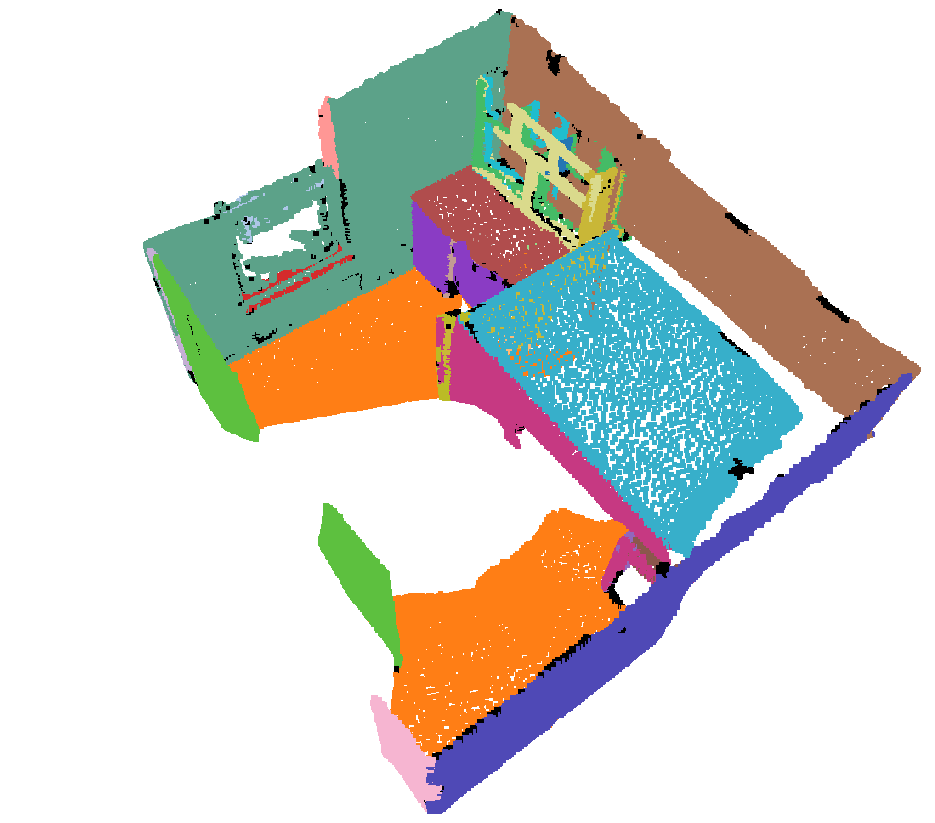}} \\ [-3pt]
\end{tabular}

\caption{RANSAC plane fitting results. Left column is room point cloud, right column is plane fitting visualization. All points belongs to same plane are colored with the same unique color. }
\label{fig:supp_plane}
\vspace{-0.15in}
\end{figure}

\noindent\textbf{Data Generation} We estimate the ground truth plane using the whole room point cloud. We use RANSAC to estimate planes, and discard plane that has multiple disconnected components. We also leverage the semantic segmentation label available on point cloud to discard plane that joins two points from different categories. Examples of resulting plane estimation can be found at Figure \ref{fig:supp_plane}. For each perspective image, we render an index map that tells which plane each pixel belongs to. The total planes we are predicting consist of two parts: planes that appears in the perspective image over certain threshold (500 pixels in our experiments), and structural planes that consist of walls, ceiling and floors. \\

\begin{figure*}
    \centering
    \begin{subfigure}{0.9\textwidth}
        \centering
        \includegraphics[width=\linewidth]{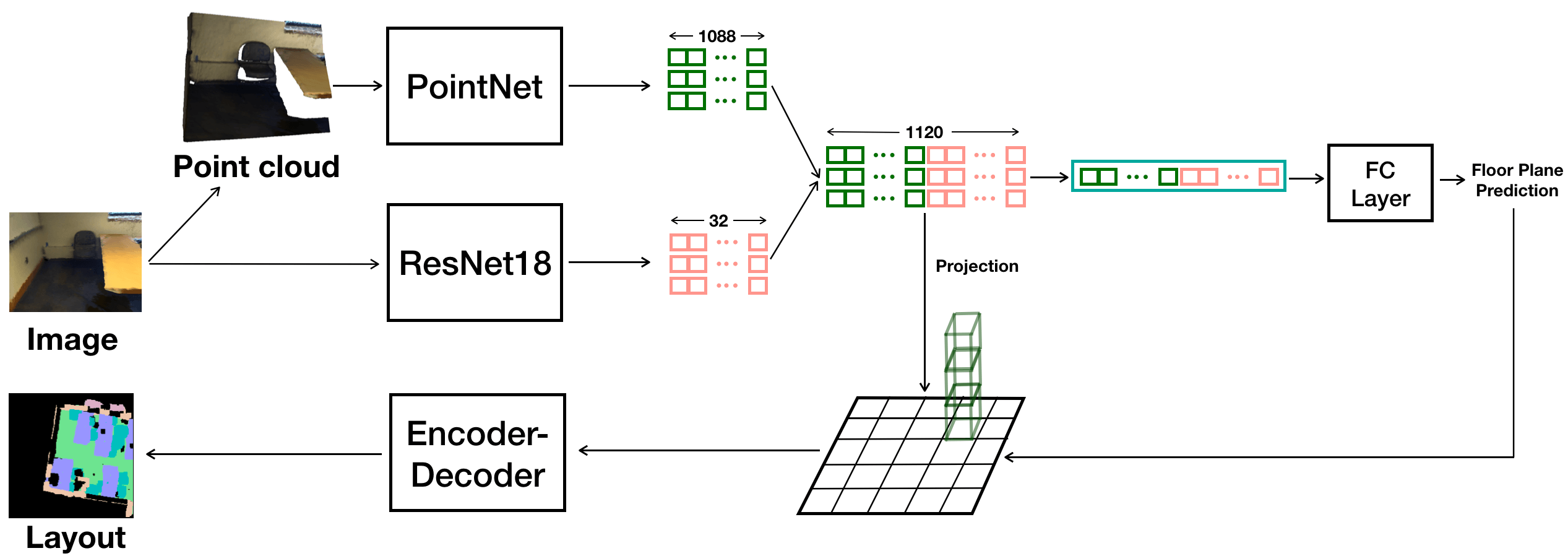} 
        \caption{2D layout completion module} \label{fig:layout_arch}
    \end{subfigure}
    \vspace{1cm}
    \begin{subfigure}{\textwidth}
    \centering
        \includegraphics[width=\linewidth]{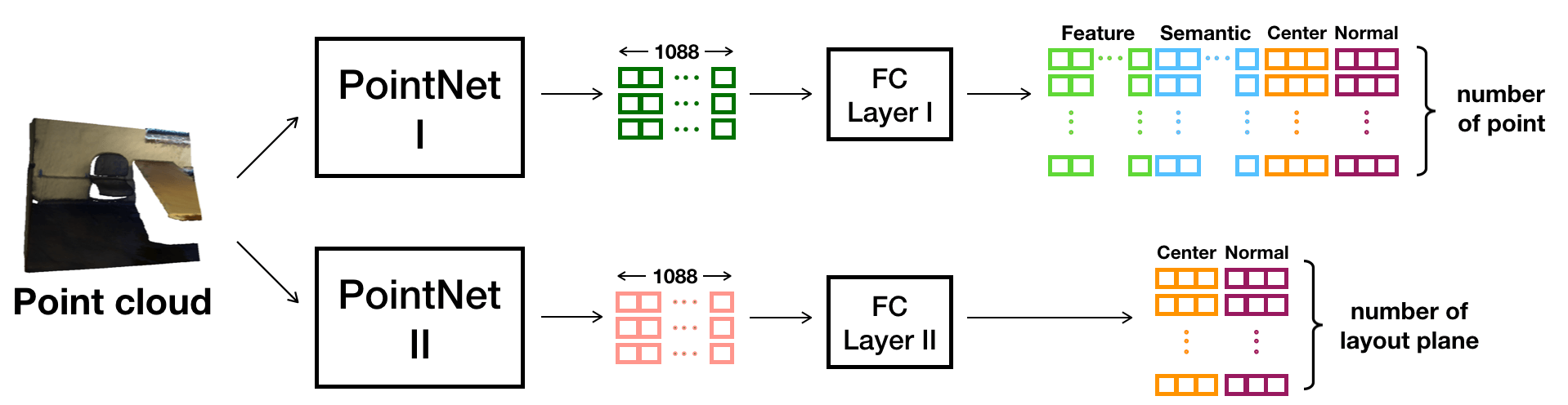} 
        \caption{plane completion module} \label{fig:plane_arch}
    \end{subfigure}
    \vspace{1cm}
    \begin{subfigure}{\textwidth}
    \centering
        \includegraphics[width=\linewidth]{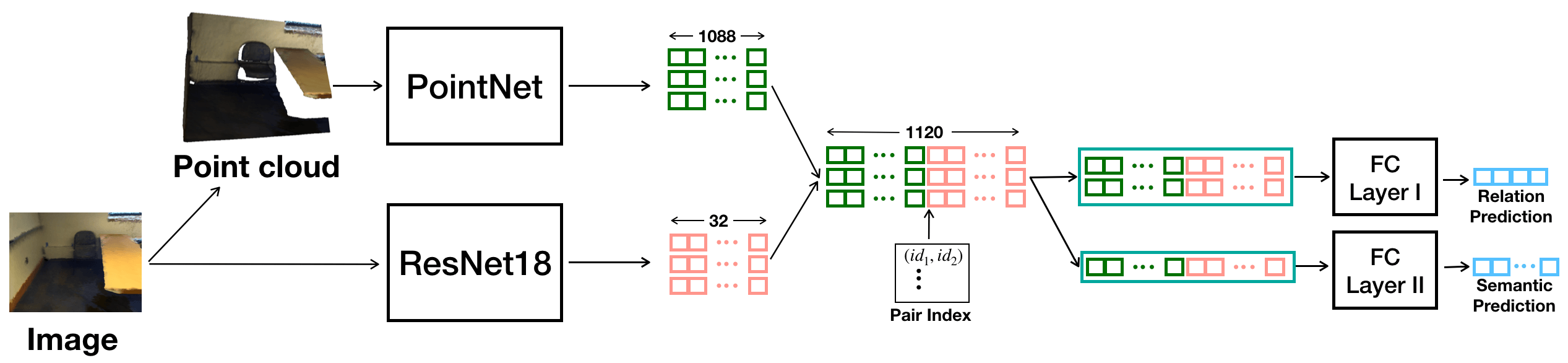} 
        \caption{local module} \label{fig:local_arch}
    \end{subfigure}
    \caption{Network architectures for different modules.}
\vspace{-0.15in}
\end{figure*}

\noindent\textbf{Network Architecture}
The planar patch module firstly extracts partial point cloud derived from depth image and then randomly sampled 8192 points as the input. To integer more information, we separated our network into two branches. The first branch predicts the plane's center and normal showing in the partial point cloud. The second branch predicts the room's structural plane center and normal which might not show in the partial point cloud. In the first branch, we use the PointNet-style network to predict a point-wise output. Each point of output contains the predicted plane normal, relative distance to the plane center, feature vector, and semantic label. In the second branch, we use another PointNet-style network to directly predict the structural plane normal and center. This branch capture precise global information of room. Please refer to Figure \ref{fig:plane_arch} for details.

\begin{figure}
\centering
\footnotesize
\def\imh{0.12\textwidth}
\def\imw{0.12\textwidth}
\newcommand{\T}[1]{\raisebox{-0.5\height}{#1}}
\setlength{\tabcolsep}{1pt}
\begin{tabular}{cccc}

\T{\includegraphics[width=\imw] {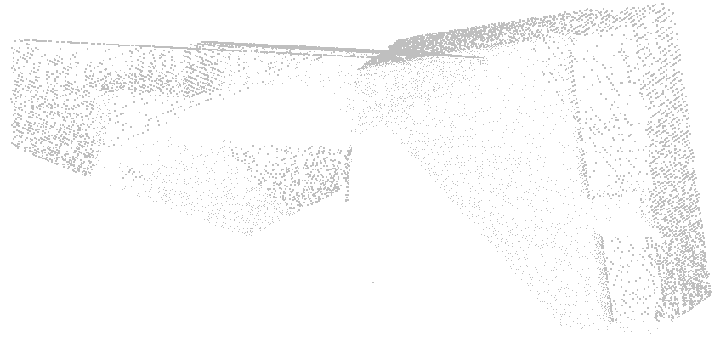}}&
\T{\includegraphics[width=\imw] {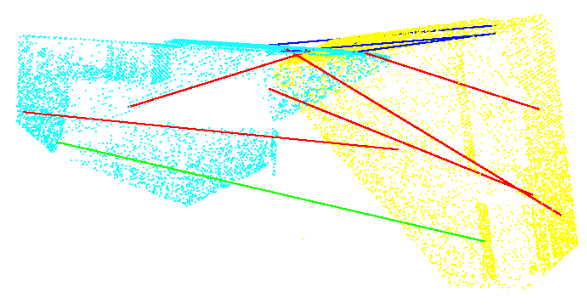}}&
\T{\includegraphics[width=\imw] {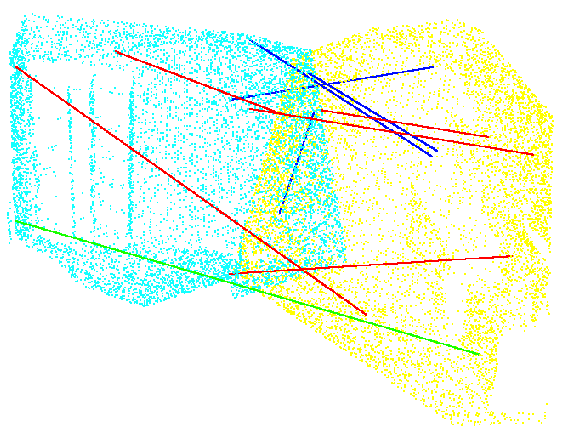}}&
\T{\includegraphics[width=\imw] {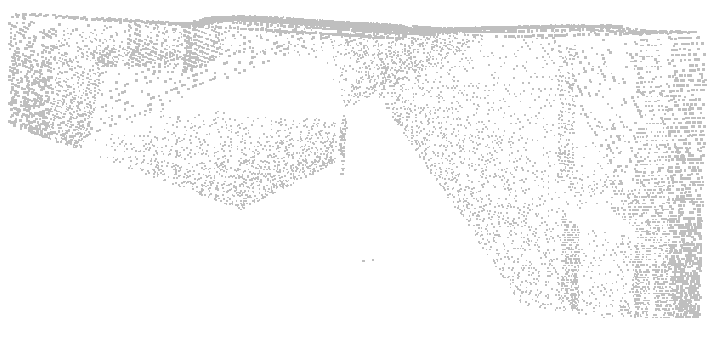}}
\\ [-3pt]

\T{\includegraphics[width=\imw] {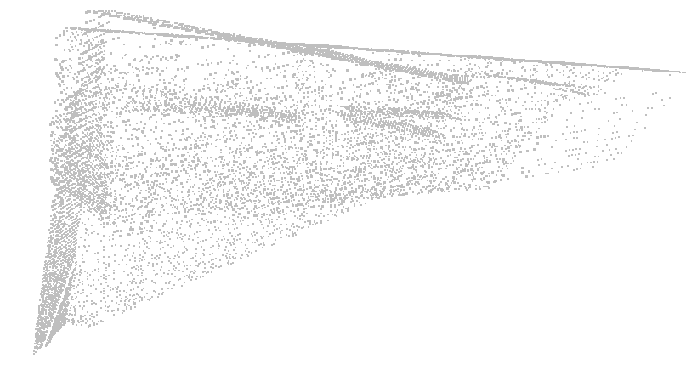}}&
\T{\includegraphics[width=\imw] {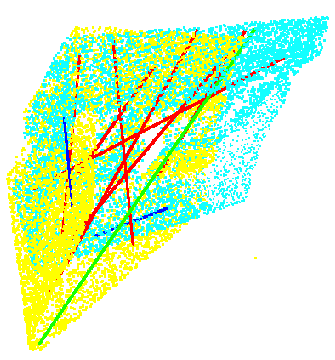}}&
\T{\includegraphics[width=\imw] {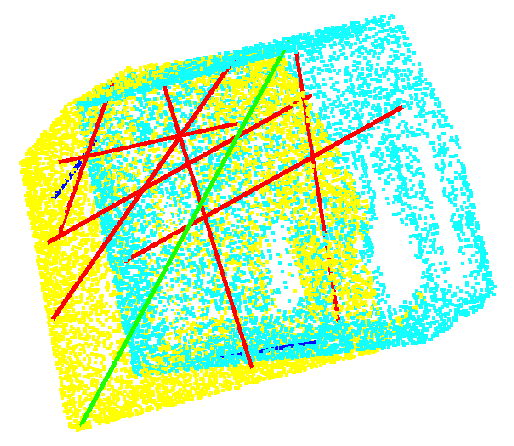}}&
\T{\includegraphics[width=\imw] {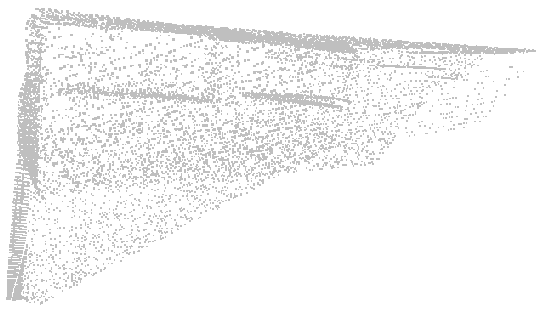}}
\\ [-3pt]
 
\T{\includegraphics[width=\imw] {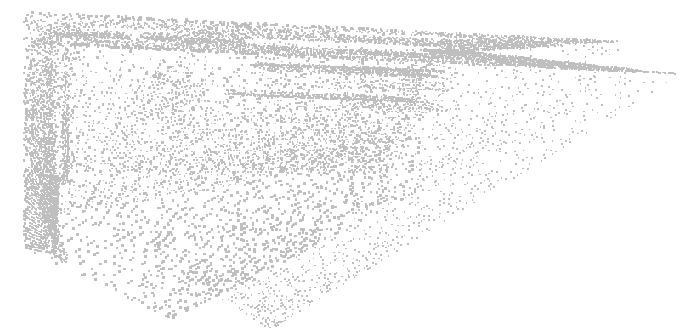}}&
\T{\includegraphics[width=\imw] {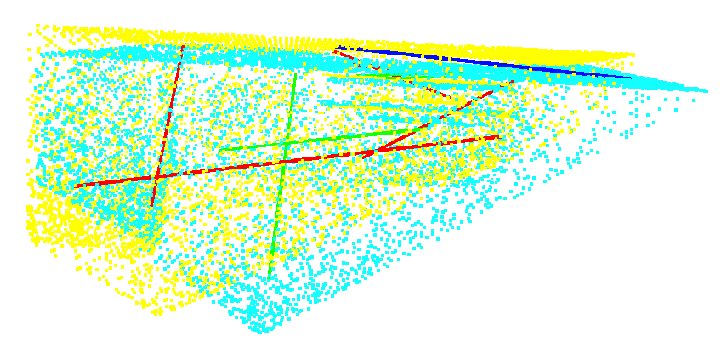}}&
\T{\includegraphics[width=\imw] {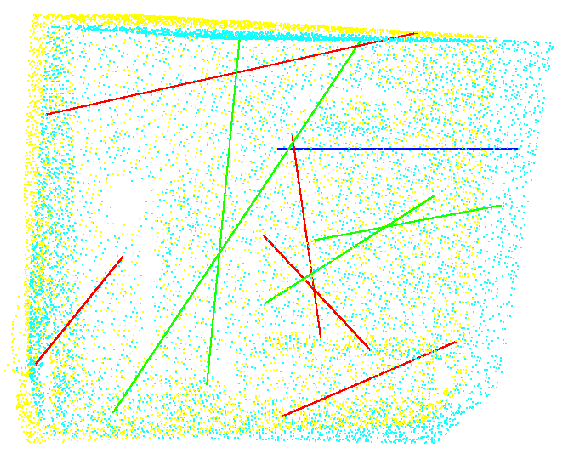}}&
\T{\includegraphics[width=\imw] {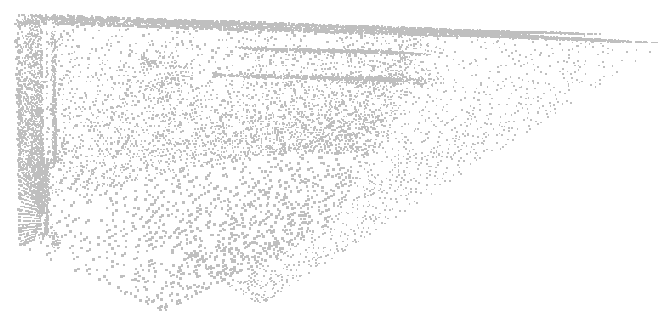}}
\\ [-3pt]

\end{tabular}

\caption{Local module results. The first column is the two input  scans, where there is noticeable mis-alignment error. The second column is visualization of detected geometric relation. Blue: coplane, red: perpendicular, green: parallel. The last column is the output of local module.}
\label{fig:supp_5}
\vspace{-0.15in}
\end{figure}

\noindent\textbf{Qualitative Results}
We show qualitative results in Figure \ref{fig:supp_2}.

\subsection{Local Relation Network}
\noindent\textbf{Network Architecture}
The input to our local relation network is 16384 points where the first 8192 points belongs to the source scan, and the later 8192 points belong to the target scan. We have 10 feature channels as input in total: we augment the xyz position with color, normal, and an {0,1} indicator indicates whether this point belongs to source or target. The input goes through a PointNet-style network to extract a per-point feature of dimension 1088. In order to provide the feature more local information, we also added an image branch similarly as in the 2D layout completion network to extract a 32 dimension convolutional per point feature. Then, using the sampled relations between source and target, we concatenate the corresponding feature vector (resulting a feature vector of 2*(1088+32) dimension) and output a relation prediction after 3 layers of fully connected layers. In addition to relation prediction, we also added per point semantic segmentation task in order to equip the feature with semantic information. We found adding the semantic prediction task increase the relation prediction's average accuracy by ~5\%. Please refer to Figure \ref{fig:local_arch} for details. \\

\noindent\textbf{Training details of local module}. Randomly sampled pairs have unbalanced distribution of relations. In order to reduce the data in-balance, we manually force equal portion of relations during training. We found this strategy works well in practice. 

\noindent\textbf{Qualitative Results}
We show qualitative results in Figure \ref{fig:supp_5}.

\begin{figure*}
\centering
\footnotesize
\def\imh{0.25\textwidth}
\def\imw{0.25\textwidth}
\newcommand{\T}[1]{\raisebox{-0.5\height}{#1}}
\setlength{\tabcolsep}{1pt}
\begin{tabular}{ccccc}

\rotatebox[origin=c]{90}{\textbf{360-image}} &
\T{\includegraphics[width=\imw] {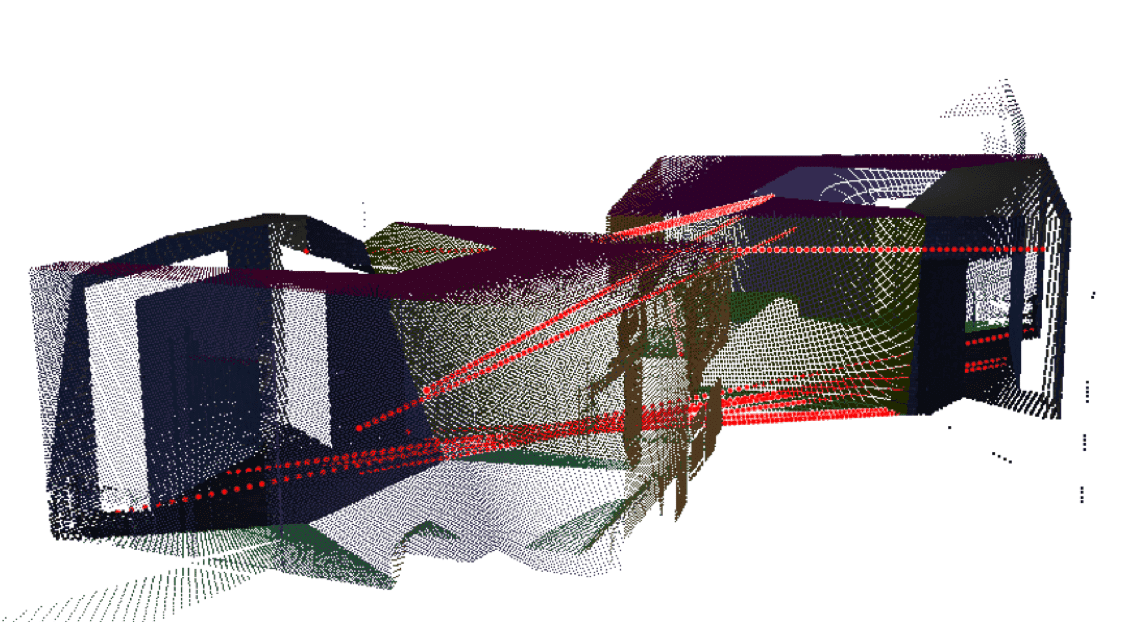}} & 
\T{\includegraphics[width=\imw] {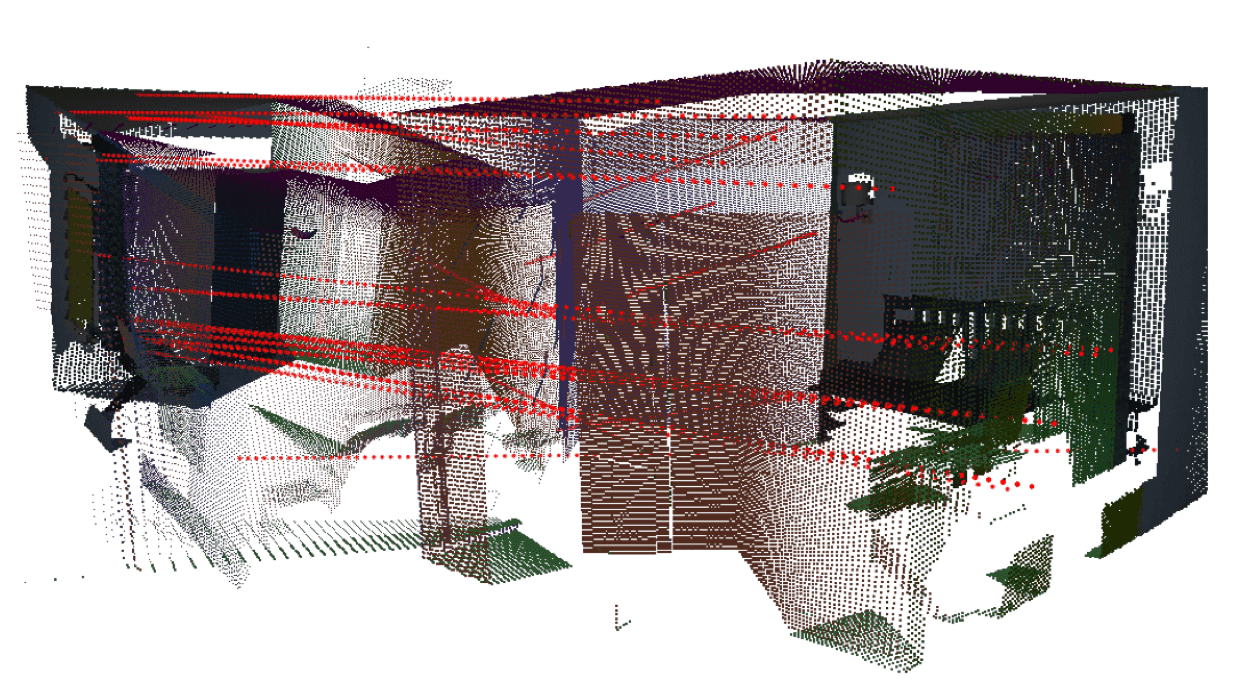}} & 
\T{\includegraphics[width=\imw] {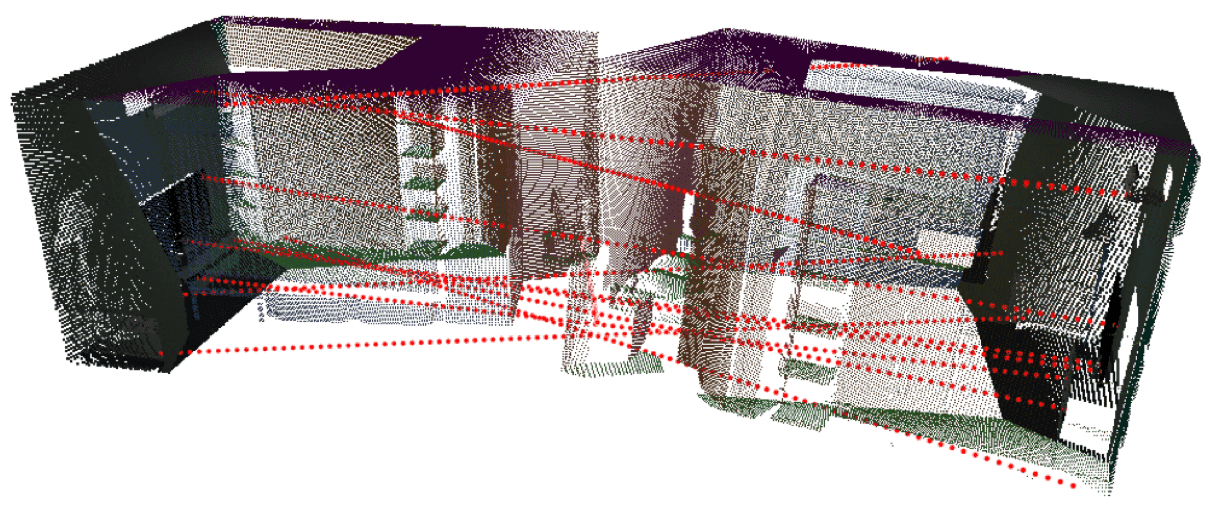}} & 
\T{\includegraphics[width=\imw] {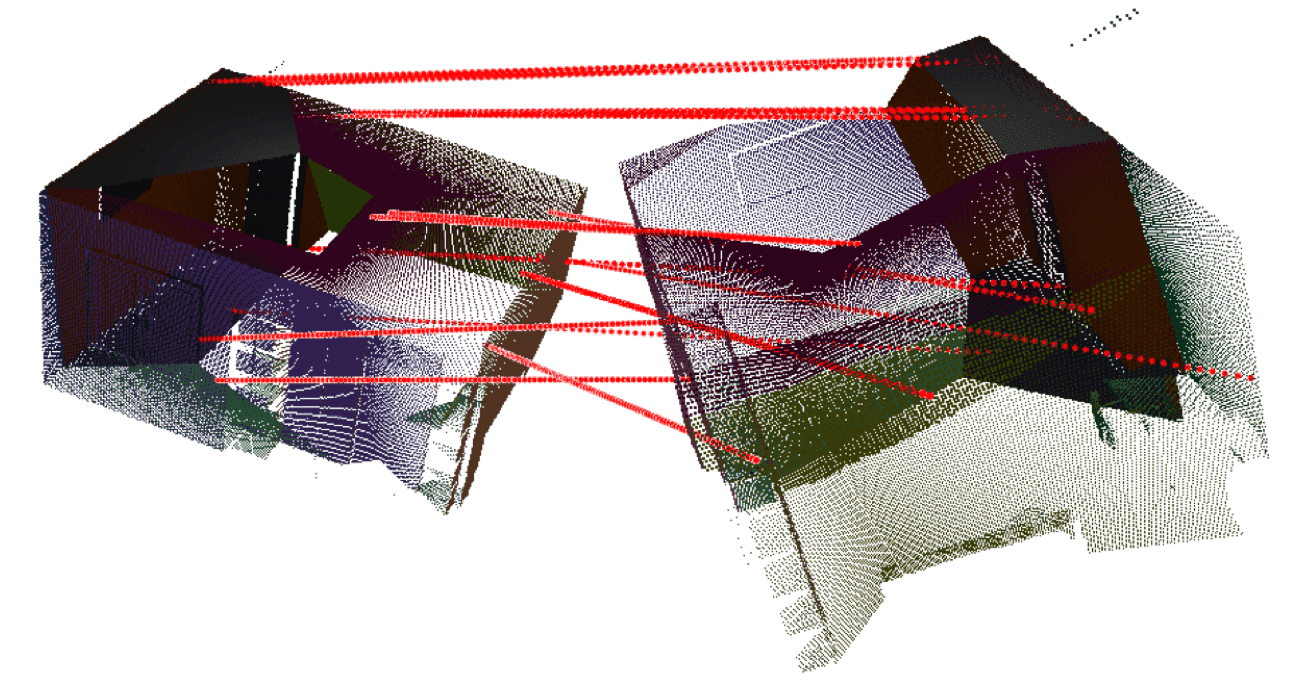}}  
\\ [+21pt]

\rotatebox[origin=c]{90}{\textbf{Planar patch}} &
\T{\includegraphics[width=\imw] {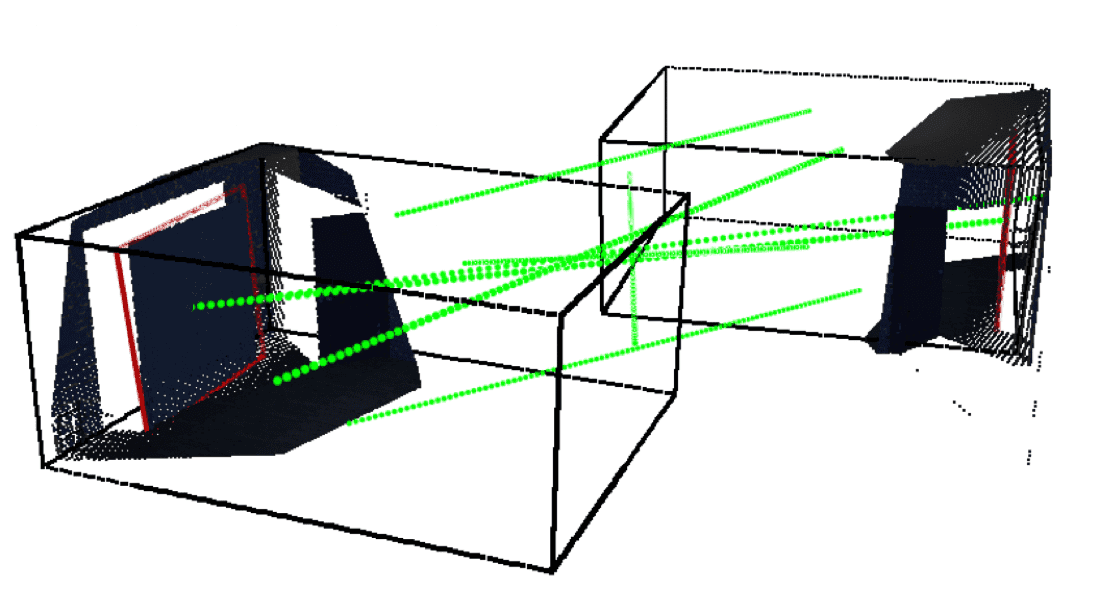}} & 
\T{\includegraphics[width=\imw] {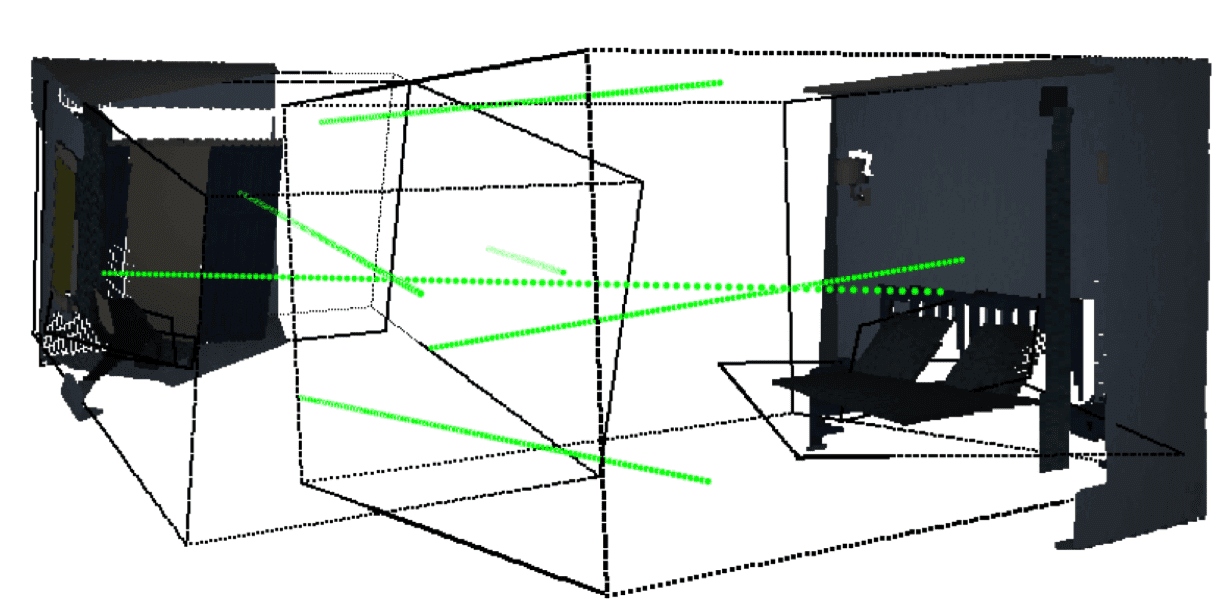}} & 
\T{\includegraphics[width=\imw] {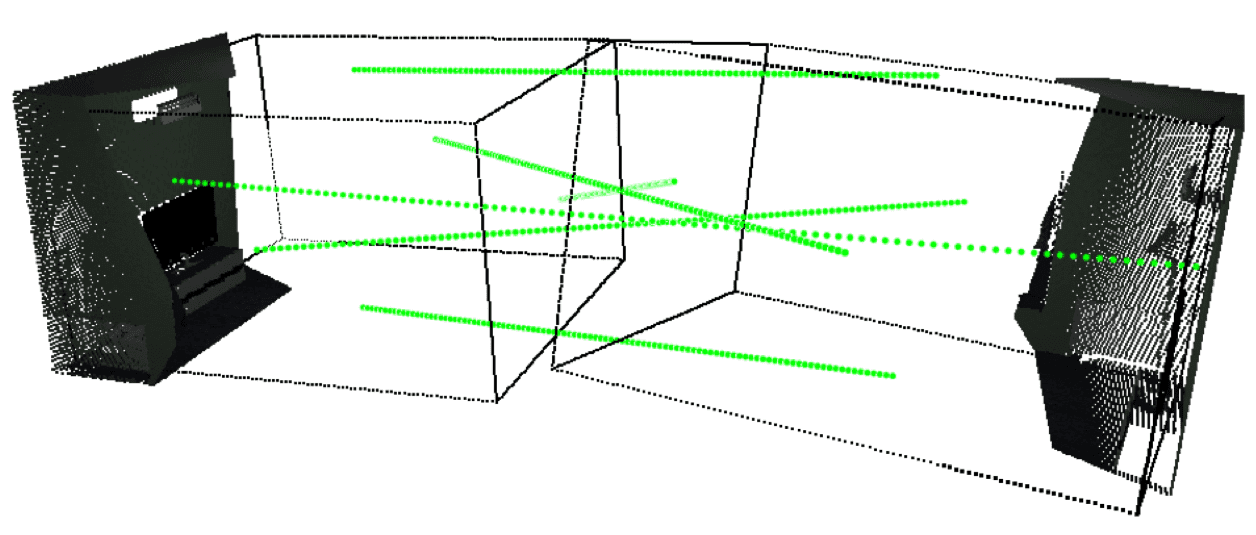}} & 
\T{\includegraphics[width=\imw] {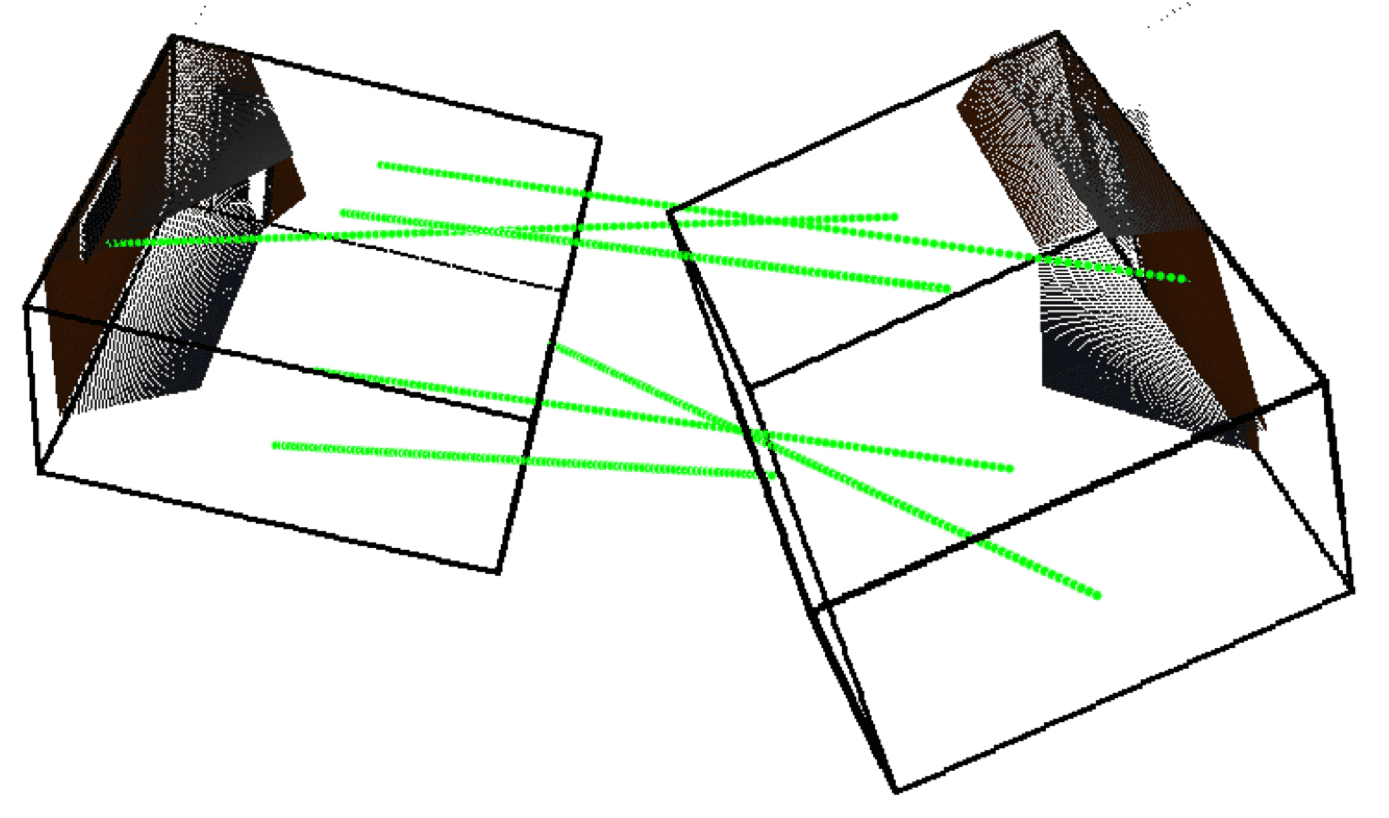}} 
\\ [+21pt]

\rotatebox[origin=c]{90}{\textbf{2D Layout}} &
\T{\includegraphics[width=\imw] {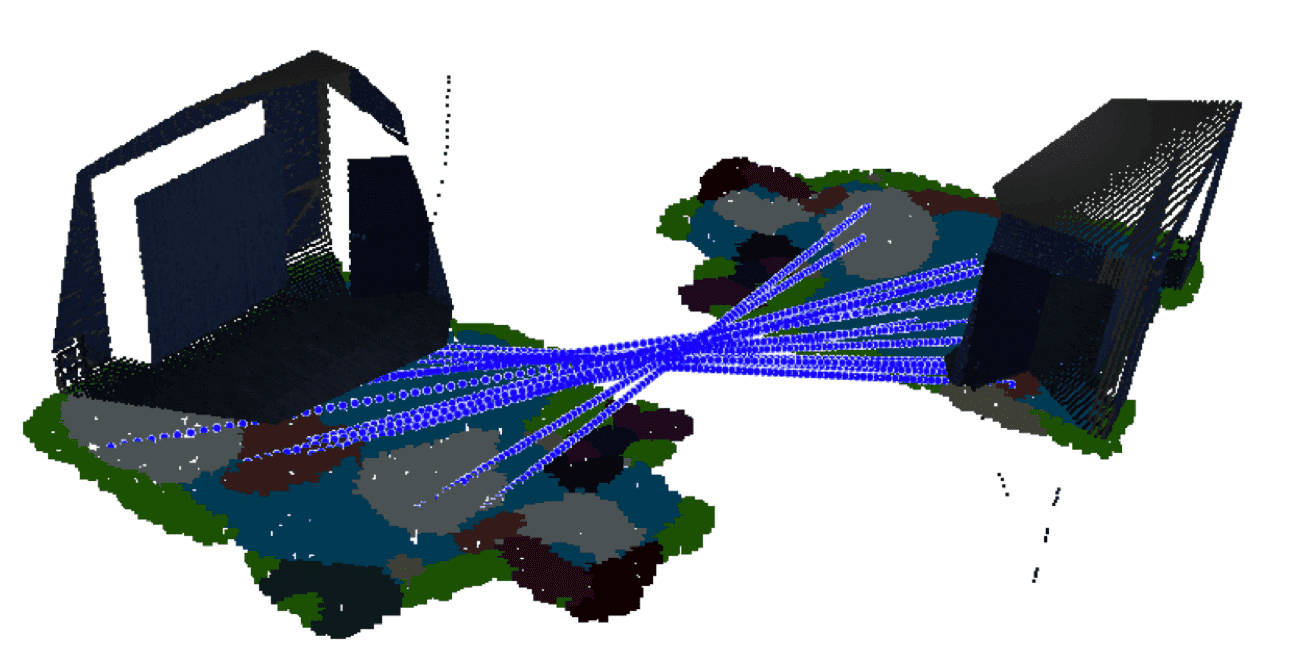}} & 
\T{\includegraphics[width=\imw] {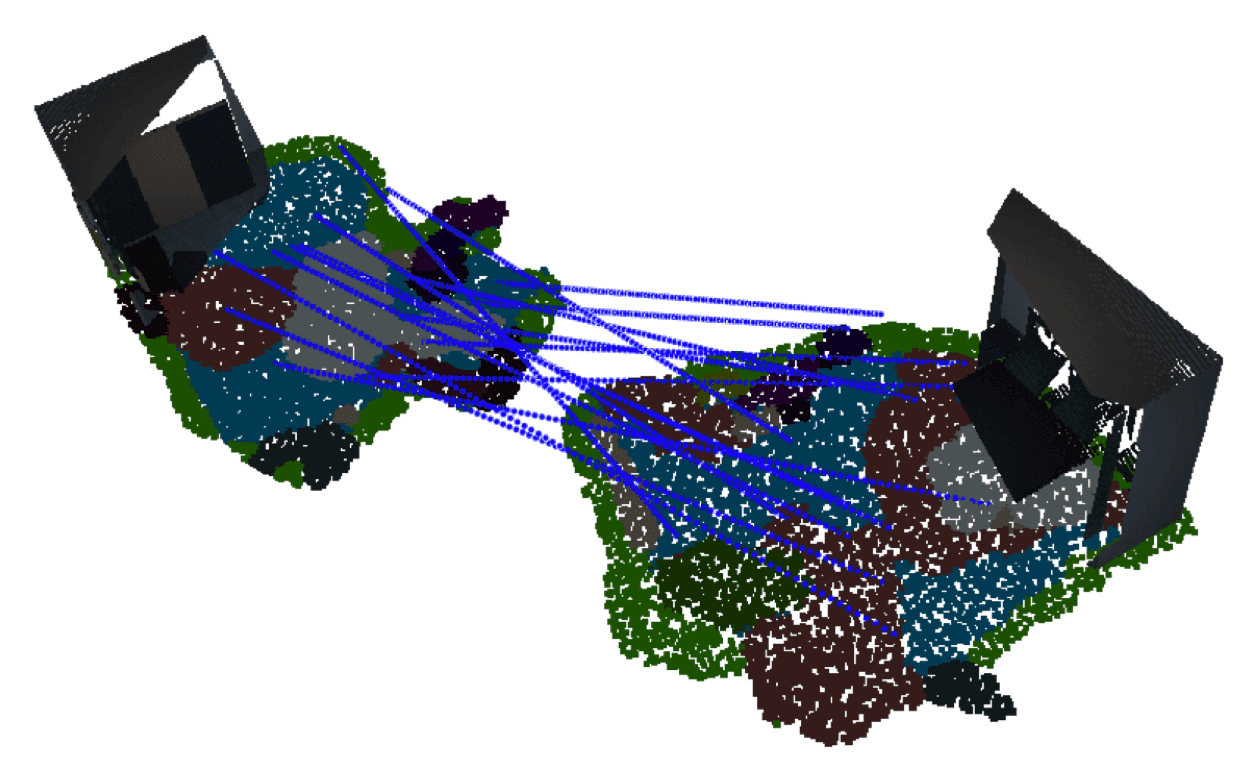}} & 
\T{\includegraphics[width=\imw] {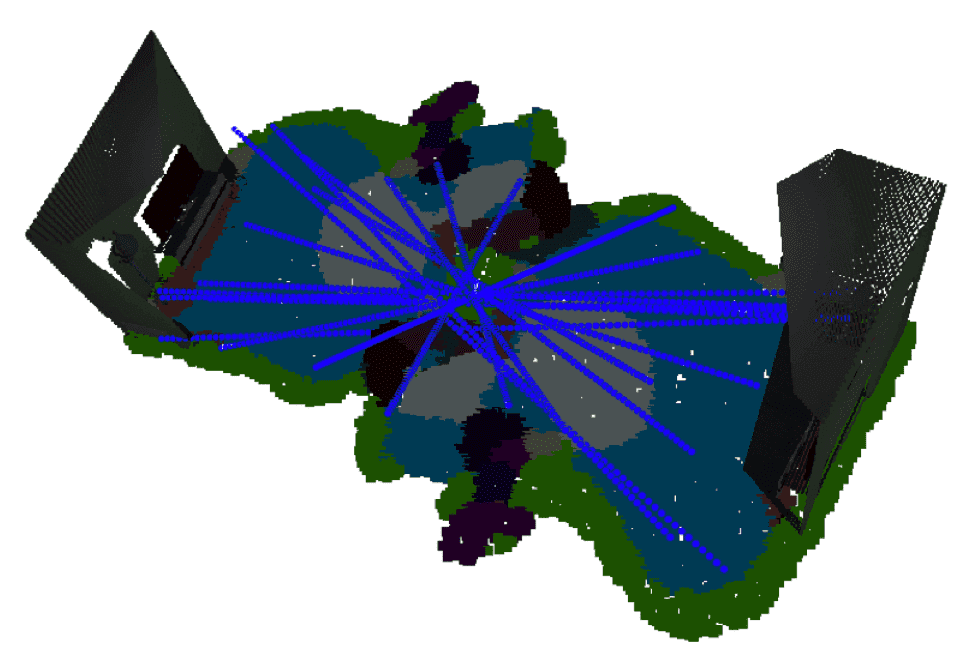}} & 
\T{\includegraphics[width=\imw] {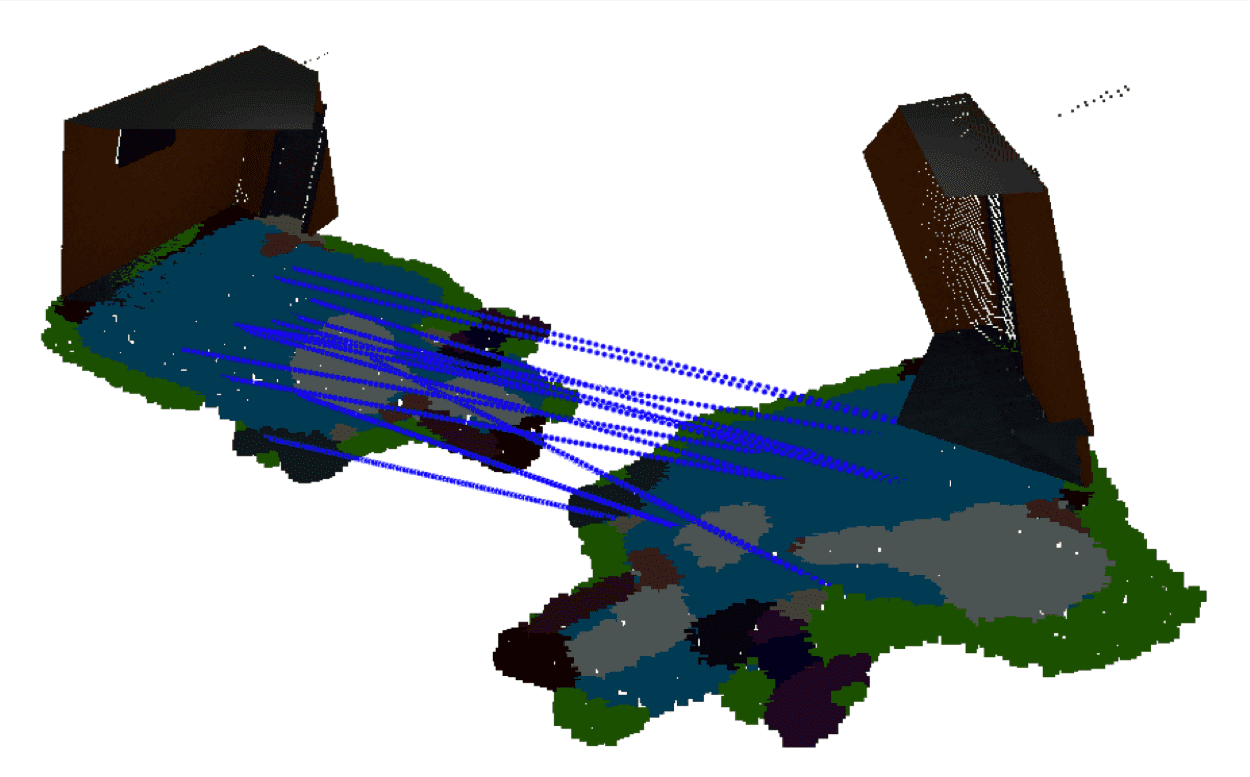}}  
\\ [+21pt]

\end{tabular}

\caption{Visualization of correspondences on three representations. Each column contains one example. The first row shows 360-image representation. The second row shows planar patch. The third row shows 2D-layout.}
\label{fig:supp_9}
\vspace{-0.15in}
\end{figure*}
\subsection{Visualization of Feature Correspondences}
We show the visualization of three representation's correspondences in Figure ~\ref{fig:supp_9}.

\subsection{Details of Spectral Matching Module}

In this section, we provide the technical details on the definition of the consistency matrix used in spectral matching. Specifically, consider two feature correspondences $(p_1,p_2)$ and $(p_1',p_2')$. With $\bs{p}_i$, $\bs{n}_i$, and $\bs{d}_i$ we denote the position, the normal, and the descriptor of $p_i$, respectively. Likewise, with $\bs{p}_i'$, $\bs{n}_i'$, and $\bs{d}_i'$  we denote the position, the normal, and the descriptor of $p_i'$, respectively. For each pair $(p_i,p_i')$, we consider four geometric quantities that are invariant under rigid motions, i.e., one distance and three angles:
\begin{align*}
l_i = \|\bs{p}_i-\bs{p}_i'\|, & \theta_{i,1} = \textup{angle}(\bs{n}_i,\bs{n}_i') \\
\theta_{i,2} = \textup{angle}(\bs{n}_i, \frac{\bs{p}_i-\bs{p}_i'}{\|\bs{p}_i-\bs{p}_i'\|}), & \theta_{i,3} = \textup{angle}(\bs{n}_i',\frac{\bs{p}_i-\bs{p}_i'}{\|\bs{p}_i-\bs{p}_i'\|}).
\end{align*}
With this setup, we define the consistency score between $p_1 p_1'$ and $p_2 p_2'$ as
\begin{align}
c(p_1 p_1', p_2 p_2') = \exp\Big(-\frac{\|\bs{d}_1-\bs{d}_1'\|^2+\|\bs{d}_2-\bs{d}_2'\|^2}{2\gamma_1^2} \nonumber\\
-\frac{(l_1-l_2)^2}{2\gamma_2^2}-\frac{(\theta_{1,2}-\theta_{1,1})^2}{2\gamma_3^2}\nonumber \\
-\frac{(\theta_{2,2}-\theta_{2,1})^2}{2\gamma_4^2}-\frac{(\theta_{3,2}-\theta_{3,1})^2}{2\gamma_5^2}\Big)
\label{Eq:Consistency:Score}
\end{align}
where $\gamma_i, 1\leq i \leq 5$ are hyper-parameters to be learned from data. 

\section{Additional Technical Details of The Training Procedure}

In this section, we provide additional technical details on the training procedure mentioned in the main paper.

\noindent\textbf{Back-propagation through implicitly defined functions.} Training $l_4$ requires us to compute the derivatives with respect to the optimal solution to an objective function. Here we represent a general formulation. Without losing generality, we assume we have an objective function $f(\bs{x},\alpha)$, where $\alpha$ denotes its hyper-parameters and/or input parameters, and where $\bs{x}$ denotes the variables to be optimized. Consider the optimal solution
\begin{equation} 
\bs{x}^{\star}(\alpha) := \underset{\bs{x}}{\textup{argmin}}\ f(\bs{x}, \alpha).
\label{Eq:F:Min}
\end{equation}
Our goal is to compute the derivatives of $\bs{x}^{\star}$ with respect to $\alpha$. To this end, notice that $\bs{x}^{\star}(\alpha)$ is a critical point of $f$. This means
\begin{equation}
\frac{\partial f}{\partial \bs{x}} (\bs{x}^{\star}, \alpha) = 0.
\label{Eq:Critical:Point}
\end{equation}
Computing the derivatives of both sides of (\ref{Eq:Critical:Point}) to $\alpha$. We arrive at
$$
\frac{\partial^2 f}{\partial^2 \bs{x}} (\bs{x}^{\star}, \alpha)\cdot \frac{\partial \bs{x}^{\star}(\alpha)}{\partial \alpha} + \frac{\partial^2 f}{\partial \bs{x}\partial \alpha} (\bs{x}^{\star}, \alpha)= 0.
$$
In other words
\begin{equation}
\frac{\partial \bs{x}^{\star}(\alpha)}{\partial \alpha} = -\left(\frac{\partial^2 f}{\partial^2 \bs{x}}\right)\cdot\frac{\partial^2 f}{\partial \bs{x} \alpha}
\label{Eq:Critical:Point2}
\end{equation}
(\ref{Eq:Critical:Point2}) allows us to optimize network parameters where the objective function involves $\bs{x}^{\star}$. Note that in this paper, $\bs{x}^{\star}\in \R^6$, so computing the second order derivatives are manageable. 

\noindent\textbf{Training details.}

\noindent\textit{360-image Completion Module} We use the same setting as \cite{Yang_2019_CVPR} without image warping process and recurrent module.

\noindent\textit{Planar Patch Completion Module} We trained the first branch of network with 60 epochs and the second branch with 30 epochs. The learning rate of both networks is $0.0002$.
\noindent\textit{2D Layout Completion Module/Local Module}
We train 60 epoch with initial learning rate $0.0002$.

\section{Additional Technical Details of the Experimental Setup}

\begin{table}
\small
\centering
\begin{tabular}{c|cccc}
 & No-relation & Co-planar & Perpend. & Parallel\\
 \hline
No-relation &0.806  &0.831& 0.077& 0.034\\
Co-planar   &0.667 &0.307& 0.019& 0.007\\
Perpend.    &0.568 &0.064& 0.314& 0.054 \\
Parallel    &0.628 &0.065& 0.006& 0.301
\end{tabular}
\caption{Confusion matrix of the prediction module. }
\label{Table:Prediction:CMat}
\end{table}

\noindent\textbf{Confusion matrix of the prediction module.} Table~\ref{Table:Prediction:CMat} shows the confusion matrix of the prediction module. Note that although the prediction module does not deliver accurate predictions. However, since we maintain a broad set of point-pairs, the predicted pairs still contain sufficient constraints for regressing the underlying relative pose. To extract correct relations, our approach utilizes robust reweighted non-linear squares and the fact that we have an initial pose to begin with, which also helps to prune wrong predictions, e.g., points with similar normals but are predicted to be perpendicular. 
\end{document}